\documentclass[acmsmall]{acmart}
\DeclareMathOperator*{\argmin}{argmin}
\usepackage{dsfont}
\usepackage{multirow}
\usepackage{array}
\usepackage{algorithm}
\usepackage{algpseudocode}
\usepackage{stackengine,scalerel}
\usepackage{wrapfig}

\AtBeginDocument{%
  }

\makeatletter
\def\@ACM@checkaffil{
    \if@ACM@instpresent\else
    \ClassWarningNoLine{\@classname}{No institution present for an affiliation}%
    \fi
    \if@ACM@citypresent\else
    \ClassWarningNoLine{\@classname}{No city present for an affiliation}%
    \fi
    \if@ACM@countrypresent\else
        \ClassWarningNoLine{\@classname}{No country present for an affiliation}%
    \fi
}
\makeatother

\setcopyright{none}

\acmJournal{TIST}
\acmYear{2026} 
\acmVolume{1} \acmNumber{1} \acmArticle{}
\acmMonth{1}
\acmDOI{10.1145/3815185}

\begin{document}

\title{Gradual Domain Adaptation for Graph Learning}



\author{Pui Ieng Lei}
\email{yc37460@um.edu.mo}
\orcid{0009-0005-6771-4598}
\affiliation{\institution{Faculty of Science and Technology, University of Macau}  \city{Macau}  \country{China}}
\author{Ximing Chen}
\affiliation{\institution{Faculty of Science and Technology, University of Macau}  \city{Macau}  \country{China}}
\email{yc37921@um.edu.mo}
\author{Yijun Sheng}
\affiliation{\institution{Faculty of Science and Technology, University of Macau}  \city{Macau}  \country{China}}
\email{yc17419@um.edu.mo}
\author{Yanyan Liu}
\affiliation{\institution{Faculty of Science and Technology, University of Macau}  \city{Macau}  \country{China}}
\email{yc07402@um.edu.mo}
\author{Zhiguo Gong}
\affiliation{\institution{Faculty of Science and Technology, University of Macau}  \city{Macau}  \country{China}}
\email{fstzgg@um.edu.mo}
\author{Qiang Yang}
\affiliation{\institution{PolyU Academy for Artificial Intelligence, Hong Kong Polytechnic University}  \city{Hong Kong}  \country{China}}
\email{profqiang.yang@polyu.edu.hk}

\renewcommand{\shortauthors}{Lei et al.}

\begin{abstract}
Existing machine learning literature lacks graph-based domain adaptation techniques capable of handling large distribution shifts, primarily due to the difficulty in simulating a coherent evolutionary path from source to target graph. To meet this challenge, we present a \emph{graph gradual domain adaptation} (GGDA) framework, which constructs a compact domain sequence that minimizes information loss during adaptation. 
Our approach starts with an efficient generation of knowledge-preserving intermediate graphs over the Fused Gromov-Wasserstein (FGW) metric. A GGDA domain sequence is then constructed upon this bridging data pool through a novel vertex-based progression, which involves selecting "close" vertices and performing adaptive domain advancement to enhance inter-domain transferability. Theoretically, our framework provides implementable upper and lower bounds for the intractable inter-domain Wasserstein distance, $W_p(\mu_t,\mu_{t+1})$, enabling its flexible adjustment for optimal domain formation. Extensive experiments across diverse transfer scenarios demonstrate the superior performance of our GGDA framework. Our code is available at \url{https://github.com/Louisa-LPI/GGDA}.
\end{abstract}



\received{28 August 2025}
\received[revised]{03 December 2025}
\received[revised]{24 February 2026}
\received[revised]{15 April 2026}
\received[accepted]{24 April 2026}

\maketitle
\footnotetext{\copyright\ [Pui Ieng Lei, Ximing Chen, Yijun Sheng, Yanyan Liu, Zhiguo Gong, and Qiang Yang] [2026]. This is the author's version of The Work. It is posted here for your personal use. Not for redistribution. The definitive version was published in ACM Trans. Intell. Syst. Technol., \url{https://doi.org/10.1145/3815185}.}

\section{Introduction}
Domain adaptation (DA) provides a powerful framework for transferring knowledge from a label-rich source domain to a relevant but label-scarce target domain under data distribution shifts \cite{pan2009survey}.
Traditionally, most DA algorithms rely on the assumption that data samples within each domain are independent and identically distributed (IID) \cite{ganin2016domain, bousmalis2016domain, long2017deep, tzeng2017adversarial, long2018conditional, wu2019domain, singh2021clda}. 
Recently, this assumption is increasingly challenged by the prevalence of non-IID data in real-world applications, such as social and traffic networks. Coupled with the rapid advancement of graph neural networks (GNNs) \cite{defferrard2016convolutional, kipf2016semi, velivckovic2017graph, hamilton2017inductive, xu2018powerful}, this trend has spurred growing interest in DA for graph-structured data \cite{wu2020unsupervised, zhang2021adversarial, mao2022augmenting, liu2023structural, wu2023non, liu2024rethinking, huang2024can, wang2024open, liu2024pairwise, chen2025smoothness}.

In graph DA, source and target graphs are defined not only by their data objects (nodes) but also by the relations between them (edges). Consequently, unlike in IID settings, data representations in graph DA must encode information from both individual nodes and their local neighborhoods. A standard approach involves learning a graph encoder that captures data dependencies via neighborhood aggregations while simultaneously aligning the source and target distributions in a latent space to facilitate transfer on a domain-invariant basis \cite{shen2020adversarial, dai2022graph, wu2020unsupervised, liu2023structural, qiao2023semi, wu2023non, you2023graph}. Nevertheless, theoretical analysis indicates that domain-invariant techniques often perform poorly under large marginal distribution shifts due to information distortion \cite{zhao2019learning}. This problem is further amplified in non-IID graph scenarios, where node attributes and graph structure may shift jointly. Motivated by these challenges, our paper delves into the critical problem of graph domain adaptation under large distribution shifts. We adhere to an unsupervised domain adaptation (UDA) setting where the target domain is entirely unlabeled.

In IID-based UDA, to handle larger distribution shifts, several algorithms integrate the use of unlabeled intermediate domains between source and target domains \cite{tan2017distant, zhu2017unpaired, hsu2020progressive, xu2020adversarial, na2021fixbi}. A specific line of research investigates \emph{gradual domain adaptation} (GDA), which leverages iterative
model adaptations across a sequence of intermediate domains that transition gradually from source to target \cite{tan2015transitive, abnar2021gradual, chen2021gradual, wang2022understanding, sagawa2022gradual, he2023gradual}. 
Conceptually, GDA decomposes a single UDA problem with a large domain gap into a series of smaller ones with reduced gaps, thereby providing a tighter bound on the generalization error. While such intermediate sequences are occasionally available from data collection (e.g., portraits taken across years or evolving social networks), it is often difficult or impossible to obtain suitable intermediate data.
To overcome this, IID methods generate virtual intermediate data through techniques like interpolation and adversarial training \cite{xu2020adversarial, na2021fixbi, abnar2021gradual, he2023gradual}.

Despite its promising performance, GDA has been confined to IID settings, and its applicability to complex non-IID data remains underexplored. Given the growing ubiquity of graph-structured data, it is critical to prevent the performance of graph DA models from plummeting under large shifts --- a scenario where GDA is poised to be highly valuable. Motivated by this gap, we introduce and explore a novel \emph{graph gradual domain adaptation} (GGDA) framework from theoretical and empirical perspectives. 
Assuming no intermediate data is initially available, our work addresses the central question: \emph{In a non-IID context of graph-structured data, how can we construct an intermediate domain sequence to ensure high target accuracy in GGDA?}

To answer this, we first identify two key criteria: (1) the intermediate domains should be constructed to \emph{minimize information loss} during adaptation, and (2) the construction scheme should allow for the  \emph{flexible adjustment} of the inter-domain Wasserstein distance, $W_p(\mu_t,\mu_{t+1})$, to optimize domain formation.
In a graph context, this presents two primary challenges: (1) $W_p(\mu_t,\mu_{t+1})$ is inherently implicit due to the lack of an explicitly defined structure space for accommodating all possible instance relations, and (2) the generation of each instance must jointly account for both the node itself and its neighborhood structure.


Based on these considerations, we propose a novel GGDA framework. We first develop an efficient method to generate a sequence of intermediate graphs with weighted Fr\'echet mean over the Fused Gromov-Wasserstein (FGW) metric \cite{vayer2019}, a graph distance metric defined as the minimal cost of transporting both node features and structural relations between two graphs. In this way, we bypass the need to know the implicit cross-domain structures. Our proposed entropy-guided matching strategy, combined with coupling constraints in FGW computations, maximizes the infusion of valid and discriminative knowledge from the original graphs into the generated data.
Upon this generated data bridge, we introduce a novel domain progression scheme to construct a sequence of GGDA domains with enhanced inter-domain compactness, facilitating robust knowledge transfer. This scheme comprises two modules: (1) a regularized vertex selection process for constructing the subsequent domain, and (2) an adaptive mass decay mechanism for removing obsolete nodes from the current domain. By traversing the generated graphs and populating each domain with "close" vertices, the resulting cross-domain adaptation effectively preserves the most transferable and discriminative knowledge.

Theoretically, we demonstrate that the otherwise intractable  $W_p(\mu_t,\mu_{t+1})$ is now concretized by its lower bound from the FGW metric and its upper bound from our vertex selection regularization. Furthermore, a hyperparameter governing the scale of vertex selection under our prioritization enables the flexible adjustment of $W_p(\mu_t,\mu_{t+1})$ to optimize domain formation without the need for expensive, repetitive data generation. Our key contributions are summarized as follows:
\begin{itemize}
\item To the best of our knowledge, we are the first to investigate \emph{gradual domain adaptation on graphs} and propose a novel solution specifically tailored to overcome \emph{large domain gaps} in non-IID graph DA tasks.
\item Our GGDA framework constructs a compact domain sequence to enable stable knowledge transfer. We address the fundamental challenge of structural implicitness by designing tractable modules with provable bounding effects on the inter-domain distance.
\item We conduct extensive experiments on fifteen benchmark datasets. The results consistently validate our theoretical insights and demonstrate the framework's effectiveness, showing an average performance gain of $+5.02\%$ over the best baseline. 
\end{itemize}

\section{Related Works}
This section provides a concise overview of prior research. A comprehensive discussion is available in the appendix.

\subsection{Graph Domain Adaptation} 
The prevailing paradigm for node-level graph domain adaptation employs adversarial training to learn domain-invariant representations that can fool a domain discriminator \cite{shen2020adversarial, wu2020unsupervised, zhang2021adversarial, dai2022graph, guo2022learning, mao2022augmenting, qiao2023semi, wu2023non, liu2023structural, you2023graph, liu2024pairwise, liu2024rethinking, wang2024open, chen2025smoothness}. Beyond adversarial alignment, alternative strategies have been explored, such as ego-graph information maximization \cite{zhu2021transfer}, spectral regularization that modulates the GNN Lipschitz constant for error bounding \cite{you2023graph}, and generation of a new source graph \cite{huang2024can}. Despite these advances, no existing approaches have addressed the critical scenario of large distribution gaps between source and target graphs.
Graph-level domain adaptation, on the other hand, focuses on graph classifications and treats each graph as an individual sample, more closely resembling traditional IID settings \cite{yin2022deal, yin2023coco}.

\subsection{Intermediate Domains for IID Domain Adaptation} 
Leveraging intermediate domains has proven effective for IID domain adaptation \cite{gong2012geodesic, gopalan2011domain, tan2015transitive, tan2017distant, hsu2020progressive, xu2020adversarial, na2021fixbi, abnar2021gradual, zhu2017unpaired, chen2021gradual, he2023gradual, kumar2020understanding, peyre2019computational, wang2022understanding}. 
Early approaches constructed these domains as subspaces on manifolds (e.g., Grassmann) \cite{gong2012geodesic, gopalan2011domain}, while later works generate them at the sample level to better integrate with deep learning techniques \cite{gong2019dlow, xu2020adversarial, na2021fixbi, abnar2021gradual, he2023gradual}. 
A specific strategy, gradual domain adaptation (GDA), progressively adapts a source model to the target along a path of intermediate domains, which are either retrieved from data or generated via interpolation and mixup strategies \cite{tan2015transitive, kumar2020understanding, chen2021gradual, wang2022understanding, he2023gradual, sagawa2022gradual, shi2024adversarial}. Despite its success in IID settings, the potential of intermediate domains and GDA for non-IID graph data has not been investigated, leaving a critical gap in handling large graph distribution shifts.

\subsection{Optimal Transport}
The optimal transport (OT) framework provides a principled way to quantify the discrepancy between probability distributions by computing the minimal cost required to transport mass from one distribution to another. The Wasserstein distance was first proposed as a metric between distributions dwelling in a common metric space \cite{kantorovich1960mathematical, rubner2000earth}, whereas the Gromov-Wasserstein (GW) distance was developed to be applied across different metric spaces by comparing their intrinsic relational structures \cite{peyre2019computational, memoli2011gromov}. Fused Gromov-Wasserstein (FGW) distance was proposed recently as a combination of the two distances above \cite{vayer2019, vayer2020}. The OT framework has found versatile applications in graph learning, including graph clustering, graph partitioning, graph representation learning, etc. \cite{vayer2019, vayer2020, xu2019scalable, xu2019gromov, chowdhury2021generalized, vincent2021online, chen2020optimal, ma2024fused, peyre2016gromov, li2023convergent}. 

\section{Preliminaries}
This section establishes our formal setting, including: (1) key mathematical notations, (2) definitions of graph space and domain distributions, and (3) specification of models and learning objectives in GGDA.

\subsection{Notations}
A simplex of $n$ bins is denoted by $\Sigma_n = \{h\in \mathbb{R}_+^n \mid \sum_i h_i = 1\}$. For two histograms $h \in \Sigma_n$ and $h' \in \Sigma_m$, we denote the set of all couplings between $h$ and $h'$ by $\Pi (h, h') = \{\pi \in \mathbb{R}_+^{n\times m} \mid \sum_i \pi_{i,j} =h'_j; \sum_j \pi_{i,j} =h_i\}$. For a space $\Omega$ and $x \in \Omega$, $\delta_x$ denotes the Dirac measure in $x$. The support of probability measure $\mu \in P(\Omega)$ is denoted by $supp(\mu)=\argmin_A\{|A| \mid A\subset \Omega; ~\mu(\Omega\symbol{92} A)=0\}$. In addition, Table \ref{notations} shows all the key notations used in this paper.

\begin{table}
\footnotesize
    \setlength{\tabcolsep}{0.8pt}
    \renewcommand{\arraystretch}{0.9}
    \caption{List of key notations in this paper.}
    \begin{tabular}{cc}
        \toprule
        \textbf{Notations}
                    &\textbf{Explanations}\\
        \midrule
        {$h \in \Sigma_n$} & {Histogram $h$ with $n$ bins}\\
        {$\pi \in \Pi (h, h')$} & {Coupling between histograms $h$ and $h'$} \\
        {$\Omega$, $\delta_x$} & {Space $\Omega$, Dirac measure in element $x$} \\
        {$\mu \in P(\Omega)$, $supp(\mu)$} & {Probability measure $\mu$ and its support} \\
        {$\mathcal{G} = (\mathcal{V}, \mathcal{E})$} & {Graph $\mathcal{G}$ with vertex set $\mathcal{V}$ and edge set $\mathcal{E}$} \\
        {$x_i, a_i, z_i, y_i$} & {Feature, structure, embedding representations and class label of vertex $v_i$} \\
        {$d_x, C, d_z, d_y$} & {Metrics on $\Omega_x$, $\Omega_a$, $\Omega_z$, and $\Omega_y$} \\
        {$\theta \in \Theta$} & {Model parameters}\\
        {$M_{\theta} = \Psi_{\theta} \circ \Phi_{\theta}$} & {Model $M$ as a composite of graph encoder $\Phi$ and classifier $\Psi$}\\
        {$\mu_0=\mu_S$, $\mu_T$} & {Source and target domains (i.e., graph distributions)}\\
        {$\mu_1, ..., \mu_{T-1}$} & {Sequence of $T-1$ constructed intermediate domains}\\
        {$\epsilon_t(\theta) \equiv \epsilon_{\mu_t}(\theta)$} & {Population loss in $t$-th domain}\\
        {$W_p(\mu, \nu)$, $\Delta$} & {$p$-Wasserstein distance between distributions, $\Delta = \frac{1}{T}\sum_{t=0}^{T-1} W_{p}(\mu_t,\mu_{t+1})$}\\
        {$FGW(\mu, \nu)$} & {Fused Gromov-Wasserstein distance between distributions} \\
        {$\hat{\mu}$, $\tilde{\mu}_k$} & {Graph partitions, $k$-th intermediate graph generated over FGW} \\
        \multirow{2}{10em}{$\mathcal{V}_B, \mathcal{V}_u^t, \mathcal{V}_l^t, \mathcal{V}_{pl}^t, \mathcal{V}_{pll}^t$} & {Vertex set from: entire generated sequence, unused vertices in the pool,} \\
        & {current domain $\mu_t$, selected vertices to pseudo-label, union of $\mathcal{V}_l^t$ and $\mathcal{V}_{pl}^t$} \\
        {$\hat{y}$, $c^t$, $\hat{c}^t$, $w^t$} & {Pseudo-label or ground-truth label, regularized score, label score, mass decay mask} \\
        {$\eta, \kappa, \beta$} & {Hyperparameters for score penalty, selection constraint, and mass decay rate} \\
        \bottomrule
    \end{tabular}
    \label{notations}
\end{table}

\subsection{Graph Space}
In this paper, a graph is viewed as a distribution sampled from a product space of features, structures, and classes.
An undirected attributed graph $\mathcal{G} = (\mathcal{V}, \mathcal{E})$ contains the set of vertices $\mathcal{V}$ and edges $\mathcal{E}$. Each vertex $v_i \in \mathcal{V}$ has (1) a feature representation $x_i$ in space $(\Omega_x, d_x)$, (2) a structure representation $a_i$ in space $(\Omega_a, C)$, and (3) a class label $y_i$ in space $(\Omega_y, d_y)$.

Specifically, the feature space $\Omega_x$ is a Euclidean space, with $d_x$ being the Euclidean distance. 
For notational simplicity, we present our formulations using binary classification in this paper, with node labels $y_i \in \{-1, 1\}$ and $\Omega_y := \mathbb{R}$.\footnote{Our theoretical proofs naturally generalize to multi-class settings, as detailed in Appendix F.}
The structure space $\Omega_a$ is an implicit space, upon which the exact positions of nodes are not observed, and $C: \Omega_a \times \Omega_a \rightarrow \mathbb{R}_+$ measures the distance between pairs of nodes reflected by their structures in the graph, such as node adjacency or shortest path distance. Note that in graph DA setting, since relational semantics persist across domains (e.g., for citation networks from different time periods, edges always represent citation relationships), we assume a shared ground structure space $(\Omega_a, C)$ for all graphs. In other words, the structural distance between nodes from different graphs is well-defined by $C$, even if the corresponding inter-graph edges  (e.g., potential citations between papers from different time periods) are unobserved. 


For a graph with $n$ vertices, a histogram $h \in \Sigma_n$ describes each vertex's relative importance within the graph (e.g., uniform or degree distribution). This allows the graph to be represented as $\mu = \sum_{i=1}^n h_i \delta_{(x_i,a_i,y_i)}$, a fully supported probability measure over the product space $\Omega_x \times \Omega_a \times \Omega_y$. 
Given that not all labels are observed, the graph can also be represented in a reduced product space $\Omega_x \times \Omega_a$, i.e., $\mu = \sum_{i=1}^n h_i \delta_{(x_i,a_i)}$. 

In a graph domain adaptation setting, we view all graphs concerned to be drawn from a shared space $\Omega_x \times \Omega_a \times \Omega_y$, but with shifted probability distributions. 



\subsection{Models and Objectives}
Consider a node classification task, where the goal is to predict node labels from node features and graph topology. Given a model family $\Theta$, for each $\theta \in \Theta$, a model $M_{\theta}: \Omega_x \times \Omega_a \rightarrow \Omega_y$ outputs the classification predictions. Each model can be viewed as a composite function with respect to some Euclidean embedding space $(\Omega_z, d_z)$, i.e., $M_{\theta} = \Psi_{\theta} \circ \Phi_{\theta}$, where $\Phi_{\theta}: \Omega_x \times \Omega_a \rightarrow \Omega_z$ (graph encoder) and $\Psi_{\theta}: \Omega_z \rightarrow \Omega_y$ (classifier). 

Consider $T+1$ graph distributions (i.e., domains), denoted by $\mu_0, \mu_1, ..., \mu_T$. For unsupervised graph DA, $\mu_0 = \mu_S$ is a fully-labeled source domain and $\mu_T$ is a fully-unlabeled target domain. For GGDA, $\mu_1,...,\mu_{T-1}$ are unknown intermediate domains transitioning from source to target. The population loss in $t$-th domain is defined as
\begin{gather}
\epsilon_t(\theta) \equiv \epsilon_{\mu_t}(\theta) = \mathbb{E}_{x,a,y \sim \mu_t}[\ell(M_{\theta}(x, a), y)],
\end{gather}
where $\ell$ is the loss function. The goal is to find a model $M_{\theta}$ that yields a low classification error on the target graph, i.e., low $\epsilon_T(\theta)$. 

\section{Assumptions}
Our framework is built upon the following assumptions. Further discussion is provided in the appendix.

\paragraph{Assumption 1} (Covariate shift) Assume the conditional distribution $p(y|x,a)$ is invariant over the ground space \cite{shimodaira2000improving}.

\paragraph{Assumption 2} (Metrics on product spaces) Assume the metric on $(\Omega_x \times \Omega_a, d_{(x,a)})$ is a linear combination of the metrics on the respective space, i.e., $d_{(x,a)}((x,a), (x',a')) = (1-\alpha)d_x(x,x')+\alpha C(a,a')$ for some $\alpha \in [0,1]$. Meanwhile, assume the metric on $(\Omega_x \times \Omega_a \times \Omega_y, d_{(x,a,y)})$ is the Manhattan distance along the input and output coordinates, i.e., $\Omega_x \times \Omega_a$ and $\Omega_y$. 

\paragraph{Assumption 3} (Bounded model complexity) Given a graph domain where the number of training samples is $n$ and the maximum degree is $d_{max}$, assume the Rademacher complexity of the considered GNN model family $\Phi$ is bounded by \cite{garg2020generalization}:
\begin{align}
\mathfrak{R}_n(\Phi) \leq \mathcal{O}\Big(\frac{d_{max}}{\sqrt{n}}\Big).
\end{align}



\paragraph{Assumption 4} (Lipschitz model) Assume for any $\theta \in \Theta$, the model $M_\theta$ is $R$-Lipschitz in the implicit input space ($\Omega_x \times \Omega_a, d_{(x,a)}$) for some $R > 0$, and the model $\Psi_\theta$ is $\tilde{R}$-Lipschitz in the Euclidean embedding space ($\Omega_z, d_z$) for some $\tilde{R} > 0$:
\begin{align}
\begin{split}
|M_\theta((x,a))-M_\theta((x',a'))| \leq R \cdot d_{(x,a)}((x,a), (x',a')),& 
 \qquad \forall (x,a), (x',a') \in \Omega_x \times \Omega_a;\\
|\Psi_\theta(z)-\Psi_\theta(z')| \leq \tilde{R} \cdot d_z(z, z'),& \qquad \forall z, z' \in \Omega_z.
\end{split}
\end{align}

\paragraph{Assumption 5} (Lipschitz loss) Assume the loss function $\ell$ is $\rho$-Lipschitz in the output space $\Omega_y$ for some $\rho > 0$ \cite{wang2022understanding}:
\begin{align}
\begin{split}
|\ell(y, \cdot) - \ell(y', \cdot)| &\leq \rho | y-y' |,\\
|\ell(\cdot, y) - \ell(\cdot, y')| &\leq \rho | y-y' |, \qquad \forall y, y' \in \Omega_y.
\end{split}
\end{align}

\section{Analysis of GGDA Domain Generation}\label{ggda_analysis}
Our goal is to transfer knowledge between the relevant source and target graphs under a large distribution shift. In the absence of naturally occurring intermediate data, we aim to generate a sequence of intermediate domains that promotes stable long-range knowledge transfer. Formal proofs of the generalization bound and all propositions in this paper are provided in the appendix.

Our primary consideration is to establish the criteria that intermediate domains should satisfy to guarantee a well-bounded target error for GGDA. Following previous GDA work \cite{wang2022understanding}, we adopt an online learning perspective and obtain the following generalization bound for graph-based gradual domain adaptation: 
\begin{gather}\label{general_bound}
\begin{split}
&\epsilon_T(\theta_T) \leq \epsilon_0(\theta_0) + \mathcal{O}\bigg(T\Delta+T\Big(e + \sqrt{\frac{log(1/\delta')}{b}}\Big)\bigg)+\tilde{\mathcal{O}}\Big(\frac{1}{\sqrt{nT}}\Big),\\
&\text{where}~~~~ \Delta = \frac{1}{T}\sum_{t=0}^{T-1} W_{p}(\mu_t,\mu_{t+1})~~ \text{and} ~~ \mu_t = \sum_{i=1}^{n_t} h^t_i \delta_{(x^t_i,a^t_i,y^t_i)}.
\end{split}
\end{gather}
Here $n$ indicates the average number of vertices in each domain. 
The terms $e$ and $\sqrt{\frac{log(1/\delta')}{b}}$ quantify the inherent non-IID nature of the given dataset and are beyond the scope of our optimization.
Our analysis focuses on the critical term $\Delta$, defined as the average $p$-Wasserstein distance (i.e., $W_p$) between consecutive domains. This metric captures the combined disparity in features, structures, and labels across domains under an optimal coupling.
We observe an inverse relationship between the optimal $\Delta$ and $T$ \cite{wang2022understanding}, suggesting that when the average distance between consecutive domains is large, the number of intermediate domains should be fewer, and vice versa. Note that this does not imply that the bound is tightest at $T=1$ (i.e., non-GDA). In fact, both Eq. \eqref{general_bound} and past empirical GDA studies \cite{kumar2020understanding, wang2022understanding} indicate that optimal performance is typically achieved at a "sweet spot" when $T$ is moderately large (i.e., $\Delta$ is moderately small). This further underscores the necessity of a graph GDA algorithm that strategically decomposes a large shift into a sequence of manageable, smaller transitions to enhance overall adaptation performance.

Based on this analysis, we identify two criteria for successful intermediate domain generation in GGDA: (1) the length of the path connecting source and target, i.e., $T\Delta$, should be small to curb information loss during transfer, and (2) the search for the optimal $\Delta$ should be conducted under an inverse relationship between $\Delta$ and T. Fulfilling these criteria presents three key challenges:
\begin{enumerate}
    \item Both criteria require generating domains based on the inter-domain distance $W_{p}(\mu_t,\mu_{t+1})$ over $\Omega_x \times \Omega_a \times \Omega_y$. This is highly non-trivial since (i) $W_{p}(\mu_t,\mu_{t+1})$ over the structural space $\Omega_a$ is inherently implicit, and (ii) during generation, each created sample (i.e.,  vertex) within domain $\mu$ must be characterized by both its own attributes and its relational information with neighbors.
    \item For criterion (1), an ideal solution is to generate a Wasserstein geodesic over $\Omega_x \times \Omega_a \times \Omega_y$, yet this is extremely challenging in practice due to the intractable optimization for an exact Wasserstein geodesic of attributed topological data.
    \item While the $p$-Wasserstein metric is defined on $\Omega_x \times \Omega_a \times \Omega_y$, only the source domain is labeled. How the intermediate domains should be generated to effectively integrate class information (i.e., $y$-dimension) poses another key challenge.
\end{enumerate}

To tackle challenge (1), we propose to facilitate domain generation by concretizing the graph metric (especially over the implicit topology on $\Omega_a$) using the FGW distance, which acts as a bridge between the abstract distance $W_{p}$ and practical implementations. Regarding challenge (2), we reformulate the goal of finding a geodesic into a tractable objective: \emph{minimizing information loss during the GGDA process}. Specifically, we aim to approximate the shortest path between source and target by generating intermediate domains that minimize information loss upon model adaptations. This is achieved by a combination of FGW optimizations and a subsequent vertex selection process for domain constructions. Additionally, the regularized vertex selection addresses challenge (3) through enhanced pseudo-labeling for domain progression. In the following sections, we will provide the detailed explanations accordingly.

\section{Graph Generation over FGW Metric}
While we wish to generate intermediate domains based on the distance defined in Eq. \eqref{general_bound}, a primary challenge is the intractability of $W_{p}(\mu_t,\mu_{t+1})$ over the product space $\Omega_x \times \Omega_a \times \Omega_y$, which arises because the metric for measuring structural distance between samples from different graphs $\Omega_a$ is implicit  (i.e., no inter-graph edges).
This presents a dilemma specific to a non-IID context, where each sample is characterized not only by itself but also by its relationships with other attributed samples. 
To bridge this gap between theory and practice, we require a well-defined and tractable inter-graph distribution distance that jointly accounts for features and structures. To this end, we adopt the Fused Gromov-Wasserstein (FGW) distance, defined as follows.

\paragraph{Definition 1} (FGW distance) Let two attributed graphs be 
$\mu = \sum_{i=1}^n h_i \delta_{(x_i,a_i)}$ and $\nu = \sum_{j=1}^m h'_j \delta_{(x'_j,a'_j)}$. 
Let $d_x(i,j) = d_x(x_i, x'_j)$ be the distance between cross-network features, while $C(i,k) = C(a_i, a_k)$ and $C'(j,l) = C(a'_j, a'_l)$ are the respective structure matrix for $\mu$ and $\nu$. Given the set of all admissible couplings $\Pi(h,h')$ and $\alpha \in [0,1]$, the FGW distance in the discrete case is defined as \cite{vayer2019}
\begin{gather}
\begin{split}
&FGW_{\alpha,p,q}(\mu,\nu) = \Big( \min_{\pi \in \Pi(h,h')} E_{\alpha,p,q}(\pi)\Big)^{\frac{1}{p}},   \quad \text{where}  \\
&E_{\alpha,p,q}(\pi) = \sum_{i,j,k,l} \Big[(1-\alpha) d_x(i,j)^q  
+\alpha \big|C(i,k)-C'(j,l)\big|^q\Big]^p \pi_{i,j} \pi_{k,l}.
\end{split}
\end{gather}
$E(\pi)$ represents the cost of transporting mass between graphs under a node matching strategy $\pi$. This cost combines the expense of transporting features across single nodes and transporting structures across pairs of nodes. The FGW distance is the minimal cost given by the optimal coupling. Hence, the computation of FGW is built upon sample coupling that preserves both feature and relation similarities. Under this formulation, node $i$ is matched to node $j$ only if they exhibit similar features and their respective neighbors (i.e., $k, l$) also present a correspondence.


\subsection{Fast Entropy-Guided Graph Generation}
Having established a tractable graph metric, we now generate intermediate graphs to bridge the source and target. Note that these "intermediate graphs" form a data pool from which "intermediate domains" will later be constructed. We propose to generate a sequence of $K-1$ intermediate graphs $\{\tilde\mu_k\}_{k=1}^{K-1}$ by computing the weighted Fr\'echet mean \cite{vayer2020} over the FGW metric on $\Omega_x \times \Omega_a$ (without $\Omega_y$ as $\mu_T$ is unlabeled).
However, directly computing the Fr\'echet mean between the \emph{entire} source and target graphs is computationally expensive for large graphs.
To improve efficiency and preserve knowledge, we introduce a partition-based, entropy-guided graph generation strategy, where each intermediate graph $\tilde\mu_k$ is generated as a combination of smaller subgraphs $\{\tilde\mu_{k_i}\}_i$.

We first transform the source and target graphs into $P_S$ and $P_T$ partitions with METIS \cite{karypis1998fast}, an algorithm that minimizes edge cuts. The core idea is to generate each subgraph $\tilde\mu_{k_i}$ from a matched pair of source and target partitions, denoted by $\hat{\mu}^S_{m(P_t)}$ and $\hat{\mu}^T_{P_t}$, with indices $m(P_t) \in \{1, 2, ..., P_S\}$ and $P_t \in \{1, 2, ..., P_T\}$. The function $m(\cdot)$ represents an index matching between source and target partitions, which we aim to optimize to enhance preservation of not only attributed structural similarity but also class semantics during our fast graph generation. 

We begin with a warm-up phase to obtain a preliminary matching $\dot{m}(\cdot)$. Let $m(\cdot)$ be a random matching, $n^S_{m(P_t)}$ and $n^T_{P_t}$ be the size of the partition pair, $Y_{m(P_t),c}^S \in \{0,1\}^{n^S_{m(P_t)}}$ be an indicator of source nodes belonging to class $c\in \{-1,1\}$, and $\pi^*(m(P_t),P_t) \in [0,1]^{n^S_{m(P_t)} \times n^T_{P_t}}$ be the normalized FGW optimal coupling on $\Omega_x \times \Omega_a$ between the partition pair. Then, for each $P_t \in \{1, 2, ..., P_T\}$, we compute the following:
\begin{gather}\label{entropy}
\begin{split}
&f_{\big(m(P_t),P_t\big),c} = \Big[diag\big(Y_{m(P_t),c}^S\big) \cdot \pi^*\big(m(P_t),P_t\big)\Big]^T\cdot \mathbf{1},\\
&F_{\big(m(P_t),P_t\big)} = \Big[f_{\big(m(P_t),P_t\big),-1}, ~f_{\big(m(P_t),P_t\big),1}~\Big],\\
&H_{\big(m(P_t),P_t\big)} = \frac{1}{n^T_{P_t}}\sum_{i=1}^{n^T_{P_t}}\Big[-\sum_{j} F_{\big(m(P_t),P_t\big),ij}\cdot logF_{\big(m(P_t),P_t\big),ij}\Big].
\end{split}
\end{gather} 
Simply put, for a candidate matching pair $(m(P_t),P_t)$, we "push" source labels onto target nodes via the optimal node coupling $\pi^*$ between the partitions, obtaining for each target node an estimated probability distribution over classes, i.e., $F_{(m(P_t),P_t)} \in [0,1]^{n^T_{P_t} \times 2}$. The average of target node class entropy, $H_{(m(P_t),P_t)} \in \mathbb{R}$, serves as a useful indicator of information compatibility between the partition pair: a large $H$ indicates that much valid class information has been lost by coupling this pair over FGW. 
Then, to ensure matched partitions are both structurally similar and class-compatible, we combine the effects of source class entropy $H^S_{m(P_t)}\in \mathbb{R}$ (i.e., the overall class label entropy of the source partition) and the FGW distance. For each pair $(m(P_t),P_t)$, we define a measure of information loss as
\begin{gather}\label{map_score}
S_{loss}\big(m(P_t),P_t\big) = \Big(H_{\big(m(P_t),P_t\big)}\big/H^S_{m(P_t)}\Big) \cdot FGW\big(\hat{\mu}^S_{m(P_t)}, \hat{\mu}^T_{P_t}\big),
\end{gather} 
where the entropy ratio prevents collapse into matches that favor class-pure source partitions, ensuring that the matching respects class diversity across the transition. During the warm-up phase, different initializations of $m(\cdot) \in \mathcal{M}(\cdot)$ are employed, and a preliminary matching is set to $\dot{m}(P_t)=\argmin_{m(P_t)\in\mathcal{M}(P_t)}S_{loss}(m(P_t),P_t)$.

We now proceed to the knowledge-preserving graph generation using the matched partition pairs. In this phase, $m(P_t)$ is set to $\dot{m}(P_t)$ with a probability $\propto 1/S_{loss}(\dot{m}(P_t),P_t)$ and is set to a new matching otherwise. This strategy prioritizes low-loss matches while allowing for further optimization. Using this matching, we generate a sequence of unlabeled intermediate graphs $\{\tilde\mu_k\}_{k=1}^{K-1}$ over $\Omega_x \times \Omega_a$ via the following weighted Fr\'echet mean optimization for each $k$: 
\begin{gather}\label{barycenter}
\begin{split}
\tilde\mu_{k} = \frac{1}{\Gamma}\sum_{i}\tilde\mu_{k_i}&,
~~~~\text{where} \\\tilde\mu_{k_i} = \argmin_{\mu} \Big[ &\Big(\frac{K-k}{K}\Big) \cdot FGW\big(\mu,\hat{\mu}^S_{m(P_{t,i})}\big) +\Big( \frac{k}{K} \Big) \cdot FGW \big(\mu, \hat{\mu}^T_{P_{t,i}}\big)\Big]\text{,}
\end{split}
\end{gather} 
with $P_{t,i}\sim Uniform\{1,2,...,P_T\}$ and $\Gamma$ being a normalizing factor. This scheme constructs a bridging graph sequence $\{\tilde\mu_k\}_{k=1}^{K-1}$ by incrementally shifting the position of $\tilde\mu_k$ between $\mu_S$ and $\mu_T$ over the FGW metric. The index matching $\dot{m}(\cdot)$ is continuously refined when a new matching yields a smaller $S_{loss}$ than the previous one. In this phase, $S_{loss}(m(P_t),P_t)$ is computed by substituting (1) the source-intermediate optimal coupling $\pi^*(m(P_{t,i}), {k_i})$ in Eq. \eqref{entropy} and (2) the distance sum $FGW(\hat{\mu}^S_{m(P_{t,i})},\tilde\mu_{k_i}) + FGW (\tilde\mu_{k_i}, \hat{\mu}^T_{P_{t,i}})$ in Eq. \eqref{map_score}. Both quantities are obtained directly as byproducts of the Fr\'echet mean optimization, requiring no additional computation. Since the entropy now reflects information loss relative to the intermediate graph $\tilde\mu_{k_i}$, the information integrity of the generated graphs is actively enhanced. Overall, our accelerated module largely reduces the time complexity of graph generation from $\mathcal{O}(n^3)$ to $\mathcal{O}(Pn_p^3)$, where $P$ is the average number of partitions and $n_p$ is the average partition size (see complexity analysis).


\subsection{Analysis: Why Is FGW-Based Graph Generation Important for a Successful GGDA?}
While the FGW-generated graphs do not directly conform to $W_{p}(\mu_t,\mu_{t+1})$ over $\Omega_x \times \Omega_a \times \Omega_y$ in Eq. \eqref{general_bound}, this proposed generation process is central to our algorithm for two reasons.

First, the minimization in Eq. \eqref{barycenter} serves as a mechanism for information preservation in the construction of intermediate graphs by aiming to reduce the FGW distance between $\tilde\mu_{k_i}$ and $\hat{\mu}_S$/$\hat{\mu}_T$. The sample correspondence constraint inherent in FGW optimization forces the ego-graphs distributed in the generated $\tilde\mu_k$ to meaningfully interpolate and preserve the characteristics of source and target ego-graphs.
In other words, this approach serves to \emph{minimize the information loss} over $\Omega_x \times \Omega_a$ when generating the bridging data sequence.

Second, as discussed in Section \ref{ggda_analysis}, generating domains requires optimizing the intractable $W_{p}(\mu_t,\mu_{t+1})$ over $\Omega_x \times \Omega_a \times \Omega_y$.
While this exact $W_p$ distance is unattainable, it is, in fact, lower bounded by the FGW distance over $\Omega_x \times \Omega_a$, as formalized below.

\paragraph{Proposition 2} For graph measures $\mu(x,a,y) = \sum_{i=1}^n h_i \delta_{(x_i,a_i,y_i)}$ and $\nu(x',a',y') = \sum_{j=1}^m h'_j \delta_{(x'_j,a'_j,y'_j)}$, along with their marginal measures $\mu(x,a) = \sum_{i=1}^n h_i \delta_{(x_i,a_i)}$ and $\nu(x',a') = \sum_{j=1}^m h'_j \delta_{(x'_j,a'_j)}$,
\begin{gather}
W_{p}\big(\mu(x,a,y),\nu(x',a',y')\big) \geq FGW_{\alpha,p,1}\big(\mu(x,a),\nu(x',a')\big)/2.
\end{gather}
This relation implies that, to generate data with a certain $W_{p}$ distance, a necessary condition is to reduce the FGW distance between the data. Intuitively, our FGW-based graph generation is a crucial
first step for creating a bridging data pool that can subsequently be
refined to achieve the desired $W_{p}(\mu_t,\mu_{t+1})$ and, ultimately, a successful GGDA process.

\section{GGDA Domain Sequence Construction}
Gradual domain adaptation (GDA) involves model training on a sequence of $T$ domains with labeled data, such that the model shifts from source domain $\mu_0$ to target domain $\mu_T$ with small steps over the intermediate domains $\{\mu_t\}_{t=1}^{T-1}$. Specifically, since only the source domain contains ground-truth labels, each intermediate domain is defined with pseudo-labels obtained during iterative training. To perform GDA on graphs, one major consideration, as discussed above, is how to properly define the intermediate domain sequence over $\Omega_x \times \Omega_a \times \Omega_y$ to minimize information loss during the GGDA process. Specifically, it is crucial to ensure that pseudo-labeling can be performed accurately when defining each labeled intermediate domain under our algorithm, thereby curbing the accumulation of label error and preserving transferable knowledge in GGDA.

\subsection{Analysis: Vertex-Based Domain Construction for GGDA}
Given the FGW-generated graph sequence, one might define each \emph{intermediate domain} $\mu_t$ directly as a single \emph{intermediate graph} $\tilde\mu_{k\mid k=t}$ with post-assigned pseudo-labels. However, such a definition is suboptimal because the FGW distance itself is not an upper bound of $|\epsilon_{k+1}(\theta) - \epsilon_k(\theta)|$, i.e., it does not inherently guarantee information transferability (and thus the correctness of pseudo-labeling) from $\tilde\mu_k$ to $\tilde\mu_{k+1}$.
To this end, we propose to construct each domain $\mu_t$ progressively during training by selecting vertices from the generated data pool.
This strategy offers two key advantages, as described below.

\begin{figure}[t]
    \centering
    \includegraphics[width=0.55\textwidth, trim=0cm 0.5cm 0cm 3cm,clip]{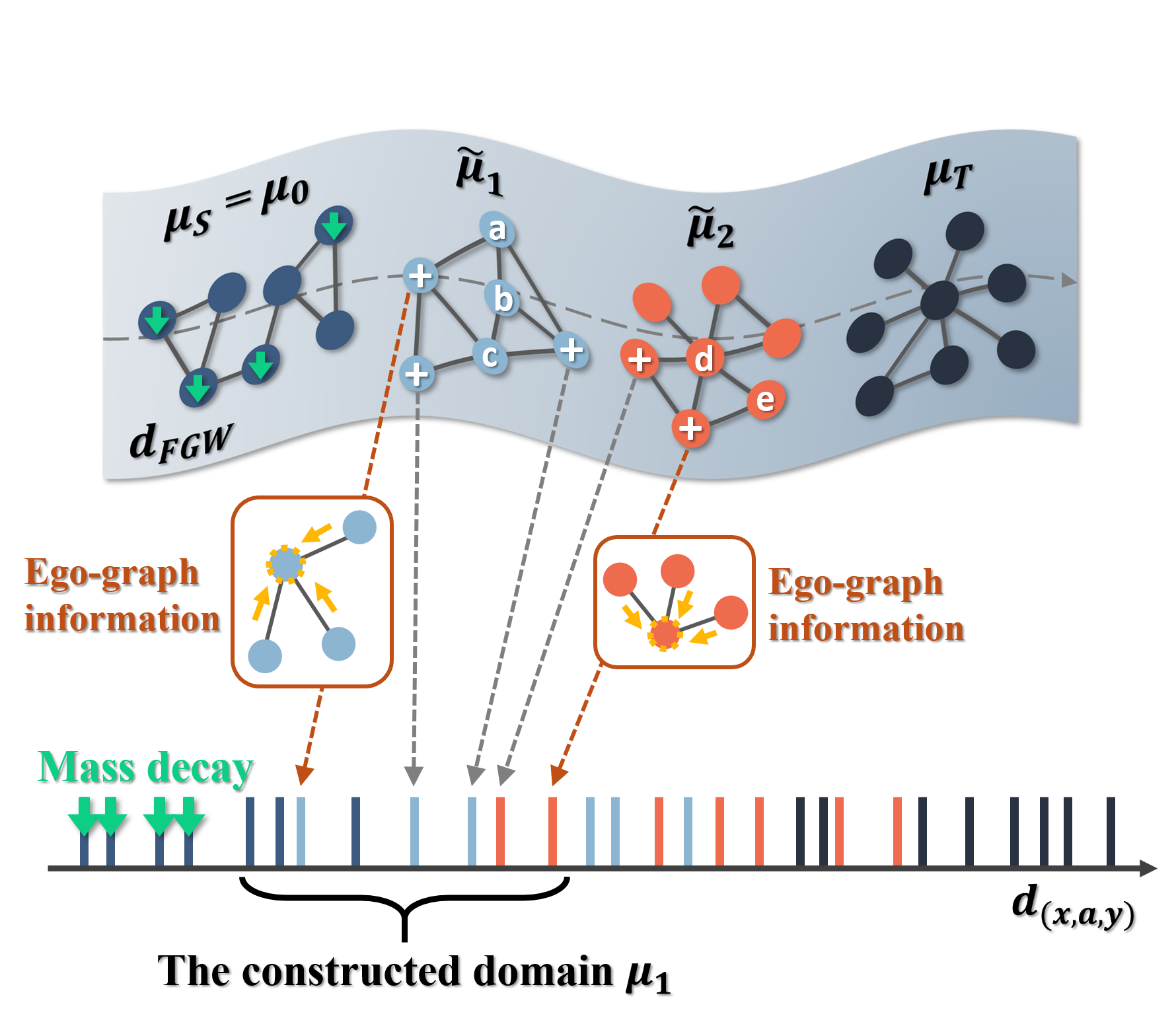} 
    \caption{Illustration of our GGDA framework. The upper part shows the sequence of intermediate graphs generated in the FGW metric space, and the lower part shows the corresponding vertex relations over product space distance $d_{(x, a, y)}$. $\mu_0$ (in dark blue) is initially labeled. For forming the subsequent domain $\mu_1$, vertices marked "+" are newly selected, while those with a green downward arrow are downweighted.
    }
    \Description[Illustration of the GGDA framework showing intermediate graph generation and vertex-based domain construction]{The figure illustrates the Graph Gradual Domain Adaptation (GGDA) framework in two distinct parts. The upper section displays a sequence of intermediate graphs generated within the Fused Gromov-Wasserstein (FGW) metric space, smoothly transitioning from the initial source domain $\mu_S=\mu_0$ (represented by dark blue nodes) to the final target domain $\mu_T$ (represented by black nodes), with intermediate graphs $\tilde{\mu}_1$ and $\tilde{\mu}_2$ situated in between. The lower section maps these graph vertices onto a horizontal axis representing the product space distance $d_{(x,a,y)}$. To construct the subsequent domain $\mu_1$, vertices from the generated intermediate graphs that are closest to the source (indicated by white plus signs) are newly selected based on their ego-graph information. Concurrently, obsolete source vertices (indicated by green downward arrows) undergo adaptive mass decay and are downweighted. This progressive, vertex-based selection strategy creates a compact constructed domain $\mu_1$ that effectively minimizes information loss during the adaptation process.}
    \label{main_graph}
\end{figure}

First, the ultimate objective of GGDA is vertex-based node classification, which depends on the characteristics encoded in each vertex's ego-graph. Therefore, instead of treating an entire FGW-generated graph as an individual domain, defining each subsequent domain as a set of vertices selected based on their knowledge relevance to the previous domain better preserves the transferable and discriminative knowledge during adaptation. As presented in Figure \ref{main_graph}, we start with the labeled domain $\mu_S = \mu_0$ in training. 
Since vertices in the graph sequence vary in their relevance to $\mu_0$ in terms of the $d_{(x, a, y)}$ metric (i.e., distance over $\Omega_x \times \Omega_a \times \Omega_y$),
selecting vertices that are close to $\mu_0$ (marked with "+") to form the subsequent domain $\mu_1$
can better enhance cross-domain knowledge transfer by minimizing information loss.

Second, since the newly constructed domain can incorporate vertices from different generated graphs, the domain's connections to unused vertices in the pool are naturally maintained through existing edges. 
Knowledge propagation across these edges improves the accuracy of pseudo-labeling on connected vertices, thereby elevating the quality of the next constructed domain.
Put differently, the presence of inter-domain edges is the key to a successful GGDA, as it indicates greater transfer potential and reduced information loss between consecutive domains. In Figure \ref{main_graph}, for example, if vertices marked with "+" are correctly pseudo-labeled and included in domain $\mu_1$, then vertices $a$ to $e$, being prospective components of domain $\mu_2$, can receive accurate pseudo-labels via message passing from their reliably labeled neighbors. This insight regarding the importance of inter-domain connectivity is further supported theoretically, as shown in the following proposition.

\paragraph{Proposition 3} Consider graph domains $\mu = \sum_{i=1}^n h_i \delta_{(x_i,a_i,y_i)}$ and $\nu = \sum_{j=1}^m h'_j \delta_{(x'_j,a'_j,y'_j)}$ with uniform weights $h$ and $h'$. Let $\mathcal{E}$ be the set of unit-weight edges linking $\mu$ and $\nu$ and $\tilde{\pi}$ be an arbitrary coupling satisfying $\tilde{\pi}_{i,j} = \frac{1}{nm} \mathds{1}_{\{(i,j) \in \mathcal{E}\}}$. For any $\theta \in \Theta$, the population loss on $\mu$ and $\nu$ given by $M_\theta$ satisfies
\begin{gather}
\begin{split}
&~~~~~~~|\epsilon_{\mu}(\theta) - \epsilon_{\nu}(\theta)| 
\leq \rho(\zeta_1 + \zeta_2), \qquad\text{where} \\ 
\zeta_1 =& \frac{\sqrt{\tilde{R}^2+1}}{n \cdot m} \sum_{(i,j)\in \mathcal{E}}\Big(\lVert z_i(\theta) - z'_j(\theta) \rVert + | y_i - y'_j |\Big), \\
\zeta_2 =& \sqrt{R^2+1} \cdot  \bigg[\min_{\tilde{\pi}}\bigg(\sum_{(i,j)\notin \mathcal{E}} d_{(x,a,y)}\Big((x_i,a_i,y_i), (x'_j,a'_j,y'_j)\Big)^p
\cdot \tilde{\pi}_{i,j}  \bigg)\bigg]^{\frac{1}{p}}.
\end{split}
\end{gather}
When transferring a model between arbitrary graph domains $\mu$ and $\nu$, the error bound consists of two components (i.e., $\zeta_1$ and $\zeta_2$) corresponding to node pairs \emph{with} and \emph{without} inter-domain edges, respectively. For $\zeta_1$, $\sum_{(i,j)\in \mathcal{E}}\lVert z_i(\theta) - z'_j(\theta) \rVert$ is small as most GNNs act as graph Laplacian regularizers that enforce similarity between the learned embeddings of two connected nodes \cite{zhu2021interpreting}. Meanwhile, $\sum_{(i,j)\in \mathcal{E}}| y_i - y'_j |$ is small if the graphs are homophilic. 
For $\zeta_2$, however, the distance $\sum_{(i,j)\notin \mathcal{E}}d_{(x,a,y)}^p$ can be unbounded.
We therefore expect a tighter bound when consecutive domains are well-connected (i.e., not mutually isolated), as this maximizes the contribution of the constrained $\zeta_1$ term and minimizes the influence of the potentially large $\zeta_2$ term.

\subsection{The Domain Progression Framework}\label{domain_progression}

Motivated by the preceding analysis, we propose to construct the domain sequence $\{\mu_t\}_{t=0}^{T}$ progressively in a vertex-based manner during training. Specifically, given the current domain $\mu_t$, we construct the subsequent domain $\mu_{t+1}$ with two novel modules: regularized vertex selection and adaptive mass decay. 

We treat the entire generated graph sequence together with the source and target graphs as a single batched graph $\mu_B$ over $\Omega_x \times \Omega_a$, with vertex set $\mathcal{V}_B:= supp(\mu_B) = \bigcup_{k=1}^{K-1} supp(\tilde\mu_k) \cup supp(\mu_S) \cup supp(\mu_T)$. Each vertex $v_b \in \mathcal{V}_B$ is associated with a mass decay mask initialized to $1$, i.e., $w^0_{v_b} = 1$.
Initially, only source vertices are labeled, so we define the first domain $\mu_0 = \mu_S = \sum_{i=1}^{n_S} h^S_i \delta_{(x^S_i,a^S_i, y^S_i)}$ with $h^S_i = 1/n_S$. For any model $M_{\theta}$, we denote the geometric margin by $|M_{\theta}(x,a)|$ and the prediction by $sign(M_{\theta}(x,a))$. In each iteration, we construct the subsequent domain $\mu_{t+1}$ from the current domain $\mu_t$ using the following two modules designed for knowledge-preserving transfer from $\mu_t$ to $\mu_{t+1}$.
\footnote{From here till the end of section \ref{domain_progression}, we denote $(x_{v_*}, a_{v_*}, \hat{y}_{v_*}, h_{v_*}, w_{v_*})$ by $(x_*, a_*, \hat{y}_*, h_*, w_*)$.}

\subsubsection{Vertex Selection for Constructing the Next Domain} Let $\mathcal{V}^t_l := supp(\mu_t)$ denote the set of vertices of the current domain $\mu_t$ and $\mathcal{V}^t_u := \mathcal{V}_B\symbol{92}(\bigcup_{t'=0}^{t}\mathcal{V}^{t'}_l)$ be the set of unused vertices in the data pool (not yet selected for constructing any prior domain).
First, we train a new model $\theta_t$ on the current domain $\mu_t$ with the loss $\mathbb{E}_{x,a,\hat{y} \sim \mu_t}\big[\ell(M_{\theta_t}(x,a), \hat{y})\big] = \sum_{v_l\in \mathcal{V}_l^t} h_l^t \cdot \ell\big(M_{\theta_t}(x_l,a_l), \hat{y_l}\big)$, where $\hat{y}$ is either a pseudo-label or a ground-truth label.
In the inference phase, for each unused vertex $v_u \in \mathcal{V}^t_u$, we compute a regularized score
\begin{gather}\label{vertex_select}
\begin{split}
c_u^t = \big|M_{\theta_t}(x_u,a_u)\big| \cdot exp\Big(-\frac{d^t_u}{\max_{\{v_{u'}\in \mathcal{V}^t_u\}}d^t_{u'}} \cdot \eta \Big),\\
\text{where} \quad d^t_u = \min_{v_l \in \mathcal{V}^t_l} d_z \big(\Phi_{\theta_t}(x_u,a_u), \Phi_{\theta_t}(x_l,a_l)\big).
\end{split}
\end{gather} 
Here $d_z$ is the Euclidean distance.
In this module, we first obtain the node embeddings for vertices in both the current domain $\mathcal{V}^t_l$ and the unused set $\mathcal{V}^t_u$ using the encoder $\Phi_{\theta_t}$, which aggregates feature and structure knowledge from each vertex's ego-graph (either given or FGW-generated).
An $\eta$-penalized scoring mechanism then prioritizes vertices in the unused set $\mathcal{V}^t_u$ with high geometric margin $|M_{\theta_t}(x_u,a_u)|$ and small embedding distance $d_u^t$ to the current domain $\mu_t$ --- vertices that are graph-proximal to current domain's labeled support nodes (via shared edges or similar ego-graph patterns) are more likely to be selected.
Vertices sorted by the score $c_u^t$ are selected subject to a constraint $\kappa$, the ratio of new pseudo-label count per class to source label count per class. The selected vertex set, denoted by $\mathcal{V}^t_{pl}$, is then incorporated into the next domain $\mu_{t+1}$, with pseudo-labels $\hat{y}_{pl} = sign(M_{\theta_t}(x_{pl}, a_{pl}))$ for $v_{pl}\in \mathcal{V}^t_{pl}$. 
We assign sharpened pseudo-labels (i.e., $-1$ or $1$) rather than probabilities to encourage more decisive model updates \cite{kumar2020understanding}. 
To justify the design of this module, we present the following proposition.

\paragraph{Proposition 4} Let $\Phi_{\tilde\theta}: \Omega_x \times \Omega_a \rightarrow \Omega_z$ be an $(L,\gamma)$-quasiisometric graph embedding \cite{dructu2018geometric}. Let $\mu(z|\tilde\theta) = \sum_{i=1}^n h_i \delta_{z_i|\tilde\theta}$, $\nu(z'|\tilde\theta) = \sum_{j=1}^m h'_j \delta_{z'_j|\tilde\theta}$, and $\pi^* = \argmin_{\pi \in \Pi(h,h')}\sum_{i,j} d_{z}(z_i|\tilde\theta, z'_j|\tilde\theta)^p \pi_{i,j}$. Then we have
\begin{align}
\begin{split}
&W_{p}\big(\mu(x,a,y),\nu(x',a',y')\big) \leq \xi
\text{~and~} 
|\epsilon_{\mu}(\theta) - \epsilon_{\nu}(\theta)| \cdot \mathcal{O}(\rho^{-1}) \leq \xi,\\ 
&\text{where } ~~~~~~\xi = L\Bigg[W_{\infty}\big(\mu(z|\tilde\theta),\nu(z'|\tilde\theta)\big) + \gamma  + \frac{1}{L}\bigg( \sum_{i,j} |y_i-y'_j|^p \pi_{i,j}^*\bigg)^{\frac{1}{p}}\Bigg].
\end{split}
\end{align}
Proposition 4 leads to two key conclusions that validate our approach: (1) Under the covariate shift assumption, our vertex selection strategy based on the sorted score $c^t_u$ adheres to $\xi$ formulation by prioritizing samples that are within the $\Omega_z$ neighborhood and distant from the decision boundary. Therefore, modifying the hyperparameter $\kappa$, which controls the scale of class-proportionate vertex selection under $c_u^t$-prioritization, enables flexible adjustment of inter-domain distance $W_{p}(\mu_t,\mu_{t+1})$ over $\Omega_x \times \Omega_a \times \Omega_y$ via its upper bound $\xi$ while satisfying an inverse relationship between $\Delta$ and $T$. This provides a tractable mechanism to realize an efficient global search for the optimal GGDA domain path without requiring any repetitive data generation. (2) Meanwhile, as our proposed prioritization favors inclusion of vertices close to $\mu_t$ over the metric $d_{(x,a,y)}$ in the next domain $\mu_{t+1}$, it ensures \emph{minimal information loss} by preserving the most transferable and discriminative knowledge from $\mu_t$ to $\mu_{t+1}$. Accordingly, a bounding effect on the local generalization error $|\epsilon_{t+1}(\theta) - \epsilon_{t}(\theta)|$ can be established to secure stable knowledge transfer upon the domain sequence.

\subsubsection{Adaptive Mass Decay of Vertices} 
Due to the stochasticity in selecting $\mathcal{V}^t_{pl}$ (newly pseudo-labeled vertices used to construct the next domain $\mu_{t+1}$), strictly excluding all current-domain vertices $\mathcal{V}_l^t$ from the subsequent domain support $\mathcal{V}_l^{t+1}$ may result in low inter-domain connectivity, degrading pseudo-label quality in the succeeding iteration. Moreover,
training a new model in each domain would require a large pool of generated data if every domain consisted entirely of new vertices, which is inefficient.
To this end, we propose an adaptive mass decay mechanism that allows a subset of current-domain vertices $\mathcal{V}_l^t$ to persist into the next domain --- but with decaying influence. Let $\mathcal{V}_{pll}^t  := \mathcal{V}_l^t \cup \mathcal{V}_{pl}^t$ be the union of the current domain support and the newly pseudo-labeled vertices. In every iteration, we store the label score $\hat{c}^t_{pll} = \hat{y}_{pll} \cdot M_{\theta_t}(x_{pll}, a_{pll})$ for $v_{pll} \in \mathcal{V}_{pll}^t$, which reflects how well the current model $M_{\theta_t}$'s predicted class aligns with the label $\hat{y}$. Then, if $t \geq 1$, we compute the mass decay for each current-domain vertex $v_l \in \mathcal{V}^t_l$ by comparing its current label score to its previously stored label score:
\begin{gather}\label{mass_decay}
\lambda_l^t = exp\bigg(-\bigg[1 - min\Big(\frac{\hat{y}_{l} \cdot M_{\theta_t}(x_{l}, a_{l})}{\hat{c}^{t-1}_{l}} , 1\Big)\bigg] \cdot \beta\bigg).
\end{gather}
The key idea is to identify the model flow after training in a new domain by quantifying the degradation in label correctness. Nodes with degraded labels will be downweighted in the next constructed domain with decay rate controlled by hyperparameter $\beta$, while reliable anchors are retained --- this effectively steers model advancement toward the target.
Specifically, we update the mask $w^{t}_l = w^{t-1}_l \cdot \lambda_l^t$ for $v_l \in \mathcal{V}^t_l$ cumulatively and finally define next domain distribution as
\begin{gather}\label{overall_dist}
\begin{split}
\mu_{t+1} &= \frac{1}{\Gamma}\bigg(\sum_{v_l \in \mathcal{V}^t_l} w^{t}_l \cdot \delta_{(x_l,a_l,\hat{y}_l)} +  \sum_{v_{pl}\in \mathcal{V}^t_{pl}}\delta_{(x_{pl}, a_{pl}, \hat{y}_{pl})}\bigg) \\
&= \sum_{v_{pll} \in \mathcal{V}^t_{pll}} h_{pll}^{t+1} \cdot \delta_{(x_{pll}, a_{pll}, \hat{y}_{pll})},  
\end{split}
\end{gather}
where $\Gamma$ is a normalizing factor ensuring $\sum h_{pll}^{t+1} = 1$. In practice, we can keep only the top-$k$ weighted samples in the new domain $\mu_{t+1}$ and assign a mass of $0$ to others, where $k$ is the average size of source and target graphs. The decay technique is conceptually illustrated in Figure \ref{main_graph}, with a minor distinction that this technique takes effect only when $t \geq 1$ (i.e., starting from the construction of $\mu_2$). Note that while some target vertices might be labeled prior to the final stage, mass decay is not applied to these vertices, as our goal is to adapt to the target distribution. The GGDA process concludes with a final training step once a sufficient number of target vertices are confidently labeled.

\section{A Unified View of GGDA \& Complexity Analysis}\label{apd_complexity}
Due to space limitations, the full algorithmic procedure of GGDA is provided in the appendix. At a high level, GGDA aims to construct a sequence of intermediate domains that minimizes information loss during adaptation from source to target. The central challenge in graphs is the intractability of the inter-domain Wasserstein distance $W_p(\mu_t, \mu_{t+1})$ over $\Omega_x \times \Omega_a \times \Omega_y$, due to the implicit nature of inter-graph structural distances (i.e., the absence of inter-graph edges). 

To bridge theory and practice, we bound $W_p(\mu_t, \mu_{t+1})$ from below via the FGW distance and from above via $\xi$ in Proposition 4 (a term related to $\Omega_z$ neighborhood and decision boundary). Guided by these bounds, GGDA proceeds in two stages: FGW-based intermediate graph generation creates a bridge using a fast entropy-guided strategy, while vertex-based domain progression iteratively constructs compact domains through regularized selection and adaptive mass decay. Together, these mechanisms enforce the theoretical bounds while actively prioritizing pseudo-label consistency, curbing information loss across domains. Importantly, GGDA’s hyperparameters serve as tunable knobs to control inter-domain compactness $W_p(\mu_t, \mu_{t+1})$, allowing practitioners to balance domain count $T$ and step size $\Delta$ --- in line with the inverse relationship in our theoretical analysis --- without costly data regeneration. In essence, GGDA transforms an otherwise intractable distribution-matching problem into a modular, bound-driven pipeline for long-range transfer.
For the generation of intermediate graphs, let $P$ be the average number of partitions and ${n}_p \ll \max(n_S, n_T)$ be the average partition size. Then, the METIS partitioning takes $\mathcal{O}(P{n}_p+ (P{n}_p)^2 +Plog(P)) = \mathcal{O}((P{n}_p)^2)$, the warmup phase takes $\mathcal{O}(P{n}_p^3)$, and the weighted Fr\'echet mean generation takes $\mathcal{O}(KPI{n}_p^3)$, with $I$ denoting the number of iterations for block coordinate descent (BCD) \cite{vayer2020}. Without loss of generality, the overall complexity for the generation phase in GGDA is $\mathcal{O}(KPI{n}_p^3)$. For comparison, generating intermediate graphs without partitioning would escalate the complexity to $\mathcal{O}(KIn^3)$, where $n$ is the average size of source and target graphs, making the computation prohibitively expensive for large networks.


For the domain progression framework, the time complexity of model training in each domain is $\mathcal{O}(h|A_t^+|d^2)$, where $h$ is the number of hops for GNN aggregations, $d$ is the latent dimension, and $|A_t^+|$ is the number of edges in the subgraph induced by the set of labeled vertices in the current domain (i.e., $\mathcal{V}_l^t:=supp(\mu_t)$) and their $h$-hop neighborhoods. The time complexity of each subsequent domain construction is $\mathcal{O}(|\mathcal{V}_l^t||\mathcal{V}_u^t|d^2)$, where $|\mathcal{V}_u^t|$ is the number of samples in the data pool that have not been labeled before.

\section{Experiments}
We examine whether GGDA improves graph knowledge transfer via its constructed domain sequence. We focus on cross-network node classification, where source labels are used for training and target labels are used for validation and testing under a 2:8 ratio \cite{liu2023structural}. 
We conduct transfer experiments on fifteen graph datasets. 
Additional experiments, implementation details, and elaborations of datasets and baseline methods are provided in the appendix.

\subsection{Datasets}
Table \ref{table_stat} shows the statistics of the datasets. For the following groups, we use one as source and the other as target:
\begin{itemize}
    \item \textbf{ACM} (pre-2008), \textbf{Citation} (post-2010), and \textbf{DBLP} (2004 to 2008)  (denoted by \textbf{A}, \textbf{C}, \textbf{D}): multi-label citation networks extracted from ArnetMiner \cite{shen2020adversarial};
    \item \textbf{ACM2} (2000 to 2010) and \textbf{DBLP2} (post-2010),  (denoted by \textbf{A2}, \textbf{D2}): another pre-processed version of single-label citation networks \cite{wu2020unsupervised};
    \item \textbf{Blog1} and \textbf{Blog2} (denoted by \textbf{B1}, \textbf{B2}): social networks from BlogCatalog, with 30\% of the binary attributes randomly flipped to enlarge discrepancy \cite{li2015unsupervised};
    \item \textbf{Arxiv 2007, 2016, 2018}: Arxiv computer science citation networks partitioned by time (from year 1950) \cite{hu2020open}.
\end{itemize}

As for citation networks \textbf{Cora} and \textbf{CiteSeer} \cite{sen2008collective}, as well as Twitch social networks \textbf{ENGB, DE}, and \textbf{PTBR} \cite{rozemberczki2021multi}, we simulate a \emph{multi-step shifting process} independently on each graph to mimic gradual data shifts. 
Specifically, we perform 5 steps of class-wise feature shifts generated with Gaussian noise.
This effectively shifts the distribution in $\Omega_x \times \Omega_a$ since both $p(x)$ and $p(a|x)$ have changed accordingly. These datasets with multi-step shifting will be used for evaluation in different ways, as detailed later.

\begin{table}[t]
    \footnotesize
    \centering
    \caption{Statistics of datasets.}
    \setlength{\tabcolsep}{3pt}
    \begin{tabular}{cccccc}
        \toprule
        \textbf{Graph}
                    &\textbf{\# Nodes}
                    &\textbf{\# Edges}
                    &\textbf{Density}
                    &\textbf{\# Features}
                    &\textbf{\# Classes}\\
        \midrule
        \textbf{ACM} & {9360}
                        & {15602}
                        & {0.0004}
                        & {6775}
                        & {5}\\
        \textbf{Citation}& {8935}
                        & {15113}
                        & {0.0004}
                        & {6775}
                        & {5}\\
        \textbf{DBLP}& {5484}
                        & {8130}
                        & {0.0005}
                        & {6775}
                        & {5}\\
        \textbf{ACM2} & {7410}
                        & {11023}
                        & {0.0004}
                        & {7537}
                        & {6}\\
        \textbf{DBLP2}& {5578}
                        & {7290}
                        & {0.0005}
                        & {7537}
                        & {6}\\
        \textbf{Blog1} & {2300}
                        & {33471}
                        & {0.0127}
                        & {8189}
                        & {6}\\
        \textbf{Blog2} & {2896}
                        & {53836}
                        & {0.0128}
                        & {8189}
                        & {6}\\
        \textbf{Arxiv 2007} & {4980}
                        & {5849}
                        & {0.0005}
                        & {128}
                        & {40}\\
        \textbf{Arxiv 2016} & {69499}
                        & {232419}
                        & {0.0001}
                        & {128}
                        & {40}\\
        \textbf{Arxiv 2018} & {120740}
                        & {615415}
                        & {0.0001}
                        & {128}
                        & {40}\\
        \textbf{Cora} & {2708}
                        & {5278}
                        & {0.0014}
                        & {1433}
                        & {7}\\
        \textbf{CiteSeer}& {3327}
                        & {4552}
                        & {0.0008}
                        & {3703}
                        & {6}\\
        \textbf{ENGB}& {7126}
                        & {35324}
                        & {0.0014}
                        & {3170}
                        & {2}\\
        \textbf{DE}& {9498}
                & {153138}
                & {0.0034}
                & {3170}
                & {2}\\
        \textbf{PTBR}& {1912}
                & {31299}
                & {0.0171}
                & {3170}
                & {2}\\
        \bottomrule
    \end{tabular}
    \label{table_stat}
\end{table}

\subsection{Baselines}
We compare our GGDA framework with the following baselines: \textbf{CDAN} \cite{long2018conditional}, \textbf{MDD} \cite{zhang2019bridging}, \textbf{UDA-GCN} \cite{wu2020unsupervised}, \textbf{ACDNE} \cite{shen2020adversarial}, \textbf{ASN} \cite{zhang2021adversarial}, \textbf{GRADE} \cite{wu2023non}, \textbf{StruRW} \cite{liu2023structural}, \textbf{SpecReg} \cite{you2023graph}, \textbf{GraphAlign} \cite{huang2024can}, \textbf{A2GNN} \cite{liu2024rethinking}, \textbf{Pair-Align} \cite{liu2024pairwise}, and \textbf{TDSS} \cite{chen2025smoothness}. Specifically, CDAN and MDD are non-graph DA methods, for which we replace the encoder with GCN \cite{kipf2016semi}. All other baselines are graph DA methods under the \textbf{domain-invariant representation (DIR)} framework, except for GraphAlign, which is the first data-centric graph DA method introduced recently. While multiple baselines incorporate advanced encoder designs to enhance transfer performance, for our \textbf{GGDA} framework, we focus on three fundamental encoders in this paper: \textbf{GCN}, \textbf{GAT}, and \textbf{GraphSAGE} \cite{kipf2016semi, velivckovic2017graph, hamilton2017inductive}. Additionally, we include two baselines trained with ground-truth information. \textbf{GGDA-GT} performs GGDA on the ground-truth intermediate graphs from the \emph{multi-step shifting process}, while \textbf{Oracle} is trained on a labeled target graph.


\subsection{Main Results} Table \ref{F1_DIR}, Table \ref{acc_DIR} and Figure \ref{lineplot_DIR} present the performance of graph domain adaptations, with F1 scores for multi-label cases and accuracy scores for single-label cases. $\mathbf{\Delta_{Mean}}$ in the tables indicates how a model's performance compares to the mean performance across all models. For the \emph{multi-step shifting} datasets, Table \ref{acc_DIR} shows the case where only the graph from the last step of shift generation is used as the target, while Figure \ref{lineplot_DIR} shows the case where graphs from each step of shift generation are treated as target graphs with different discrepancies from the source graph.

As shown in the tables, our proposed GGDA framework consistently outperforms other baselines given the intermediate domain construction that aims to minimize information loss upon adaptations
and flexibly handles transfer scenarios with varied discrepancies. 
Moreover, the constructed domains effectively simulate the underlying data flow that bridges the gap, as evidenced by the small difference in target accuracy between our GGDA framework and GGDA-GT, which uses the actual intermediate data from the multi-step shifting process. Meanwhile, Figure \ref{lineplot_DIR} illustrates that GGDA-GCN and GGDA-GAT are the least susceptible to an increasing source-target discrepancy, and the high vulnerability of DIR-based models suggests that enforcing an alignment on domain distributions can lead to information distortion and backfire when the domain discrepancy is large. Such a limitation of DIR has also been theoretically validated in previous studies \cite{zhao2019learning}.

GGDA's adaptation performance varies with different encoders, governed by their inductive bias in steering domain progression. Specifically, GGDA-GCN favors exploitation --- its smoothing and spectral properties drive conservative evolution of the sequence into nodes embedded in strong homophilous neighborhoods while minimizing the influence of spurious patterns. As for GGDA-GAT, it favors exploration --- its attention mechanism encodes nodes based on salient local patterns, enabling more aggressive progression but risking distraction by outlier signals, a trade-off reflected in its strong performance on citation networks with stable patterns versus its overfitting to social networks with complex structures (e.g., Blogs). GGDA-SAGE underperforms relative to other variants in most scenarios due to its confined neighborhood sampling that dilutes the strong, unique structural signals of individual nodes. This is particularly detrimental in GGDA, where critical adaptation knowledge is embedded within the rich structures of generated graphs. Consequently, GGDA-SAGE can make erratic decisions in both vertex selection and mass decay processes. Please see the appendix for further analysis of encoder selection for GGDA.

\begin{table*}[h]
    \scriptsize
    \centering
    \setlength{\tabcolsep}{0.5pt}
    \caption{Performance of node classifications under the unsupervised graph DA setting (from a fully labeled source graph to an unlabeled target graph). Results are reported using micro-F1 (F1MI) and macro-F1 (F1MA) scores for the multi-label datasets below. $\dag$ indicates $p < 0.05$ and $\ddag$ indicates $p < 0.01$ using paired t-tests against the strongest baseline.}
    \begin{tabular}{ccccccccccccc|c}
        \toprule   
        \textbf{} &
        \multicolumn{2}{c}{\textbf{A to C}}  
        & \multicolumn{2}{c}{\textbf{A to D}}
        & \multicolumn{2}{c}{\textbf{C to A}}
        & \multicolumn{2}{c}{\textbf{C to D}}
        & \multicolumn{2}{c}{\textbf{D to A}}
        & \multicolumn{2}{c}{\textbf{D to C}}
        & \multicolumn{1}{c}{\multirow{2}{*}{$\mathbf{\Delta_{Mean}}$}}\\
        \cmidrule(r){2-3} \cmidrule(r){4-5}
        \cmidrule(r){6-7} \cmidrule(r){8-9}
        \cmidrule(r){10-11} \cmidrule(r){12-13}
        \textbf{} & \textbf{F1MI}
                    &\textbf{F1MA}
                    &\textbf{F1MI}
                    &\textbf{F1MA}
                    &\textbf{F1MI}
                    &\textbf{F1MA}
                    &\textbf{F1MI}
                    &\textbf{F1MA}
                    &\textbf{F1MI}
                    &\textbf{F1MA}
                    &\textbf{F1MI}
                    &\textbf{F1MA}\\
        \midrule
        \textbf{CDAN}   & $77.4 $\scalebox{0.7}{$\pm  0.3$}
                        & $74.7 $\scalebox{0.7}{$\pm  0.5$}
                        & $69.6 $\scalebox{0.7}{$\pm  0.5$}
                        & $65.3 $\scalebox{0.7}{$\pm  1.5$}
                        & $73.7 $\scalebox{0.7}{$\pm  0.3$}
                        & $72.5 $\scalebox{0.7}{$\pm  0.5$}
                        & $73.9 $\scalebox{0.7}{$\pm  0.3$}
                        & $70.9 $\scalebox{0.7}{$\pm  0.6$}
                        & $69.0 $\scalebox{0.7}{$\pm  0.4$}
                        & $66.3 $\scalebox{0.7}{$\pm  0.7$}
                        & $74.3 $\scalebox{0.7}{$\pm  0.3$}
                        & $70.8 $\scalebox{0.7}{$\pm  0.6$}
                        & $+0.3$\\
        \textbf{MDD}    & $76.4 $\scalebox{0.7}{$\pm  0.4$}
                        & $72.9 $\scalebox{0.7}{$\pm  0.7$}
                        & $67.6 $\scalebox{0.7}{$\pm  0.6$}
                        & $61.2 $\scalebox{0.7}{$\pm  1.6$}
                        & $74.5 $\scalebox{0.7}{$\pm  0.5$}
                        & $73.4 $\scalebox{0.7}{$\pm  0.7$}
                        & $73.2 $\scalebox{0.7}{$\pm  0.4$}
                        & $70.2 $\scalebox{0.7}{$\pm  0.5$}
                        & $69.8 $\scalebox{0.7}{$\pm  0.4$}
                        & $66.6 $\scalebox{0.7}{$\pm  0.8$}
                        & $74.5 $\scalebox{0.7}{$\pm  0.5$}
                        & $70.3 $\scalebox{0.7}{$\pm  0.9$}
                        & $-0.4$\\
        \midrule
        \textbf{UDA-GCN} & $74.0 $\scalebox{0.7}{$\pm  0.7$}
                        & $71.2 $\scalebox{0.7}{$\pm  1.5$}
                        & $69.0 $\scalebox{0.7}{$\pm  0.8$}
                        & $61.3 $\scalebox{0.7}{$\pm  2.4$}
                        & $72.6 $\scalebox{0.7}{$\pm  0.5$}
                        & $71.4 $\scalebox{0.7}{$\pm  0.6$}
                        & $71.5 $\scalebox{0.7}{$\pm  0.6$}
                        & $66.0 $\scalebox{0.7}{$\pm  2.7$}
                        & $70.4 $\scalebox{0.7}{$\pm  0.7$}
                        & $67.7 $\scalebox{0.7}{$\pm  2.3$}
                        & $74.5 $\scalebox{0.7}{$\pm  0.4$}
                        & $71.0 $\scalebox{0.7}{$\pm  1.1$}
                        & $-1.2$\\
        \textbf{ACDNE} & $ 78.9$\scalebox{0.7}{$\pm  0.3$}
                        & $ 77.3$\scalebox{0.7}{$\pm  0.3$}
                        & $ 72.2$\scalebox{0.7}{$\pm  0.3$}
                        & $ 69.9$\scalebox{0.7}{$\pm  0.3$}
                        & $ \underline{76.3}$\scalebox{0.7}{${\pm  0.1}$}
                        & $ 74.6$\scalebox{0.7}{$\pm  0.1$}
                        & $ 74.3$\scalebox{0.7}{$\pm  0.3$}
                        & $ 72.9$\scalebox{0.7}{$\pm  0.3$}
                        & $ 71.6$\scalebox{0.7}{$\pm  0.4$}
                        & $ 70.5$\scalebox{0.7}{$\pm  0.4$}
                        & $ 79.2$\scalebox{0.7}{$\pm  0.2$}
                        & $ 77.1$\scalebox{0.7}{$\pm  0.3$}
                        & $+3.3$\\
        \textbf{ASN} & $ 78.6$\scalebox{0.7}{$\pm  0.7$}
                        & $ 73.6$\scalebox{0.7}{$\pm  1.9$}
                        & $ 72.7$\scalebox{0.7}{$\pm  1.1$}
                        & $ 66.7$\scalebox{0.7}{$\pm  2.8$}
                        & $ 72.8$\scalebox{0.7}{$\pm  0.8$}
                        & $ 72.7$\scalebox{0.7}{$\pm  1.4$}
                        & $ 74.5$\scalebox{0.7}{$\pm  0.7$}
                        & $ 72.0$\scalebox{0.7}{$\pm  1.2$}
                        & $ 69.0$\scalebox{0.7}{$\pm  1.2$}
                        & $ 68.5$\scalebox{0.7}{$\pm  3.0$}
                        & $ 78.1$\scalebox{0.7}{$\pm  0.3$}
                        & $ 74.1$\scalebox{0.7}{$\pm  1.2$}
                        & $+1.5$\\
        \textbf{GRADE}& $ 70.8$\scalebox{0.7}{$\pm  0.6$}
                        & $ 68.4$\scalebox{0.7}{$\pm  0.8$}
                        & $ 66.4$\scalebox{0.7}{$\pm  0.4$}
                        & $ 61.6$\scalebox{0.7}{$\pm  2.1$}
                        & $ 67.9$\scalebox{0.7}{$\pm  0.3$}
                        & $ 66.3$\scalebox{0.7}{$\pm  0.4$}
                        & $ 69.6$\scalebox{0.7}{$\pm  0.5$}
                        & $ 66.3$\scalebox{0.7}{$\pm  1.0$}
                        & $ 65.0$\scalebox{0.7}{$\pm  0.7$}
                        & $ 62.1$\scalebox{0.7}{$\pm  1.0$}
                        & $ 68.3$\scalebox{0.7}{$\pm  0.5$}
                        & $ 65.2$\scalebox{0.7}{$\pm  1.2$}
                        & $-4.8$\\
        \textbf{StruRW} & $ 65.1$\scalebox{0.7}{$\pm  2.4$}
                        & $ 63.2$\scalebox{0.7}{$\pm  1.8$}
                        & $ 68.0$\scalebox{0.7}{$\pm  2.1$}
                        & $ 62.4$\scalebox{0.7}{$\pm  2.4$}
                        & $ 70.6$\scalebox{0.7}{$\pm  2.4$}
                        & $ 66.3$\scalebox{0.7}{$\pm  2.4$}
                        & $ 71.8$\scalebox{0.7}{$\pm  2.0$}
                        & $ 70.4$\scalebox{0.7}{$\pm  1.6$}
                        & $ 65.6$\scalebox{0.7}{$\pm  2.1$}
                        & $ 61.7$\scalebox{0.7}{$\pm  1.5$}
                        & $ 69.2$\scalebox{0.7}{$\pm  2.0$}
                        & $ 60.7$\scalebox{0.7}{$\pm  2.2$}
                        & $-5.0$\\
        \textbf{SpecReg}& $ 67.5$\scalebox{0.7}{$\pm  1.6$}
                        & $ 64.3$\scalebox{0.7}{$\pm  2.3$}
                        & $ 66.4$\scalebox{0.7}{$\pm  1.4$}
                        & $ 61.8$\scalebox{0.7}{$\pm  2.3$}
                        & $ 68.7$\scalebox{0.7}{$\pm  1.4$}
                        & $ 65.2$\scalebox{0.7}{$\pm  3.1$}
                        & $ 71.8$\scalebox{0.7}{$\pm  1.0$}
                        & $ 65.9$\scalebox{0.7}{$\pm  2.1$}
                        & $ 65.9$\scalebox{0.7}{$\pm  1.5$}
                        & $ 58.8$\scalebox{0.7}{$\pm  3.8$}
                        & $ 67.0$\scalebox{0.7}{$\pm  1.8$}
                        & $ 58.9$\scalebox{0.7}{$\pm  1.7$}
                        & $-6.1$\\
        \textbf{GraphAlign} & $ 73.2$\scalebox{0.7}{$\pm  1.1$}
                            & $ 70.1$\scalebox{0.7}{$\pm  1.1$}
                            & $ 67.5$\scalebox{0.7}{$\pm  1.4$}
                            & $ 64.6$\scalebox{0.7}{$\pm  1.2$}
                            & $ 67.2$\scalebox{0.7}{$\pm  1.1$}
                            & $ 66.6$\scalebox{0.7}{$\pm  1.1$}
                            & $ 71.6$\scalebox{0.7}{$\pm  1.1$}
                            & $ 68.2$\scalebox{0.7}{$\pm  1.8$}
                            & $ 59.6$\scalebox{0.7}{$\pm  1.6$}
                            & $ 58.8$\scalebox{0.7}{$\pm  1.0$}
                            & $ 66.0$\scalebox{0.7}{$\pm  0.8$}
                            & $ 62.9$\scalebox{0.7}{$\pm  1.7$}
                            & $-4.9$\\
        \textbf{A2GNN}& $ 75.5$\scalebox{0.7}{$\pm  0.3$}
                        & $ 77.4$\scalebox{0.7}{$\pm  0.3$}
                        & $ 62.1$\scalebox{0.7}{$\pm  0.4$}
                        & $ 66.3$\scalebox{0.7}{$\pm  0.3$}
                        & $ \underline{76.3}$\scalebox{0.7}{${\pm  0.2}$}
                        & $ 74.9$\scalebox{0.7}{$\pm  0.1$}
                        & $ 70.5$\scalebox{0.7}{$\pm  0.4$}
                        & $ 73.5$\scalebox{0.7}{$\pm  0.3$}
                        & $ 71.7$\scalebox{0.7}{$\pm  0.3$}
                        & $ 71.5$\scalebox{0.7}{$\pm  0.2$}
                        & $ 76.5$\scalebox{0.7}{$\pm  0.2$}
                        & $ 77.8$\scalebox{0.7}{$\pm  0.1$}
                        & $+1.6$\\
         \textbf{Pair-Align}& $ 72.4$\scalebox{0.7}{$\pm  0.6$}
                        & $ 68.3$\scalebox{0.7}{$\pm  0.6$}
                        & $ 67.5$\scalebox{0.7}{$\pm  0.5$}
                        & $ 64.3$\scalebox{0.7}{$\pm  0.3$}
                        & $ 71.2$\scalebox{0.7}{$\pm  0.7$}
                        & $ 68.5$\scalebox{0.7}{$\pm  0.3$}
                        & $ 71.3$\scalebox{0.7}{$\pm  0.3$}
                        & $ 70.4$\scalebox{0.7}{$\pm  0.2$}
                        & $ 66.1$\scalebox{0.7}{$\pm  0.4$}
                        & $ 62.9$\scalebox{0.7}{$\pm  0.6$}
                        & $ 72.4$\scalebox{0.7}{$\pm  0.5$}
                        & $ 67.2$\scalebox{0.7}{$\pm  0.4$}        & $-2.7$\\
        \textbf{TDSS}& $ 79.6$\scalebox{0.7}{$\pm  0.6$}
                        & $ 77.4$\scalebox{0.7}{$\pm  0.9$}
                        & $ 74.1$\scalebox{0.7}{$\pm  0.5$}
                        & $ 69.7$\scalebox{0.7}{$\pm  0.5$}
                        & $ 74.3$\scalebox{0.7}{$\pm  0.2$}
                        & $ 75.3$\scalebox{0.7}{$\pm  0.3$}
                        & $ 75.2$\scalebox{0.7}{$\pm  0.3$}
                        & $ 73.2$\scalebox{0.7}{$\pm  0.6$}
                        & $ 71.3$\scalebox{0.7}{$\pm  0.8$}
                        & $ 71.6$\scalebox{0.7}{$\pm  0.7$}
                        & $ 80.2$\scalebox{0.7}{$\pm  0.3$}
                        & $ 77.4$\scalebox{0.7}{$\pm  0.9$} 
                        & $+3.7$\\         
        \midrule
        \textbf{GGDA-GCN} & $\dag \underline{80.3} $\scalebox{0.7}{${\pm  0.3}$}
                        & $\ddag \underline{78.8} $\scalebox{0.7}{${\pm  0.2}$}
                        & $ \dag \underline{74.7}$\scalebox{0.7}{${\pm  0.3}$}
                        & $\ddag \underline{72.2}$\scalebox{0.7}{${\pm  0.3}$}
                        & $ 76.0$\scalebox{0.7}{$\pm  0.1$}
                        & $ \ddag \underline{75.9}$\scalebox{0.7}{${\pm  0.2}$}
                        & $ \underline{75.7}$\scalebox{0.7}{${\pm  0.5}$}
                        & $ \ddag\underline{74.3}$\scalebox{0.7}{${\pm  0.5}$}
                        & $ \ddag\underline{73.7}$\scalebox{0.7}{${\pm  0.3}$}
                        & $ \ddag\underline{73.2}$\scalebox{0.7}{${\pm  0.3}$}
                        & \scalebox{1.05}{$ \ddag\mathbf{81.6}$}\scalebox{0.7}{${\pm  0.3}$}
                        & \scalebox{1.05}{$ \ddag\mathbf{80.7}$}\scalebox{0.7}{${\pm  0.2}$}
                        & $\underline{+5.2}$\\
        \textbf{GGDA-GAT}& \scalebox{1.05}{$ \ddag\mathbf{81.9}$}\scalebox{0.7}{${\pm  0.2}$}
                        & \scalebox{1.05}{$ \ddag\mathbf{80.4}$}\scalebox{0.7}{${\pm  0.2}$}
                        & \scalebox{1.05}{$ \ddag \mathbf{77.2}$}\scalebox{0.7}{${\pm  0.2}$}
                        & \scalebox{1.05}{$ \ddag \mathbf{74.9}$}\scalebox{0.7}{${\pm  0.4}$}
                        & \scalebox{1.05}{$ \ddag\mathbf{77.1}$}\scalebox{0.7}{${\pm  0.2}$}
                        & \scalebox{1.05}{$ \ddag\mathbf{76.6}$}\scalebox{0.7}{${\pm  0.2}$}
                        & \scalebox{1.05}{$ \ddag\mathbf{77.4}$}\scalebox{0.7}{${\pm  0.3}$}
                        & \scalebox{1.05}{$ \ddag\mathbf{75.9}$}\scalebox{0.7}{${\pm  0.4}$}
                        & \scalebox{1.05}{$ \ddag\mathbf{75.8}$}\scalebox{0.7}{${\pm  0.3}$}
                        & \scalebox{1.05}{$ \ddag\mathbf{75.8}$}\scalebox{0.7}{${\pm  0.3}$}
                        & $ \ddag\underline{81.5}$\scalebox{0.7}{${\pm  0.2}$}
                        & $ \ddag\underline{80.3}$\scalebox{0.7}{${\pm  0.2}$}
                        & \scalebox{1.05}{$\mathbf{+6.6}$}\\
        \textbf{GGDA-SAGE}& $ 78.0$\scalebox{0.7}{$\pm  0.3$}
                        & $ 76.8$\scalebox{0.7}{$\pm  0.3$}
                        & $ 72.4$\scalebox{0.7}{$\pm  0.6$}
                        & $ 70.0$\scalebox{0.7}{$\pm  0.5$}
                        & $ 75.6$\scalebox{0.7}{$\pm  0.2$}
                        & $ 75.6$\scalebox{0.7}{$\pm  0.3$}
                        & $ 74.1$\scalebox{0.7}{$\pm  0.5$}
                        & $ 72.3$\scalebox{0.7}{$\pm  0.7$}
                        & $ 70.8$\scalebox{0.7}{$\pm  0.3$}
                        & $ 71.3$\scalebox{0.7}{$\pm  0.4$}
                        & $ 76.3$\scalebox{0.7}{$\pm  0.5$}
                        & $ 76.0$\scalebox{0.7}{$\pm  0.5$}
                        & $+2.8$\\
        \midrule
        \textbf{Oracle}& $ 97.3$\scalebox{0.7}{$\pm  0.1$}
                        & $ 96.9$\scalebox{0.7}{$\pm  0.1$}
                        & $ 96.9$\scalebox{0.7}{$\pm  0.1$}
                        & $ 96.8$\scalebox{0.7}{$\pm  0.1$}
                        & $ 97.0$\scalebox{0.7}{$\pm  0.1$}
                        & $ 97.0$\scalebox{0.7}{$\pm  0.1$}
                        & $ 96.9$\scalebox{0.7}{$\pm  0.1$}
                        & $ 96.8$\scalebox{0.7}{$\pm  0.1$}
                        & $ 97.0$\scalebox{0.7}{$\pm  0.1$}
                        & $ 97.0$\scalebox{0.7}{$\pm  0.1$}
                        & $ 97.3$\scalebox{0.7}{$\pm  0.1$}
                        & $ 96.9$\scalebox{0.7}{$\pm  0.1$}\\
        \bottomrule
    \end{tabular}
    \label{F1_DIR}
\end{table*}

\begin{table*}[h]
    \scriptsize
    \centering
    \setlength{\tabcolsep}{1pt}
    \caption{Performance of node classifications under the unsupervised graph DA setting (from a fully labeled source graph to an unlabeled target graph). The five rightmost columns are transfers from the original source graph to the last graph in the generated shift sequence. Results are reported using accuracy scores for the single-label datasets below. $\dag$ indicates $p < 0.05$ and $\ddag$ indicates $p < 0.01$ using paired t-tests against the strongest baseline. OOM indicates out-of-memory errors.}
    \begin{tabular}{cccccccccccc|c}
        \toprule
        \textbf{} &
        \multicolumn{2}{c}{\textbf{}}  
        & \multicolumn{2}{c}{\textbf{}}
        & \multicolumn{2}{c}{\textbf{Arxiv 2007 to}}
        & \multicolumn{5}{c}{\textbf{To the last step of shift generation}}
        & \multicolumn{1}{c}{\multirow{2}{*}{$\mathbf{\Delta_{Mean}}$}}\\
        \cmidrule(r){6-7}
        \cmidrule(r){8-12}
        \textbf{} & \textbf{A2 to D2}
                    &\textbf{D2 to A2}
                    &\textbf{B1 to B2}
                    &\textbf{B2 to B1}
                    &\textbf{Arxiv 2016}
                    &\textbf{Arxiv 2018}
                    &\textbf{Cora}
                    &\textbf{CiteSeer}
                    &\textbf{ENGB}
                    &\textbf{DE}
                    &\textbf{PTBR}\\
        \midrule
        \textbf{CDAN}   & $75.8$\scalebox{0.7}{$\pm 1.8$}
                        & $67.5$\scalebox{0.7}{$\pm 0.6$}
                        & $48.0$\scalebox{0.7}{$\pm 1.0$}
                        & $45.7$\scalebox{0.7}{$\pm 0.9$}
                        & $47.4$\scalebox{0.7}{$\pm 0.1$}
                        & $42.6$\scalebox{0.7}{$\pm 0.4$}
                        & $70.2$\scalebox{0.7}{$\pm 4.2$}
                        & $67.8$\scalebox{0.7}{$\pm 3.6$}
                        & $68.6$\scalebox{0.7}{$\pm 1.9$}
                        & $70.7$\scalebox{0.7}{$\pm 1.8$}
                        & $66.5$\scalebox{0.7}{$\pm 2.3$}
                        & $+1.1$\\
        \textbf{MDD}    & $74.2$\scalebox{0.7}{$\pm 1.5$}
                        & $69.8$\scalebox{0.7}{$\pm 1.3$}
                        & $44.0$\scalebox{0.7}{$\pm 0.7$}
                        & $43.5$\scalebox{0.7}{$\pm 0.8$}
                        & $52.8$\scalebox{0.7}{$\pm 0.5$}
                        & $51.3$\scalebox{0.7}{$\pm 1.3$}
                        & $59.2$\scalebox{0.7}{$\pm 1.7$}
                        & $64.1$\scalebox{0.7}{$\pm 3.2$}
                        & $65.6$\scalebox{0.7}{$\pm 4.2$}
                        & $67.9$\scalebox{0.7}{$\pm 2.2$}
                        & $65.9$\scalebox{0.7}{$\pm 0.4$}
                        & $-0.1$\\
        \midrule                
        \textbf{UDA-GCN} 
                        & $79.1$\scalebox{0.7}{$\pm 2.1$}
                        & \scalebox{1.05}{$\mathbf{71.5}$}\scalebox{0.7}{${\pm 1.0}$}
                        & $43.6$\scalebox{0.7}{$\pm 1.8$}
                        & $45.8$\scalebox{0.7}{$\pm 2.5$}
                        & $54.1$\scalebox{0.7}{$\pm 0.2$}
                        & $53.5$\scalebox{0.7}{$\pm 0.6$}
                        & $56.5$\scalebox{0.7}{$\pm 3.6$}
                        & $71.4$\scalebox{0.7}{$\pm 4.5$}
                        & $62.2$\scalebox{0.7}{$\pm 3.4$}
                        & $69.2$\scalebox{0.7}{$\pm 2.8$}
                        & $71.6$\scalebox{0.7}{$\pm 2.6$}
                        & $+1.8$\\
        \textbf{ACDNE}
                        & $70.1$\scalebox{0.7}{$\pm 1.4$}
                        & $60.3$\scalebox{0.7}{$\pm 1.5$}
                         & \scalebox{1.05}{$\mathbf{53.1}$}\scalebox{0.7}{${\pm 1.3}$}
                        & \scalebox{1.05}{$\mathbf{51.3}$}\scalebox{0.7}{${\pm 1.3}$}
                        & $43.5$\scalebox{0.7}{$\pm 0.6$}
                        & $40.9$\scalebox{0.7}{$\pm 1.3$}
                        & $60.4$\scalebox{0.7}{$\pm 3.9$}
                        & $53.6$\scalebox{0.7}{$\pm 3.1$}
                        & $50.8$\scalebox{0.7}{$\pm 2.7$}
                        & $56.8$\scalebox{0.7}{$\pm 2.6$}
                        & $68.1$\scalebox{0.7}{$\pm 2.2$}
                        & $-4.5$\\
        \textbf{ASN}
                        & $\underline{85.2}$\scalebox{0.7}{${\pm 1.2}$}
                        & $64.9$\scalebox{0.7}{$\pm 1.4$}
                        & $30.1$\scalebox{0.7}{$\pm 1.3$}
                        & $38.2$\scalebox{0.7}{$\pm 2.3$}
                        & \text{OOM}
                        & \text{OOM}
                        & $78.8$\scalebox{0.7}{$\pm 1.9$}
                        & $70.9$\scalebox{0.7}{$\pm 3.0$}
                        & $56.5$\scalebox{0.7}{$\pm 1.0$}
                        & $59.6$\scalebox{0.7}{$\pm 2.8$}
                        & $67.7$\scalebox{0.7}{$\pm 2.7$}
                        & $+1.4$\\
        \textbf{GRADE}
                        & $ 61.1$\scalebox{0.7}{$\pm 1.9$}
                        & $ 64.4$\scalebox{0.7}{$\pm 2.4$}
                        & $ 42.5$\scalebox{0.7}{$\pm 1.3$}
                        & $ 41.4$\scalebox{0.7}{$\pm 1.0$}
                        & $ 46.8$\scalebox{0.7}{$\pm 0.4$}
                        & \text{OOM}
                        & $ 57.9$\scalebox{0.7}{$\pm 3.2$}
                        & $ 66.5$\scalebox{0.7}{$\pm 2.5$}
                        & $ 58.1$\scalebox{0.7}{$\pm 2.8$}
                        & $ 61.2$\scalebox{0.7}{$\pm 3.3$}
                        & $ 63.4$\scalebox{0.7}{$\pm 3.3$}
                        & $-3.6$\\
        \textbf{StruRW} & $ 63.7$\scalebox{0.7}{$\pm 2.8$}
                        & $ 66.6$\scalebox{0.7}{$\pm 2.7$}
                        & $ 38.7$\scalebox{0.7}{$\pm 2.6$}
                        & $ 31.9$\scalebox{0.7}{$\pm 1.3$}
                        & $ 31.7$\scalebox{0.7}{$\pm 0.7$}
                        & $ 27.2$\scalebox{0.7}{$\pm 0.7$}
                        & $ 67.6$\scalebox{0.7}{$\pm 1.9$}
                        & $ 62.1$\scalebox{0.7}{$\pm 4.3$}
                        & $ 55.6$\scalebox{0.7}{$\pm 2.5$}
                        & $ 60.8$\scalebox{0.7}{$\pm 2.4$}
                        & $ 61.7$\scalebox{0.7}{$\pm 2.5$}
                        & $-8.3$\\
        \textbf{SpecReg}
                        & $ 75.6$\scalebox{0.7}{$\pm 6.9$}
                        & $ 68.0$\scalebox{0.7}{$\pm 2.0$}
                        & $ 44.7$\scalebox{0.7}{$\pm 1.0$}
                        & $ 43.7$\scalebox{0.7}{$\pm 1.0$}
                        & $ 53.2$\scalebox{0.7}{$\pm 0.5$}
                        & $ 47.6$\scalebox{0.7}{$\pm 1.9$}
                        & $ 72.0$\scalebox{0.7}{$\pm 4.3$}
                        & $ 75.6$\scalebox{0.7}{$\pm 3.7$}
                        & $ 68.1$\scalebox{0.7}{$\pm 3.6$}
                        & $ 69.7$\scalebox{0.7}{$\pm 1.4$}
                        & $ 69.5$\scalebox{0.7}{$\pm 1.2$}
                        & $+2.6$\\
        \textbf{GraphAlign} & $ 67.8$\scalebox{0.7}{$\pm 1.4$}
                            & $ 68.0$\scalebox{0.7}{$\pm 1.6$}
                            & $ 45.1$\scalebox{0.7}{$\pm 1.2$}
                            & $ 45.9$\scalebox{0.7}{$\pm 0.4$}
                            & $ 30.1$\scalebox{0.7}{$\pm 0.3$}
                            & $ 27.5$\scalebox{0.7}{$\pm 0.7$}
                            & $ 61.0$\scalebox{0.7}{$\pm 2.4$}
                            & $ 56.2$\scalebox{0.7}{$\pm 5.2$}
                            & $ 69.4$\scalebox{0.7}{$\pm 3.8$}
                            & $ 69.9$\scalebox{0.7}{$\pm 4.8$}
                            & $ 67.8$\scalebox{0.7}{$\pm 2.0$}
                            & $-4.6$\\
        \textbf{A2GNN}
                        & $ 76.1$\scalebox{0.7}{$\pm 0.1$}
                        & $ 67.9$\scalebox{0.7}{$\pm 0.7$}
                        & $ 37.2$\scalebox{0.7}{$\pm 1.6$}
                        & $ 39.5$\scalebox{0.7}{$\pm 1.2$}
                        & $ 48.4$\scalebox{0.7}{$\pm 1.1$}
                        & $ 40.1$\scalebox{0.7}{$\pm 2.6$}
                        & $ 79.7$\scalebox{0.7}{$\pm 1.9$}
                        & $ 70.4$\scalebox{0.7}{$\pm 1.7$}
                        & $ 67.7$\scalebox{0.7}{$\pm 2.2$}
                        & $ 67.1$\scalebox{0.7}{$\pm 2.1$}
                        & $ 68.9$\scalebox{0.7}{$\pm 1.7$}
                        & $+0.4$\\
        \textbf{Pair-Align}
                        & $ 63.3$\scalebox{0.7}{$\pm 0.4$}
                        & $ 65.7$\scalebox{0.7}{$\pm 0.4$}
                        & $ 39.1$\scalebox{0.7}{$\pm 0.7$}
                        & $ 34.0$\scalebox{0.7}{$\pm 0.6$}
                        & $ 39.2$\scalebox{0.7}{$\pm 0.4$}
                        & $ 37.2$\scalebox{0.7}{$\pm 0.7$}
                        & $ 60.7$\scalebox{0.7}{$\pm 1.4$}
                        & $ 71.4$\scalebox{0.7}{$\pm 2.7$}
                        & $ 61.5$\scalebox{0.7}{$\pm 2.0$}
                        & $ 60.4$\scalebox{0.7}{$\pm 1.3$}
                        & $ 67.9$\scalebox{0.7}{$\pm 1.9$}
                        & $-5.3$\\
        \textbf{TDSS}
                        & $ 72.5$\scalebox{0.7}{$\pm 0.7$}
                        & $ 64.2$\scalebox{0.7}{$\pm 0.7$}
                        & $ 34.1$\scalebox{0.7}{$\pm 1.1$}
                        & $ 32.9$\scalebox{0.7}{$\pm 0.8$}
                        & $ 43.3$\scalebox{0.7}{$\pm 0.4$}
                        & \text{OOM}
                        & $ 79.7$\scalebox{0.7}{$\pm 1.4$}
                        & $ 70.5$\scalebox{0.7}{$\pm 2.2$}
                        & $ 62.3$\scalebox{0.7}{$\pm 1.3$}
                        & $ 62.0$\scalebox{0.7}{$\pm 1.4$}
                        & $ 69.1$\scalebox{0.7}{$\pm 1.5$}
                        & $-0.9$\\
        \midrule
        \textbf{GGDA-GCN} 
                        & \scalebox{1.05}{$\ddag\mathbf{91.5}$}\scalebox{0.7}{${\pm 0.5}$}
                        & $70.0$\scalebox{0.7}{$\pm 0.8$}
                        & $\underline{52.4}$\scalebox{0.7}{${\pm 1.7}$}
                        & $49.3$\scalebox{0.7}{$\pm 0.4$}
                        & \scalebox{1.05}{$ \ddag\mathbf{56.8}$}\scalebox{0.7}{$\pm 0.5$}
                        & \scalebox{1.05}{$ \ddag\mathbf{56.6}$}\scalebox{0.7}{$\pm 0.5$}
                        & $\ddag\underline{86.6}$\scalebox{0.7}{${\pm 2.1}$}
                        & $74.1$\scalebox{0.7}{$\pm 3.4$}
                        & \scalebox{1.05}{$\ddag\mathbf{75.3}$}\scalebox{0.7}{${\pm 0.6}$}
                        & \scalebox{1.05}{$\ddag\mathbf{74.6}$}\scalebox{0.7}{${\pm 0.4}$}
                        & $\underline{72.2}$\scalebox{0.7}{${\pm 0.8}$}
                        & \scalebox{1.05}{$\mathbf{+9.1}$}\\
        \textbf{GGDA-GAT}
                        & \scalebox{1.05}{$ \ddag\mathbf{91.5}$}\scalebox{0.7}{${\pm 0.4}$}
                        & $ \underline{70.3}$\scalebox{0.7}{${\pm 0.5}$}
                        & $ 31.5$\scalebox{0.7}{$\pm 0.6$}
                        & $ 30.9$\scalebox{0.7}{$\pm 0.9$}
                        & $ \ddag 54.7$\scalebox{0.7}{$\pm 0.4$}
                        & $ \dag \underline{54.1}$\scalebox{0.7}{$\pm 0.5$}
                        & $ 82.6$\scalebox{0.7}{$\pm 2.2$}
                        & $\ddag \underline{82.7}$\scalebox{0.7}{${\pm 1.1}$}
                        & $ \ddag\underline{74.2}$\scalebox{0.7}{${\pm 1.6}$}
                        & $\ddag \underline{74.5}$\scalebox{0.7}{${\pm 1.7}$}
                        & \scalebox{1.05}{$ \ddag\mathbf{76.9}$}\scalebox{0.7}{${\pm 1.5}$}
                        & $\underline{+5.9}$\\
        \textbf{GGDA-SAGE}
                        & $ 78.5$\scalebox{0.7}{$\pm 0.6$}
                        & $ 58.2$\scalebox{0.7}{$\pm 0.5$}
                        & $ 50.2$\scalebox{0.7}{$\pm 1.0$}
                        & $ \underline{50.3}$\scalebox{0.7}{${\pm 1.0}$}
                        & $ \ddag\underline{55.0}$\scalebox{0.7}{$\pm 0.1$}
                        & $ 53.2$\scalebox{0.7}{$\pm 0.4$}
                        & \scalebox{1.05}{$ \ddag\mathbf{87.4}$}\scalebox{0.7}{${\pm 1.2}$}
                        & \scalebox{1.05}{$ \ddag\mathbf{83.6}$}\scalebox{0.7}{${\pm 1.1}$}
                        & $ 61.8$\scalebox{0.7}{$\pm 1.2$}
                        & $ 64.8$\scalebox{0.7}{$\pm 1.0$}
                        & $ 70.5$\scalebox{0.7}{$\pm 0.9$}
                        & $+4.0$\\
        \midrule
        \textbf{GGDA-GT}& /
                        & /
                        & /
                        & /
                        & /
                        & /
                        & $ 93.8$\scalebox{0.7}{$\pm 0.2$}
                        & $ 88.5$\scalebox{0.7}{$\pm 0.2$}
                        & $ 77.7$\scalebox{0.7}{$\pm 0.4$}
                        & $ 74.8$\scalebox{0.7}{$\pm 0.7$}
                        & $ 77.3$\scalebox{0.7}{$\pm 0.3$}\\
        \textbf{Oracle}
                        & $ 92.4$\scalebox{0.7}{$\pm 0.1$}
                        & $ 99.2$\scalebox{0.7}{$\pm 0.1$}
                        & $ 71.4$\scalebox{0.7}{$\pm 2.2$}
                        & $ 72.3$\scalebox{0.7}{$\pm 0.9$}
                        &$ 67.5$\scalebox{0.7}{$\pm 0.1$}
                        &$ 67.9$\scalebox{0.7}{$\pm 0.1$}
                        & $ 99.0$\scalebox{0.7}{$\pm 0.2$}
                        & $ 95.0$\scalebox{0.7}{$\pm 0.2$}
                        & $ 81.1$\scalebox{0.7}{$\pm 0.4$}
                        & $ 77.6$\scalebox{0.7}{$\pm 0.4$}
                        & $ 78.0$\scalebox{0.7}{$\pm 0.4$}\\
        \bottomrule
    \end{tabular}
    \label{acc_DIR}
\end{table*}

\begin{figure*}[t]
    \centering
    \setlength{\tabcolsep}{0.4em}
        \begin{tabular}{ccc} 
        \includegraphics[height=84pt, trim=0cm 0cm 0cm 0cm,clip]{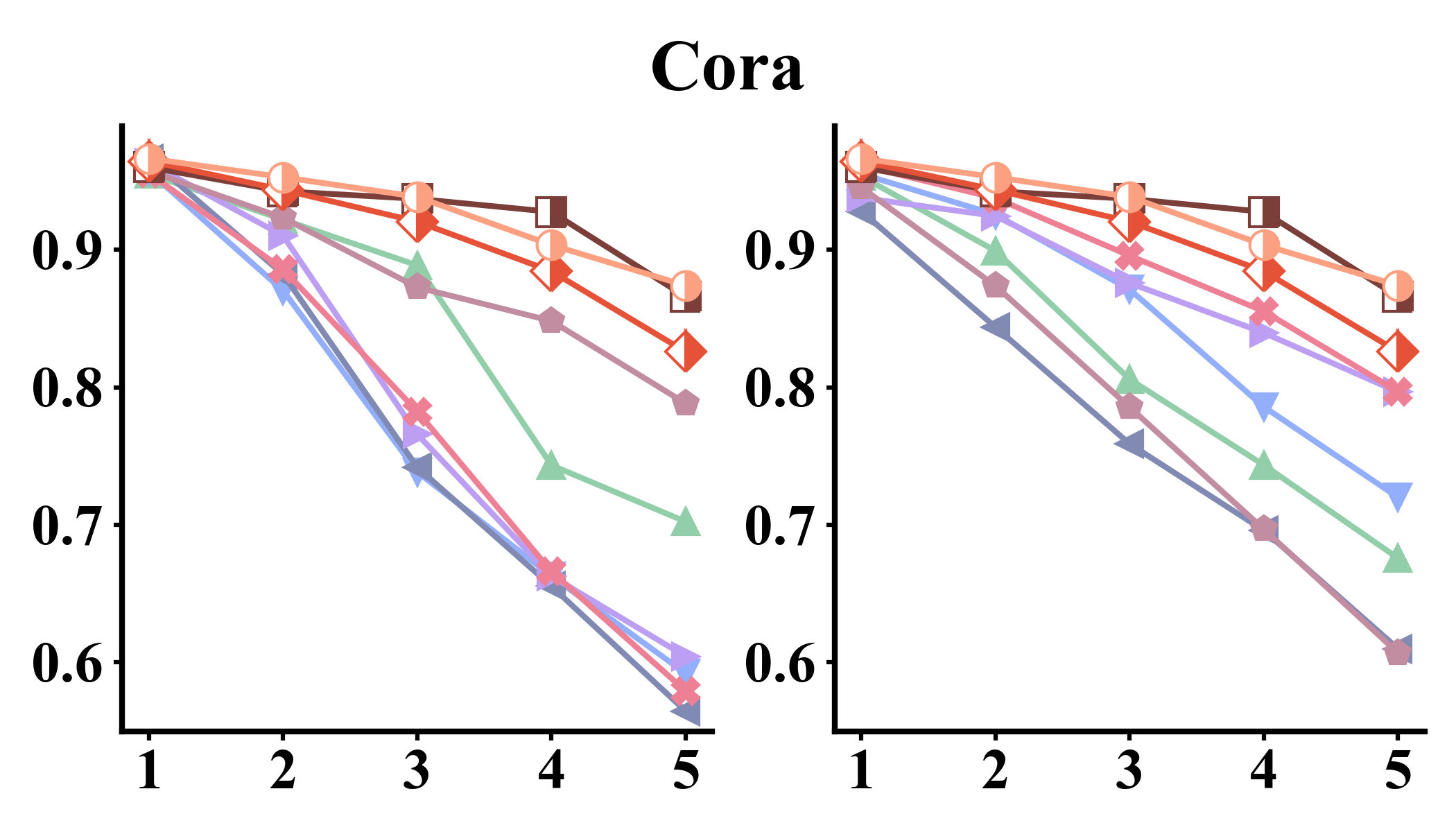} & 
        \includegraphics[height=84pt, trim=0cm 0cm 0cm 0cm,clip]{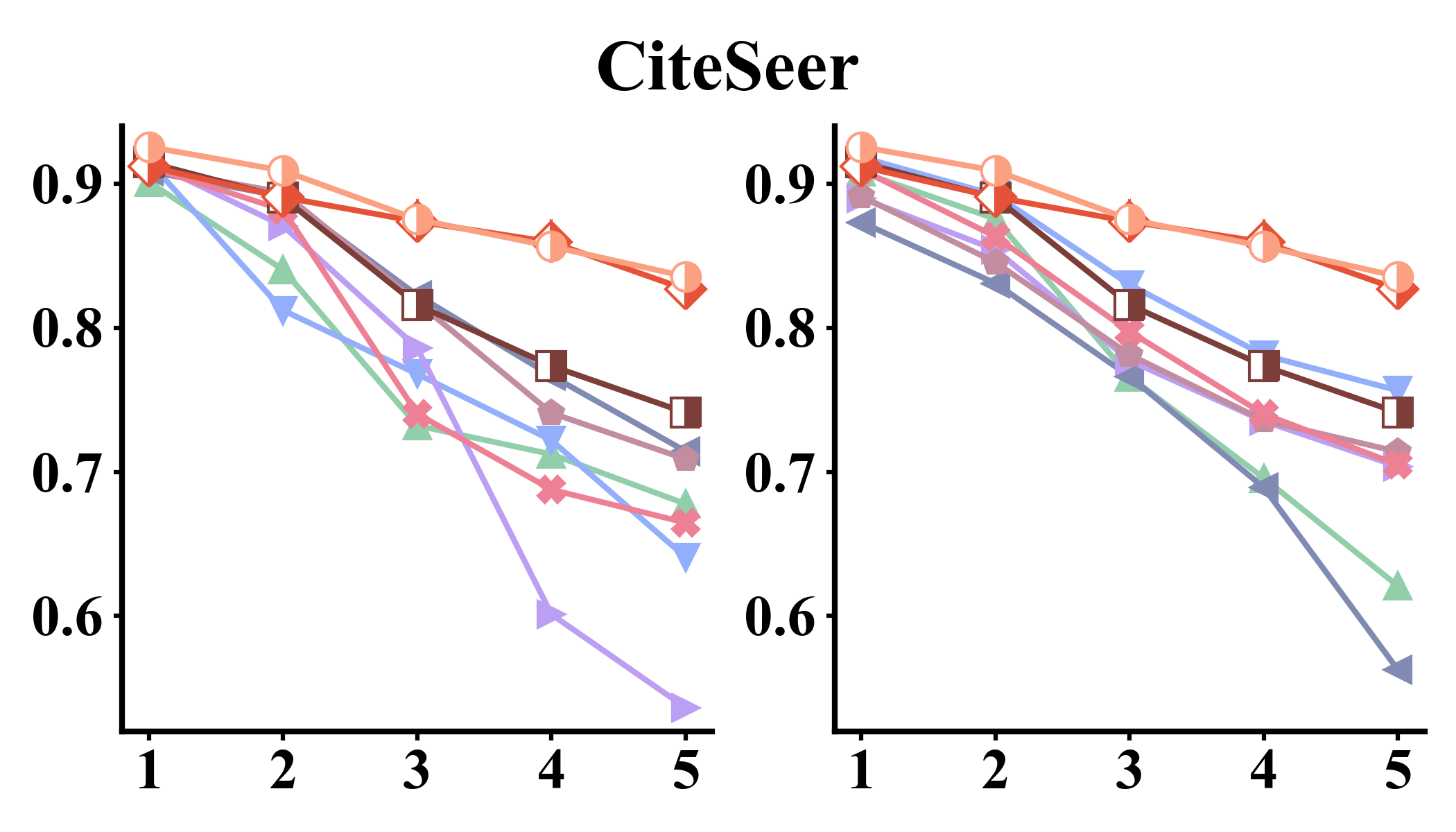} & 
        \multirow{2}{*}[73pt]{\includegraphics[height=162pt, trim=0cm 0cm 0cm 0cm,clip]{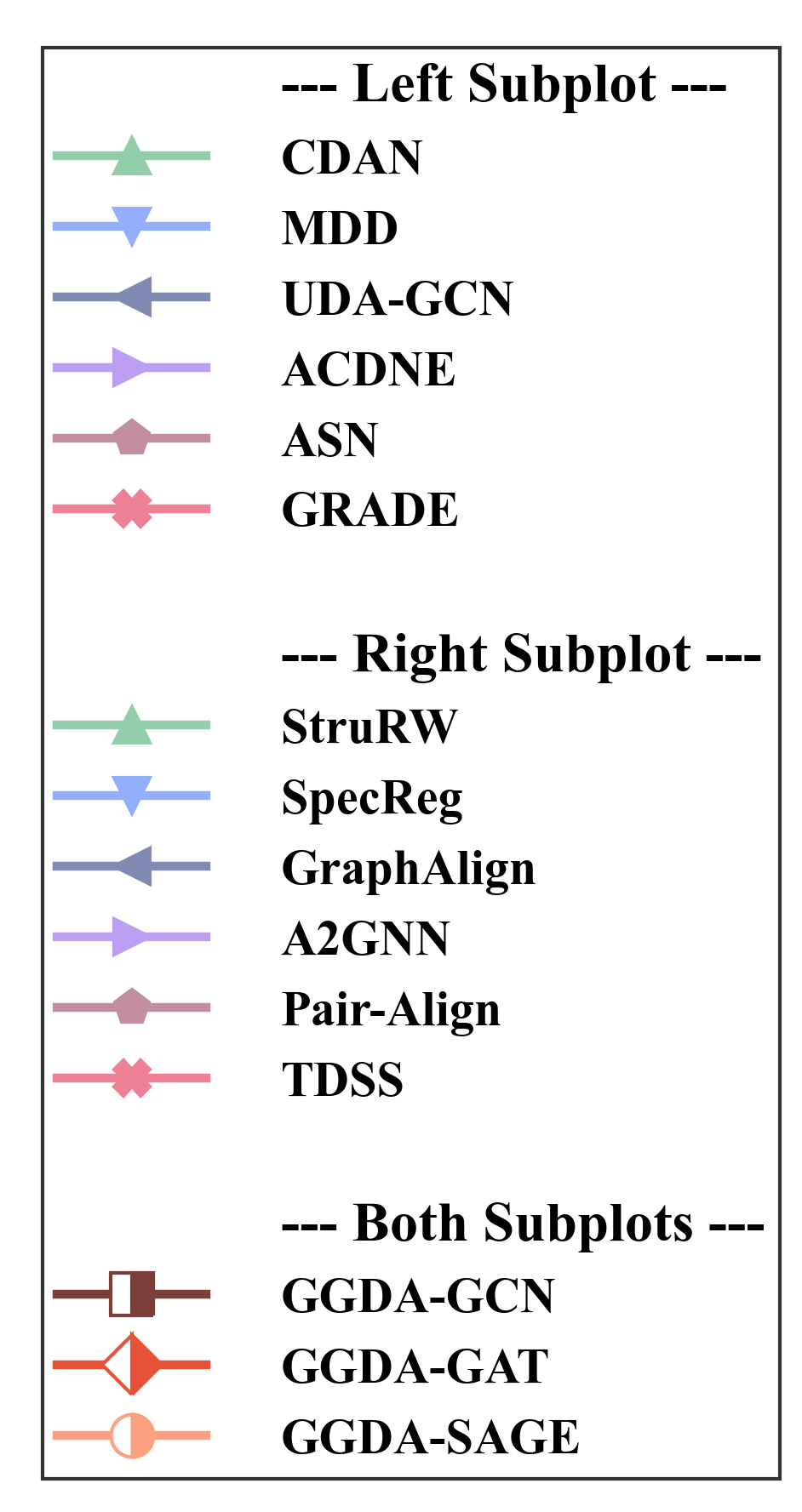}}\\
        \includegraphics[height=84pt, trim=0cm 0cm 0cm 0cm,clip]{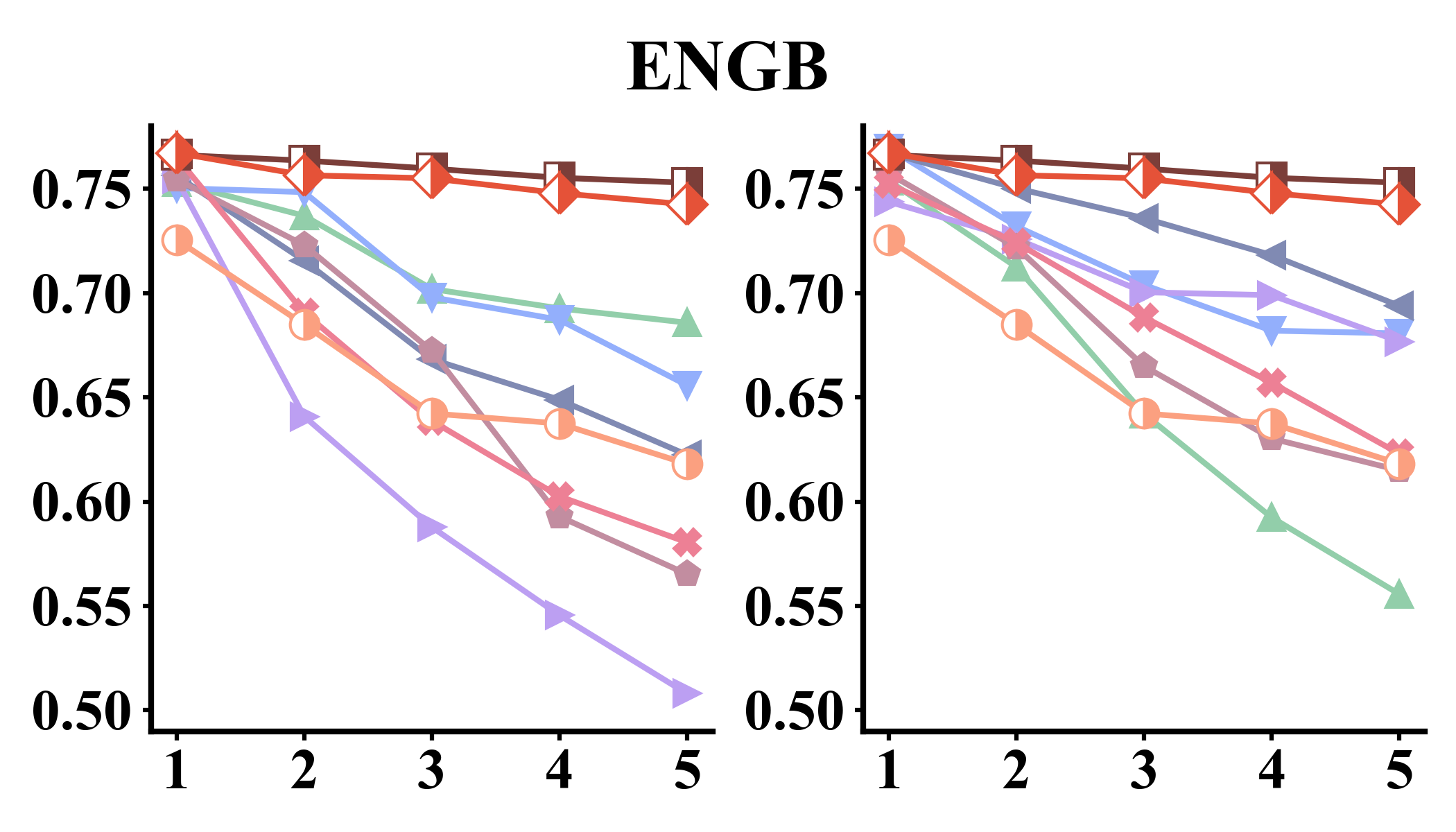} & 
        \includegraphics[height=84pt, trim=0cm 0cm 0cm 0cm,clip]{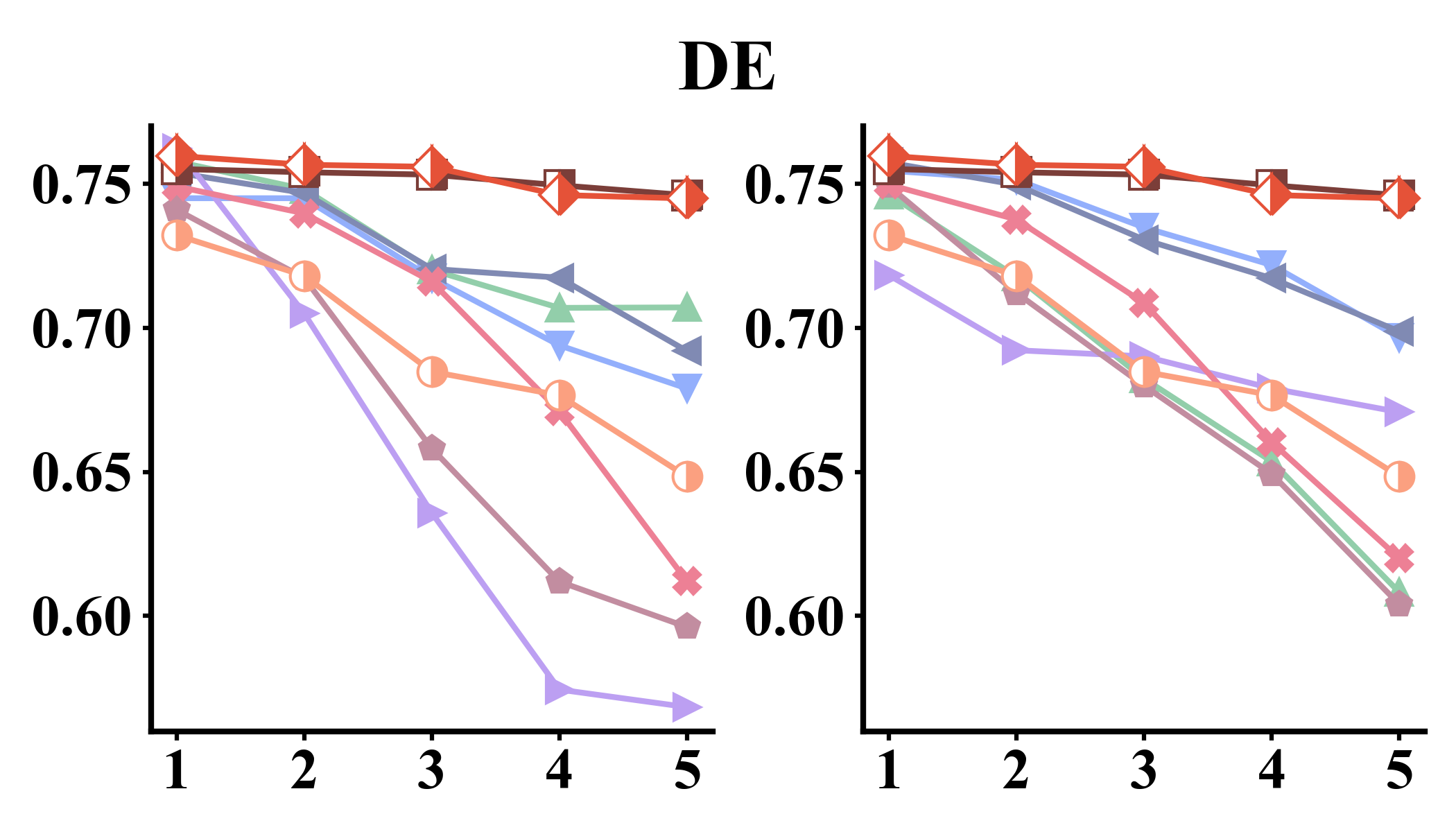} &        
        \end{tabular}
    \caption{Node classification accuracy on target graphs with varying discrepancy from source. The x-axis shows source-target discrepancy level ranked 1 (closest) to 5 (farthest). The y-axis shows classification accuracy. For each transfer, the left subplot shows GGDA vs. first‑half baselines, the right subplot shows GGDA vs. second‑half baselines. The legend indicates subplot membership (left/right/both).}
    \label{lineplot_DIR}
    \Description[Line graphs comparing GGDA model accuracy against twelve baselines as source-target discrepancy increases across four datasets.]{Eight line plots in a 2 times 4 grid show node classification accuracy (y-axis) versus source-target discrepancy (x-axis) for Cora, CiteSeer, ENGB, and DE datasets. For each dataset, three GGDA variants (GGDA-GCN, GGDA-GAT, GGDA-SAGE) are compared against two sets of baselines. In all plots, GGDA models generally exhibit higher stability and accuracy than the twelve baselines. As discrepancy increases from 1 to 5, baseline performance drops sharply, while GGDA models maintain a shallow decline, consistently forming the top-most lines in every subplot.}
\end{figure*}

\begin{table*}
\caption{Performance drop in target graph node classifications across variants compared to the complete GGDA framework. All drops against complete GGDA are statistically significant, i.e., $p < 0.05$ using paired t-tests, except those marked $(ns)$.}
    \scriptsize
    \centering
    \setlength{\tabcolsep}{1.8pt}
    \begin{tabular}{ccccccccccc|c}
        \toprule
        \textbf{} & \textbf{A to D}
                    &\textbf{D to C}
                    &\textbf{C to A}
                    &\textbf{A2 to D2}
                    &\textbf{B1 to B2}
                    &\textbf{Arxiv\scalebox{0.7}{(07-16)}}
                    &\textbf{Arxiv\scalebox{0.7}{(07-18)}}
                    &\textbf{CiteSeer}
                    &\textbf{DE}
                    &\textbf{PTBR}
                    &\textbf{Avg. Drop}\\
        \midrule
        \textbf{Source} & $-10.7$\scalebox{0.7}{$\pm 0.7$}
                        & $-11.0$\scalebox{0.7}{$\pm 0.9$}
                        & $-6.0$\scalebox{0.7}{$\pm 0.2$}
                        & $-17.0$\scalebox{0.7}{$\pm 0.9$}
                        & $-15.1$\scalebox{0.7}{$\pm 1.7$}
                        & $-5.6$\scalebox{0.7}{$\pm 0.9$}
                        & $-7.8$\scalebox{0.7}{$\pm 0.9$}
                        & $-15.8$\scalebox{0.7}{$\pm 4.9$}
                        & $-34.9$\scalebox{0.7}{$\pm 0.6$}
                        & $-37.2$\scalebox{0.7}{$\pm 1.1$}
                        & $-16.1$\\
        \textbf{Direct ST} & $-6.7$\scalebox{0.7}{$\pm 0.6$}
                        & $-5.0$\scalebox{0.7}{$\pm 0.5$}
                        & $-3.2$\scalebox{0.7}{$\pm 0.3$}
                        & $-11.8$\scalebox{0.7}{$\pm 1.2$}
                        & $-11.2$\scalebox{0.7}{$\pm 2.4$}
                        & $-5.1$\scalebox{0.7}{$\pm 0.8$}
                        & $-7.6$\scalebox{0.7}{$\pm 1.1$}
                        & $-7.9$\scalebox{0.7}{$\pm 5.0$}
                        & $-7.3$\scalebox{0.7}{$\pm 6.4$}
                        & $-22.7$\scalebox{0.7}{$\pm 2.9$}
                        & $-8.9$\\
        \textbf{DIR}    & $-9.6$\scalebox{0.7}{$\pm 1.3$}
                        & $-13.1$\scalebox{0.7}{$\pm 1.2$}
                        & $-8.9$\scalebox{0.7}{$\pm 0.8$}
                        & $-11.1$\scalebox{0.7}{$\pm 3.5$}
                        & $-4.3$\scalebox{0.7}{$\pm 3.4$}
                        & $-4.3$\scalebox{0.7}{$\pm 1.0$}
                        & $-4.1$\scalebox{0.7}{$\pm 1.1$}
                        & $-21.2$\scalebox{0.7}{$\pm 6.5$}
                        & $-3.1$\scalebox{0.7}{$\pm 3.8(ns)$}
                        & $-6.6$\scalebox{0.7}{$\pm 1.4$}
                        & $-8.6$\\
        \textbf{GGDA-I}& $-2.5$\scalebox{0.7}{$\pm 0.7$}
                        & $-5.2$\scalebox{0.7}{$\pm 0.6$}
                        & $-1.5$\scalebox{0.7}{$\pm 0.4$}
                        & $-7.5$\scalebox{0.7}{$\pm 2.6$}
                        & $-10.2$\scalebox{0.7}{$\pm 3.1$}
                        & $-2.1$\scalebox{0.7}{$\pm 1.0$}
                        & $-2.9$\scalebox{0.7}{$\pm 1.1$}
                        & $-4.1$\scalebox{0.7}{$\pm 7.1(ns)$}
                        & $-6.1$\scalebox{0.7}{$\pm 3.7$}
                        & $-4.7$\scalebox{0.7}{$\pm 4.1$}
                        & $-4.7$\\
        \textbf{GGDA-R} & $-2.7$\scalebox{0.7}{$\pm 0.7$}
                        & $-1.3$\scalebox{0.7}{$\pm 0.7$}
                        & $-1.7$\scalebox{0.7}{$\pm 0.6$}
                        & $-0.7$\scalebox{0.7}{$\pm 0.6$}
                        & $-3.8$\scalebox{0.7}{$\pm 2.1$}
                        & $-2.0$\scalebox{0.7}{$\pm 0.5$}
                        & $-2.0$\scalebox{0.7}{$\pm 0.9$}
                        & $-2.0$\scalebox{0.7}{$\pm 5.9(ns)$}
                        & $-0.7$\scalebox{0.7}{$\pm 0.7$}
                        & $-1.5$\scalebox{0.7}{$\pm 1.0$}
                        & $-1.9$\\
        \bottomrule
    \end{tabular}
    \label{ablation}
\end{table*}

\subsection{Ablation Studies}
To investigate the effectiveness of different components, we conduct ablation studies as follows. (1) \textbf{Source}: direct transfer with a classifier trained on source; (2) \textbf{Direct ST}: self-training without intermediate samples (i.e., iterative pseudo-labeling but omitting intermediate domain construction); (3) \textbf{DIR}: a prevalent graph DA framework with an adversarial domain classifier for distribution alignment; (4) \textbf{GGDA-I} (\textbf{I}solated): GGDA without vertex‑based progression, treating each isolated graph $\tilde{\mu}_k$ as a separate domain; (5) \textbf{GGDA-R} (\textbf{R}andom): GGDA with random matching function $m(\cdot)$ for graph generation (i.e., omitting entropy‑guided matching). We adopt GCN in all cases, with the results shown in Table \ref{ablation}. The significant performance drop of Direct ST and DIR highlights the crucial role of intermediate domains in exploiting the underlying data patterns for a robust graph DA. Meanwhile, DIR shows signs of negative transfer (i.e., lower accuracy than Source), underscoring the instability of domain alignment and its heavy reliance on additional model designs.
The performance drop of GGDA‑I confirms that vertex‑based progression is essential to translate FGW‑generated graphs into consistent gains, as it actively binds knowledge‑preserving vertices across domains (e.g., in D to C and B1 to B2, intermediate graphs alone fail to outperform Direct ST or DIR).
Meanwhile, the performance drop of GGDA-R reflects the improved information integrity from entropy-guided matching in graph generation.

\subsection{Hyperparameter Analysis}
\begin{figure}[t]
    \setlength{\tabcolsep}{0.7em}
        \begin{tabular}{cc} 
        \includegraphics[height=92pt, trim=0cm 0cm 0cm 0cm,clip]{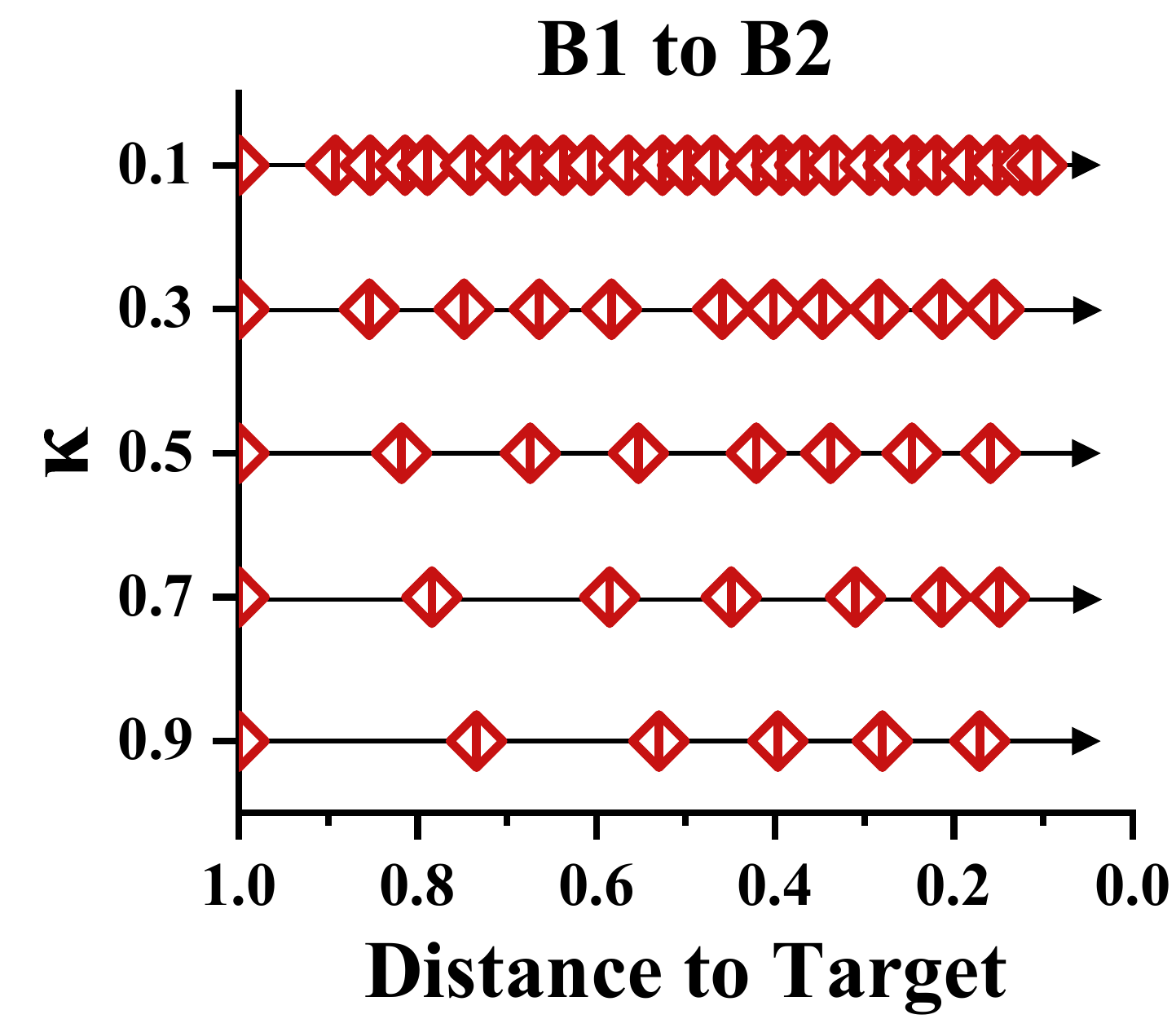} & 
        \includegraphics[height=92pt, trim=0cm 0cm 0cm 0cm,clip]{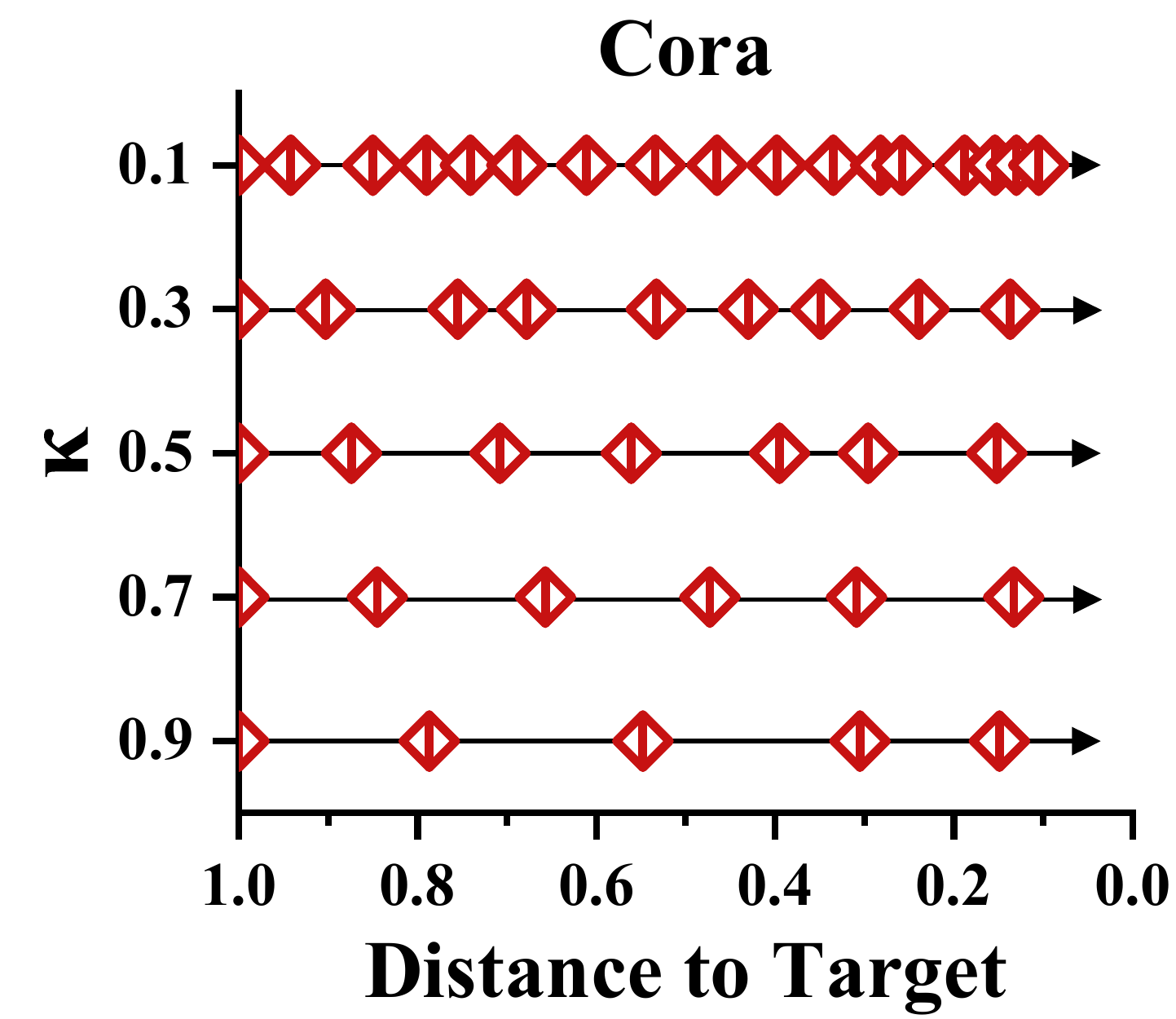}
        \end{tabular}
    \caption{Progression of the constructed domain sequence under different $\kappa$. The x-axis shows the normalized embedding distance between the current and target domains.}
    \Description[Two dot plots showing the progression of intermediate domains between source and target for different kappa values.]{Two horizontal plots for the B1-to-B2 and Cora datasets illustrate the adaptation path from source to target. The y-axis shows kappa values from 0.1 to 0.9, while the x-axis shows the normalized embedding distance to the target domain (from 1.0 to 0.0). Red diamonds mark the positions of constructed intermediate domains. For a low kappa of 0.1, the diamonds are numerous and densely packed across the distance. As kappa increases toward 0.9, the number of intermediate domains decreases significantly, visualizing how kappa controls the granularity of the domain sequence construction.}
    \label{emb}
\end{figure}

GGDA is governed by four interpretable hyperparameters: number of intermediate graphs $K-1$, score penalty $\eta$, vertex selection ratio $\kappa$, and mass decay rate $\beta$. Their effects are illustrated in Figure \ref{hyperparameters} and \ref{emb}:
\begin{figure}[t]
    \setlength{\tabcolsep}{0.4em}
        \begin{tabular}{cccc} 
        \includegraphics[height=92pt, trim=2cm 6cm 3cm 5cm,clip]{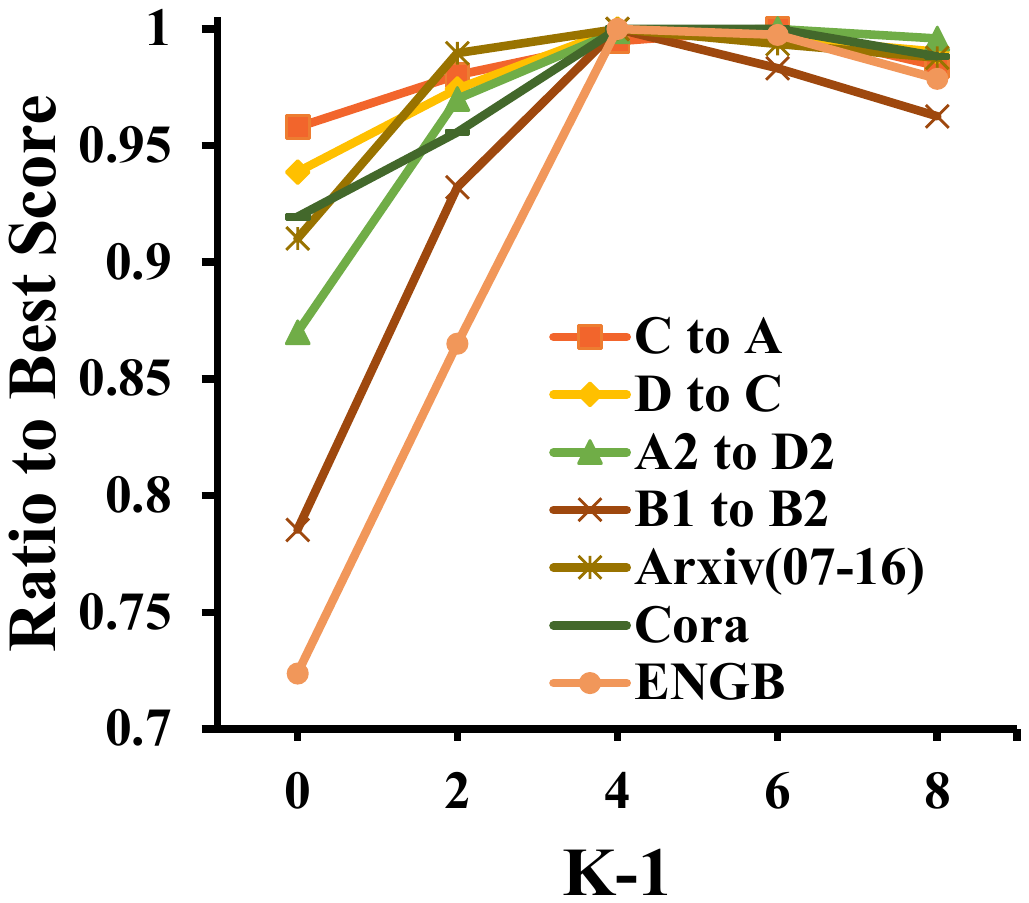} & 
        \includegraphics[height=92pt, trim=2cm 6cm 2.5cm 5cm,clip]{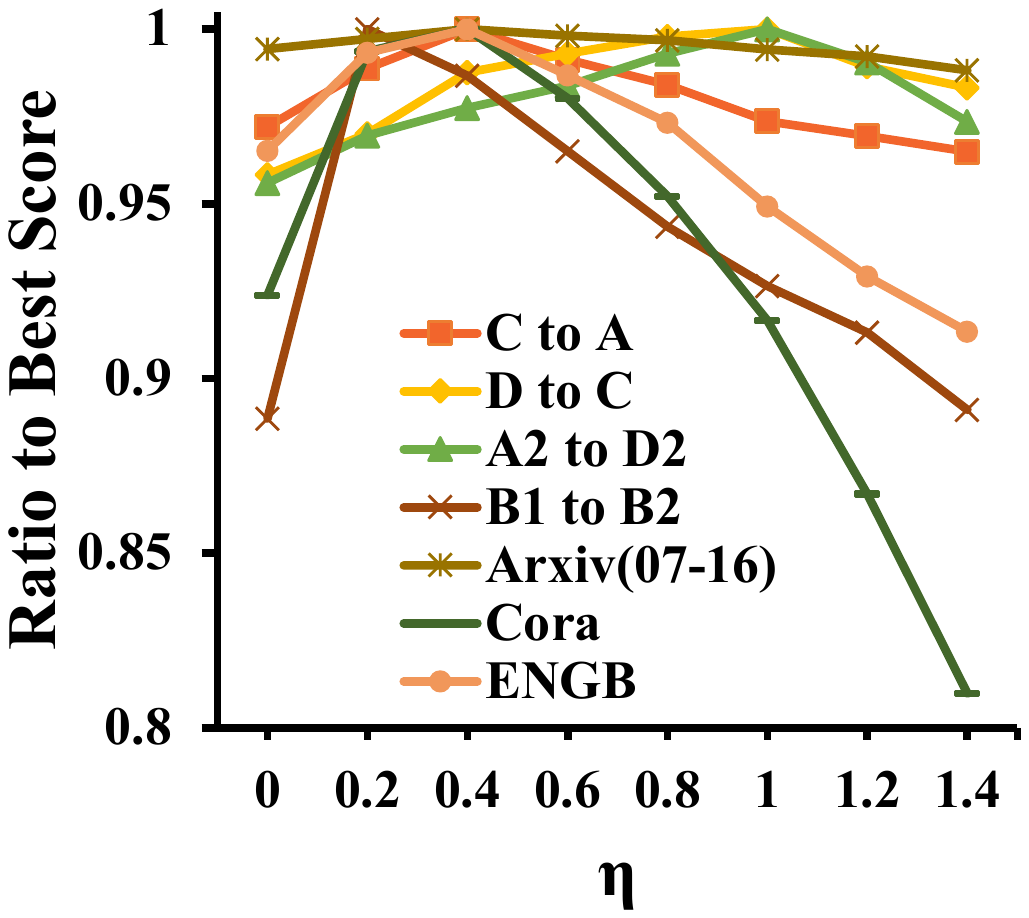} &
        \includegraphics[height=92pt, trim=4.8cm 2.3cm 5.3cm 1.2cm,clip]{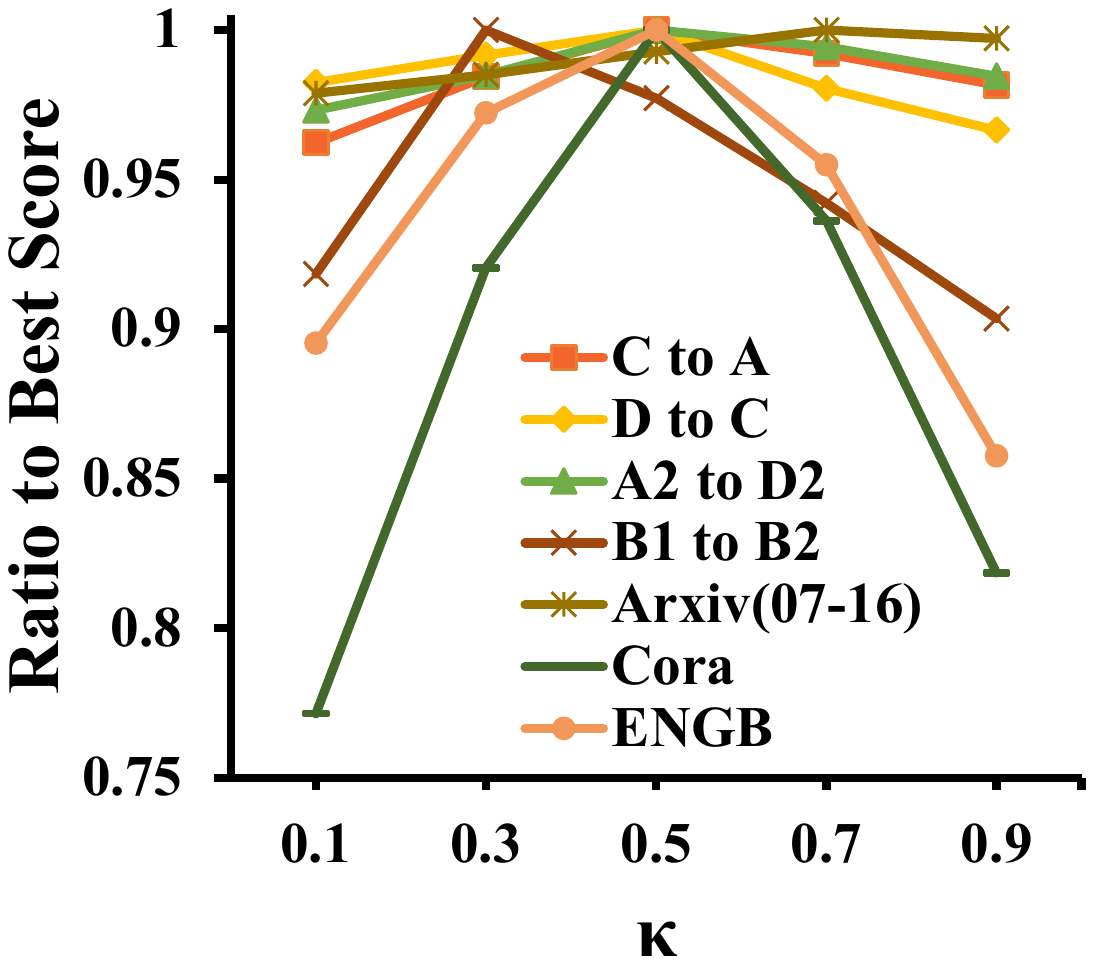} & 
        \includegraphics[height=92pt, trim=2cm 6cm 2.7cm 5cm,clip]{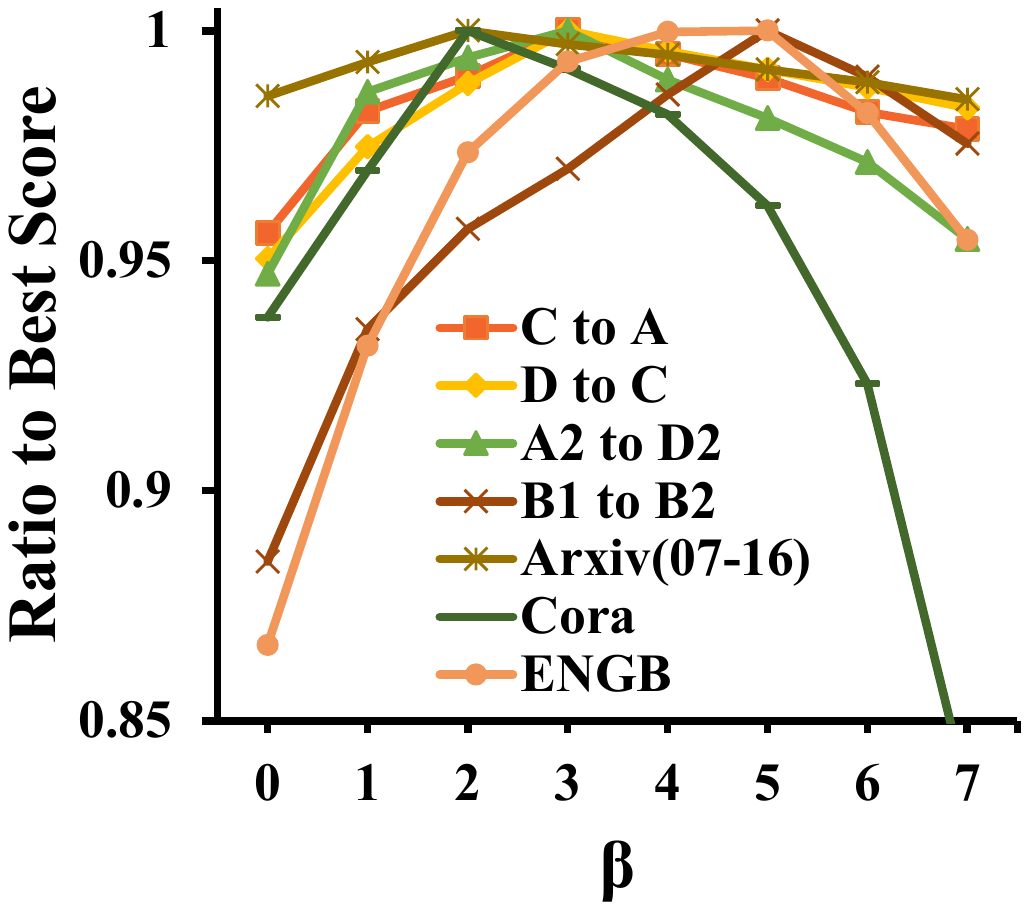} 
        \end{tabular}
    \caption{Performance of our GGDA framework with varying values of hyperparameters $K$, $\eta$, $\kappa$ and $\beta$.}
    \Description[Four sensitivity plots showing the performance ratio of GGDA across seven datasets as hyperparameters K-1, eta, kappa, and beta vary.]{Four side-by-side line plots illustrate how the "Ratio to Best Score" (y-axis) changes with four hyperparameters (x-axes). The first plot shows performance rising sharply and plateauing after K-1 equals 4. The second plot shows stable performance for eta between 0.2 and 0.6, with declines at higher values for some datasets. The third and fourth plots show bell-shaped curves for kappa and beta, with optimal performance peaking around 0.5 for kappa and 3 for beta.}
    \label{hyperparameters}
\end{figure}
(1) Regarding $K-1$, the number of intermediate graphs generated via FGW, setting it to $0$ corresponds to direct self-training without intermediate graphs, which leads to unreliable pseudo-labeling under large domain discrepancies. Performance improves sharply once intermediate graphs are introduced ($K-1 \geq 1$), and stabilizes after $K-1 \geq 4$, indicating that a modest number of intermediate graphs suffices to bridge the domain gap effectively. 
(2) The score penalty $\eta$ governs the degree of penalty applied to the regularized score $c_u^t$. An overly small $\eta$ risks selecting distant, mislabeled vertices due to distribution shift, while a large $\eta$ over-penalizes confidence and hinders domain progression. On some datasets, the ranking based on regularized scores is sensitive to small $\eta$, suggesting that the high-ranked vanilla scores are concentrated, and a light regularization is sufficient to disperse them.
(3) As shown in Figure \ref{emb}, adjusting $\kappa$ enables flexible domain constructions under varying inter-domain distance $W_p(\mu_t, \mu_{t+1})$ while maintaining an inverse relationship between the average distance $\Delta$ and the number of domains $T$. Meanwhile, Figure \ref{hyperparameters} shows that the optimal GGDA domain path occurs at a "sweet spot" when $\kappa$, and thus $T$, is neither too small nor too large, which aligns with our theoretical analysis. 
(4) $\beta$ controls the pace of weight decay --- a smaller $\beta$ retains more weight on past-domain samples, and vice versa. 
Setting $\beta=0$ reduces GGDA to self-training over the full graph sequence, ignoring distribution shifts, while adopting an excessively large $\beta$ causes rapid information decay before full utilization, both of which can deteriorate transfer performance.

In practice, GGDA is robust to moderate hyperparameter variation.
For new transfer scenarios, we recommend starting with the following defaults: $K-1=4$, $\eta=0.4$, $\kappa=0.5$, and $\beta=3$. Intuitively, $K-1$ and $\kappa$ control adaptation granularity, $\eta$ safeguards selection reliability, and $\beta$ governs forgetting, so further tuning can be guided by observable dataset properties, such as graph heterophily, pseudo-label confidence and its fluctuations, and validation accuracy, etc.

\begin{figure}[t]
    \setlength{\tabcolsep}{0.3em}
        \begin{tabular}{ccc} 
        \includegraphics[height=85pt, trim=0cm 0cm 6cm 1cm,clip]{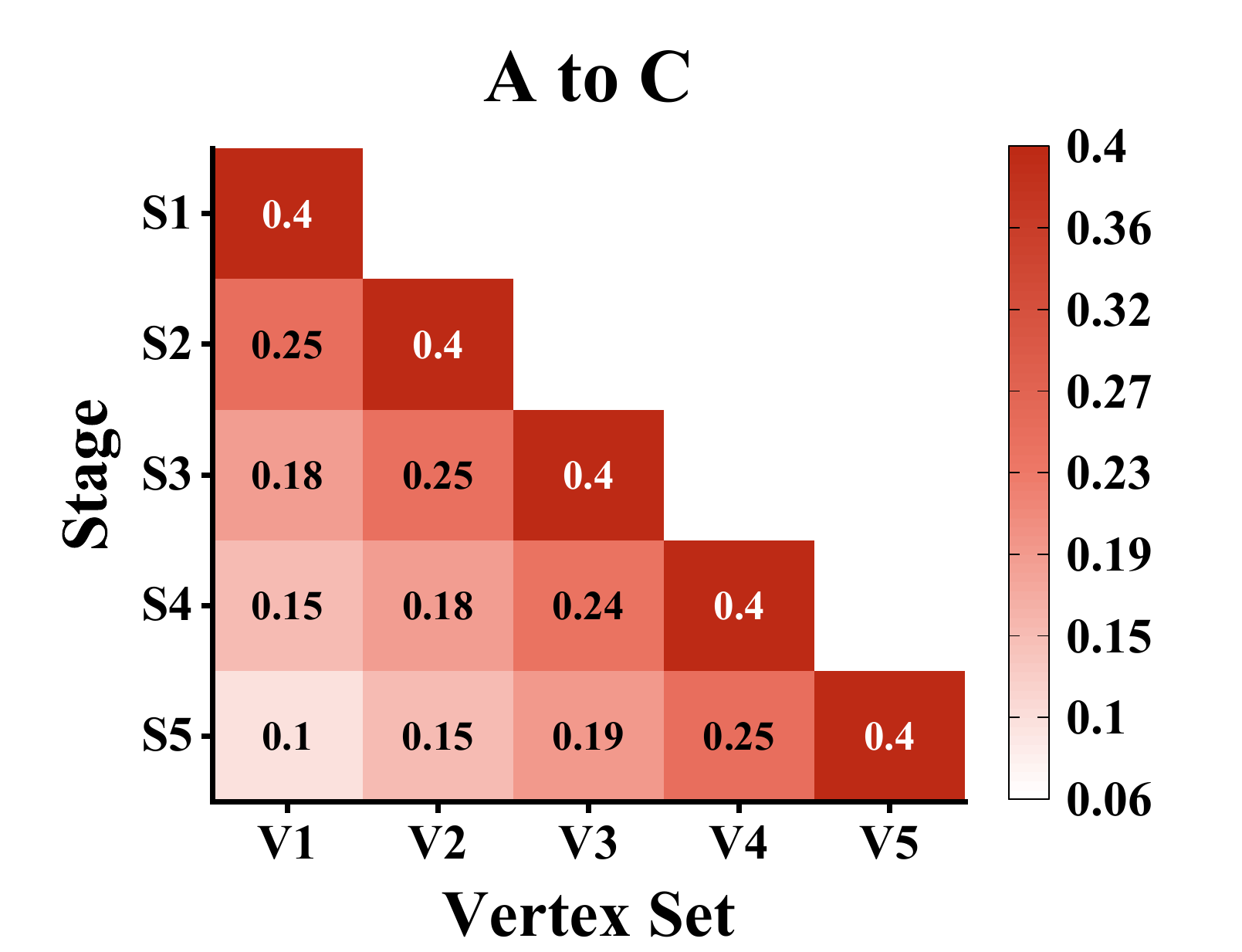} & 
        \includegraphics[height=85pt, trim=0cm 0cm 6cm 1cm,clip]{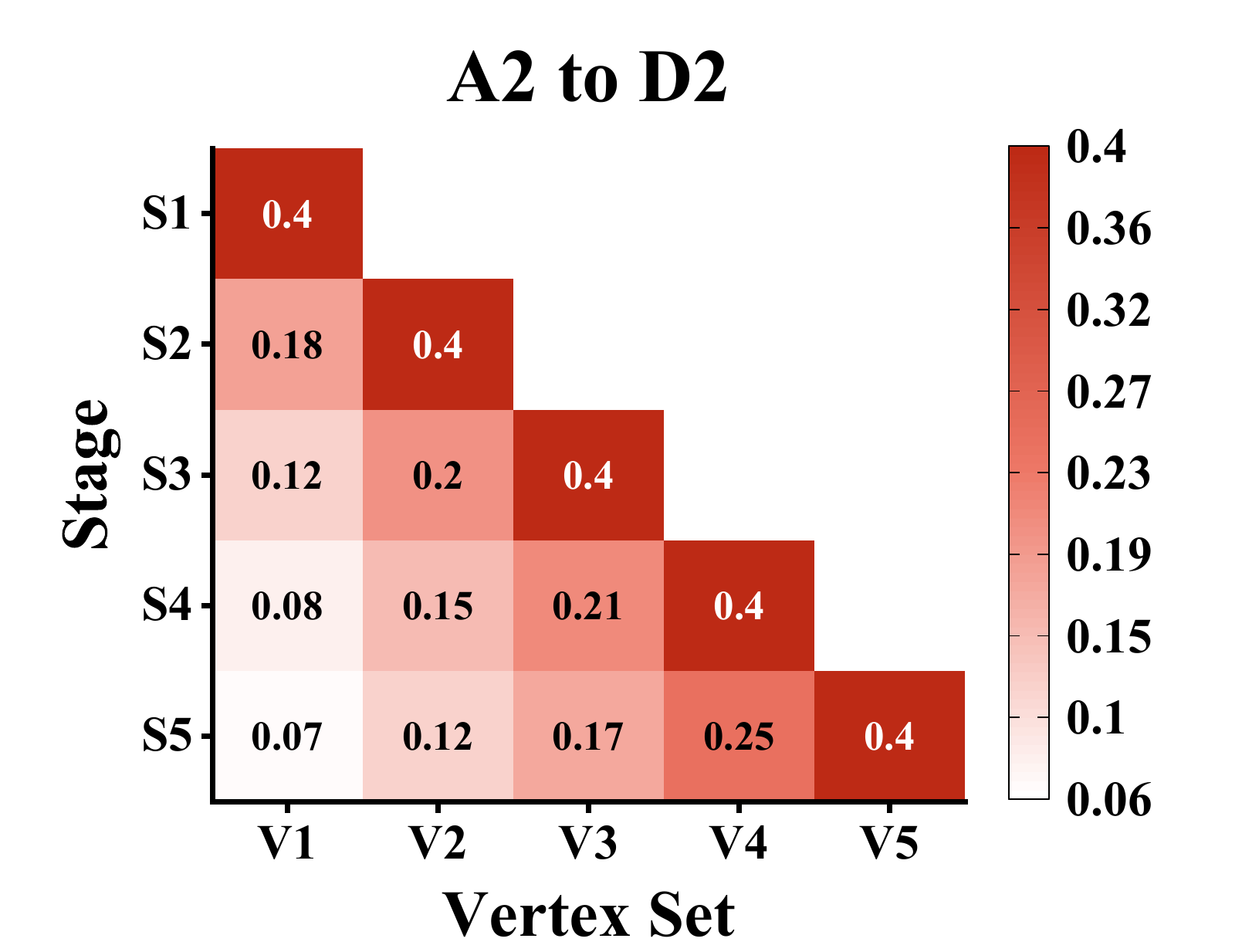} &
        \includegraphics[height=85pt, trim=0cm 0cm 0cm 1cm,clip]{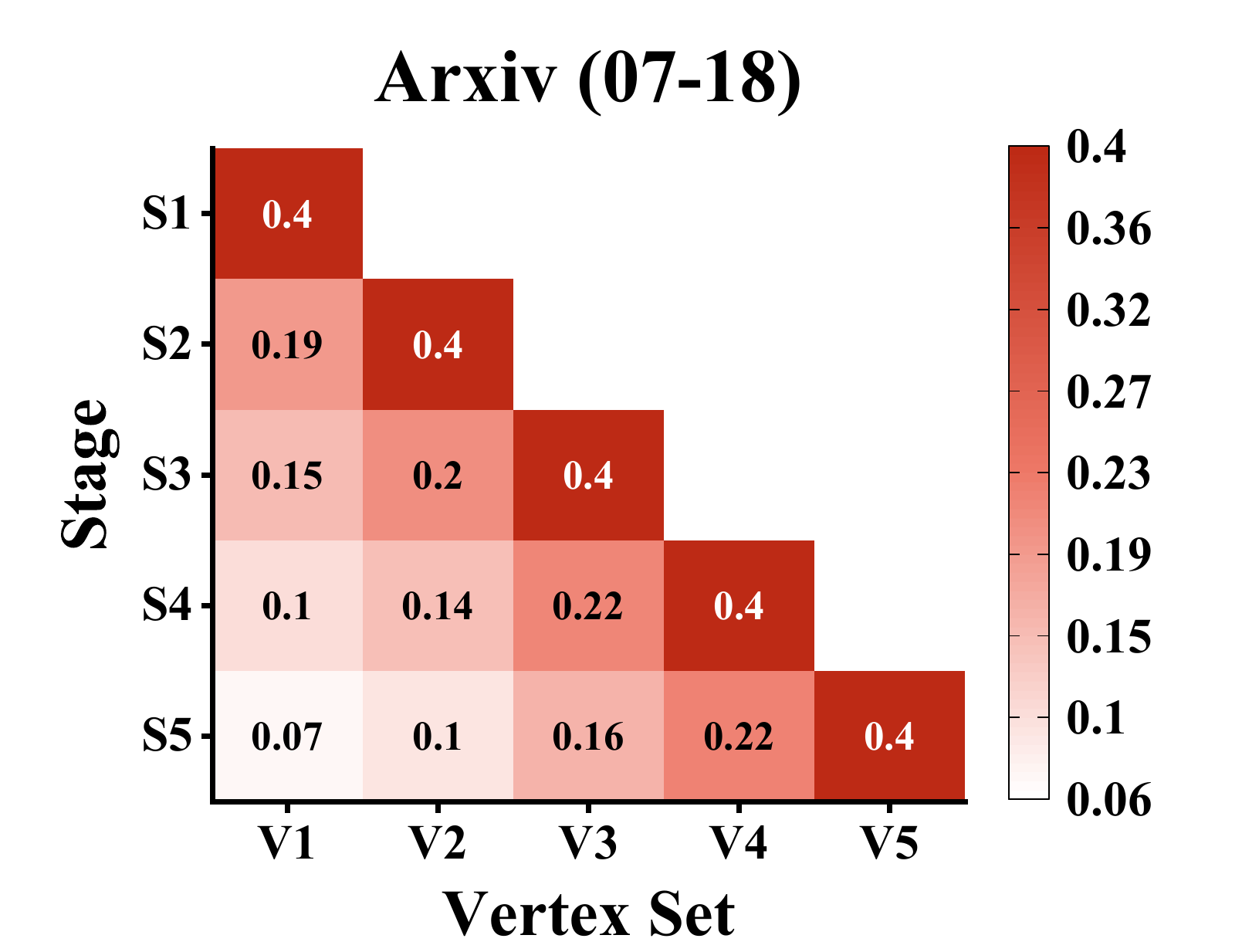}
        \end{tabular}
    \caption{Visualization of mass decay in each stage of GGDA. The x-axis represents the set of the top-weighted vertices constituting 40\% of each
domain distribution, and the y-axis represents the stage of GGDA. The color of the squares indicates the sum of weights over each vertex set in each stage.}
    \Description[Three lower-triangular heatmaps visualizing the cumulative mass decay of vertex weights across five stages of GGDA.]{Three side-by-side heatmaps for datasets A to C, A2 to D2, and Arxiv (07-18) illustrate weight distribution across stages S1–S5 (y-axis) and vertex sets V1–V5 (x-axis). Each plot follows a lower-triangular format where the diagonal cells consistently hold the maximum weight of 0.4. Moving left from the diagonal, the colors and numerical values fade progressively, reaching as low as 0.07. This visual gradient from right to left and top to bottom effectively demonstrates the cumulative decay of past domain importance as the model advances and shifts its focus toward new dominant nodes.}
    \label{decay}
\end{figure}

\subsection{Visualization of Adaptive Mass Decay}
Figure \ref{decay} demonstrates the effect of mass decay across GGDA stages. For a clearer interpretation, here we omit the operation that zeroes out lower-ranked weights in domain constructions. 
We observe a progressive fading of color from top to bottom and from right to left, indicating that the importance of past domains cumulatively decays as the model advances, and in each stage, the domain distribution shifts towards new dominant nodes. We also observe that consecutive vertex sets could carry higher weights in later stages. This is because the normalized $\eta$-penalty could slightly slow down the domain shift when approaching the target domain, resulting in more overlap between adjacent vertex sets. Notably, the bottom-left grid can be interpreted as the proportion of source samples that explicitly share common ego-graph properties with the final domain.

\begin{figure*}[b]
    \setlength{\tabcolsep}{-1pt}
        \begin{tabular}{ m{5.7em} m{17.2em} m{5.7em} m{25em}} 
       \begin{tabular}{@{}c@{}}\scriptsize{UDA-GCN:} \\ \scriptsize{ACCURACY = 62.7}\end{tabular} & \includegraphics[height=61.5pt, trim=0.4cm 0.3cm 0cm 0.1cm,clip]{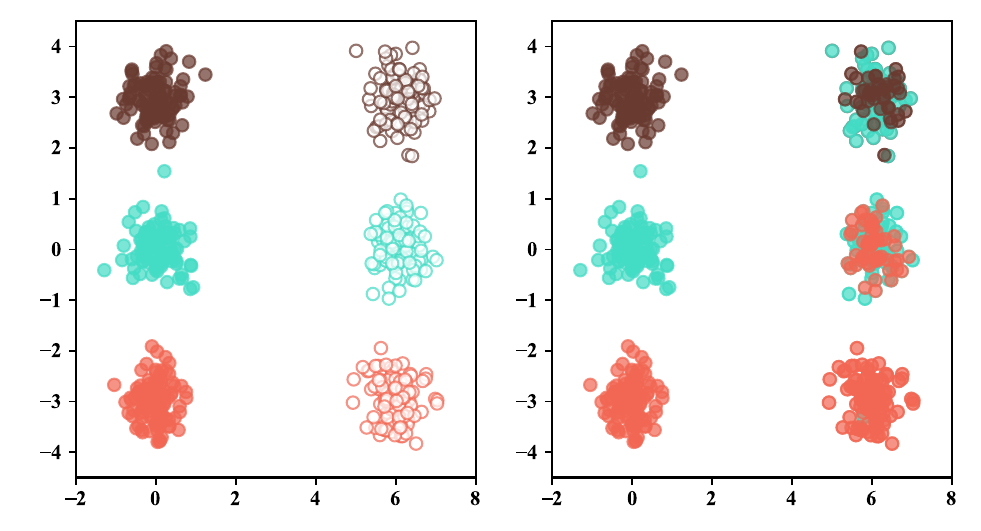} &
      \begin{tabular}{@{}c@{}}\scriptsize{TDSS:} \\ \scriptsize{ACCURACY = 77.5}\end{tabular} & \includegraphics[height=60pt, trim=0.1cm 0.1cm 0cm 0.1cm,clip]{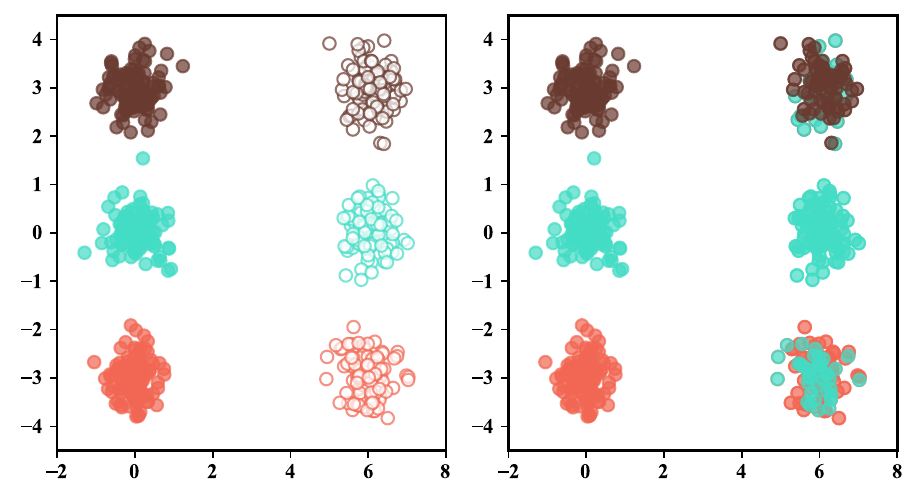} \\
       \begin{tabular}{@{}c@{}}\scriptsize{Direct ST:} \\ \scriptsize{ACCURACY = 33.8}\end{tabular} & \includegraphics[height=62pt, trim=0cm 0.3cm 0cm 0.1cm,clip]{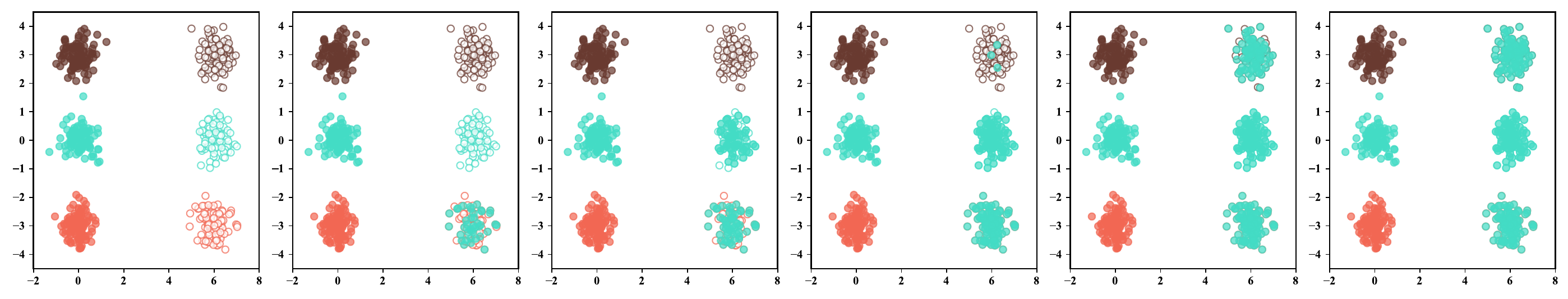}  \\
        \begin{tabular}{@{}c@{}}\scriptsize{GGDA-GCN:} \\ \scriptsize{ACCURACY = 95.4}\end{tabular} & \includegraphics[height=62.8pt, trim=0cm 0.3cm 0cm 0.1cm,clip]{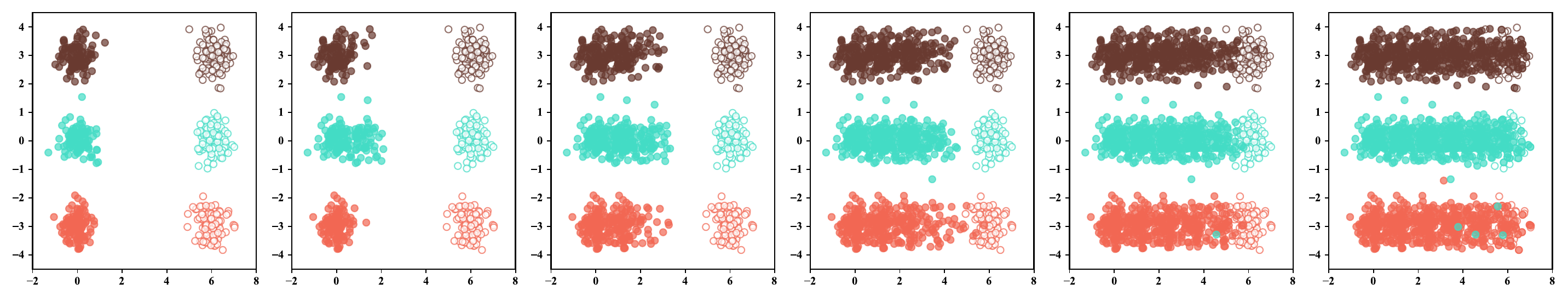}
        \end{tabular}
    \caption{Each plane represents a 2-dimensional space $\Omega_x$ (note that all computations are performed over the product space $\Omega_x \times \Omega_a$, i.e., with both features and structures on graph (non-IID), but the visualization is shown on $\Omega_x$ for clarity). Solid circles represent the ground-truth source labels or the assigned pseudo-labels (i.e., predictions) on each sample, whereas hollow circles represent the ground-truth target labels. Distinct colors indicate different classes. The assignments of labels throughout the training process are shown from left to right, with the leftmost planes showing the source and target labels before training. Specifically, UDA-GCN and TDSS employ single-stage training, while Direct ST and GGDA-GCN employ multi-stage training.}
    \Description[A grid of scatter plots visualizing the multi-stage training process and label assignments for four different models on a synthetic dataset.]{A large grid of 2D scatter plots compares UDA-GCN, TDSS, Direct ST, and GGDA-GCN. Rows represent models, and columns show the progression of label assignments from source to target. Solid circles denote source or pseudo-labels, while hollow circles denote ground-truth target labels. For UDA-GCN and TDSS (single-stage), the target nodes remain largely misaligned or mixed (with 62.7\% and 77.5\% accuracy). For Direct ST, early errors in pseudo-labeling propagate, leading to a cluttered final state with 33.8\% accuracy. In contrast, the GGDA-GCN row shows a smooth, iterative transition where clusters gradually shift and align. The final GGDA-GCN plot shows nearly perfect overlap between solid and hollow circles across three distinct horizontal bands, achieving the highest accuracy of 95.4\%.}
    \label{x_plot}
\end{figure*}

\subsection{Visualization of Knowledge Transfer in GGDA}
To delve into the process of knowledge transfer in GGDA, we conduct an experiment on a synthetic dataset generated using contextual stochastic block model (CSBM) \cite{deshpande2018contextual}, where node attributes and edges are generated independently given node labels. Specifically, we construct the source and target graphs to each contain 300 nodes, with 100 nodes from each of the 3 classes. We generate the node attributes for the three classes with Gaussians $\mathbb{P}_0=\mathcal{N}([0, -3], I/2)$, $\mathbb{P}_1=\mathcal{N}([0, 0], I/2)$, and $\mathbb{P}_2=\mathcal{N}([0, 3], I/2)$ for the source graph, and Gaussians $\mathbb{P}_0=\mathcal{N}([6, -3], I/2)$, $\mathbb{P}_1=\mathcal{N}([6, 0], I/2)$, and $\mathbb{P}_2=\mathcal{N}([6, 3], I/2)$ for the target graph. In both graphs, the intra-class edge probability is set to 0.1, while the inter-class edge probability is set to 0.02. To widen the distribution gap, we further modify the target graph by replacing 25\% of the randomly generated intra-class edges with intra-class edges that connect the top-k dissimilar nodes based on their attributes. 

Figure \ref{x_plot} demonstrates knowledge transfer in UDA-GCN, TDSS, Direct ST, and GGDA-GCN via the label assignments from training. While visualized in $\Omega_x$ for clarity, we emphasize that the actual computations operate over $\Omega_x \times \Omega_a$ with interactions between node features and local graph structures --- where the key complexity arises. The figure reveals that, through careful graph generation and iterative domain construction strategy, GGDA effectively creates an optimal domain path over the manifold. It regularizes transfer errors over consecutive domains, thus overcoming the significant distribution gap with minimal noise disturbance to allow smooth and accurate transition to the target graph. 

Knowledge distortion occurs with methods other than GGDA-GCN. For DIR models (i.e., UDA-GCN and TDSS), domain alignment forcibly matches the overall distributions while neglecting source-target correlations in the joint space, rendering them unreliable under a widened distribution gap. As for Direct ST, pseudo-labels are initially assigned to target nodes solely based on the source model, but with large domain discrepancies, many become inaccurate. The erroneous information propagates upon the target graph during subsequent training stages, impeding proper learning.

\begin{table}[b]
\footnotesize
    \caption{Comparison of per-epoch runtime on transfers A to C (18,295 nodes) and Arxiv\scalebox{0.7}{(07-18)} (125,720 nodes), with each model's average performance score across all datasets. OOM indicates out-of-memory errors.}
    \begin{tabular}{cccc}
        \toprule
        \multirow{2}{*}{\textbf{Model}}
                    &\textbf{Per-Epoch Runtime on} 
                    &\textbf{Per-Epoch Runtime on} 
                    &\textbf{Average Score across}\\
                 {}  & \textbf{{A to C} (seconds)} & \textbf{Arxiv\scalebox{0.7}{(07-18)} (seconds)} & \textbf{All Datasets} \\
        \midrule
        \textbf{CDAN} & {0.012} & {0.050} & 66.5\\
        \textbf{MDD} & {0.007} & {0.024} & 65.6\\
        \textbf{UDA-GCN} & {0.156} & {0.398} & 66.0\\
        \textbf{ACDNE} & {3.337} & {9.250} & 65.4\\
        \textbf{ASN} & {0.150} & {OOM} & 67.9\\
        \textbf{GRADE} & {0.006} & {OOM} & 61.9\\
        \textbf{StruRW} & {0.040} & {0.356} & 59.2\\
        \textbf{SpecReg} & {0.182} & {0.532} & 63.9\\
        \textbf{GraphAlign} & {2.026} & {7.114} & 61.1\\
        \textbf{A2GNN} & {0.077} & { 0.643 } & 66.8\\
        \textbf{Pair-Align} & {1.076} & {15.691} & 61.9\\
        \textbf{TDSS} & {0.191} & {OOM} & 67.7\\
        \midrule
        \textbf{GGDA-GCN} & {0.329} & 0.877 & 72.9\\
        \bottomrule
    \end{tabular}
    \label{per_epoch_runtime}
\end{table}

\section{Runtime Comparison \& Memory Requirements}
To assess the practical applicability of GGDA, we compare the per-epoch runtime of each method across two transfer scenarios: A to C (18,295 nodes) and Arxiv\scalebox{0.7}{(07-18)} (125,720 nodes, the largest dataset in our study). Runtime is measured by averaging over 1000 training epochs, and the results are summarized in Table \ref{per_epoch_runtime}. 

Overall, despite the non-generative nature of DIR models, they rely heavily on advanced encoder designs and regularizations, leading to costly computations such as eigendecomposition and topological matching. Consequently, ASN, GRADE, and TDSS fail to scale to larger graphs, encountering out-of-memory (OOM) errors, while ACDNE, GraphAlign, and Pair-Align remain feasible but incur high computational overhead. Notably, for GraphAlign, a data-centric method, graph generation is computationally intensive when the generated graph exceeds 1\% of the target graph size, i.e., a huge trade-off between time and performance. 

In contrast, our novel GGDA framework achieves a runtime comparable to more efficient baselines (with the time ratio of graph generation to domain progression maintained around 3.5:1 in both transfers) while delivering significantly improved target predictions. This favorable balance stems from our accelerated graph generation technique and the careful domain constructions that minimize transfer information loss. While GGDA is not the fastest in absolute terms, it offers an advantageous efficiency-performance trade-off, enabling scalable and accurate graph domain adaptation without prohibitive computational cost.

Regarding memory requirement, the peak memory consumption of GGDA on the largest transfer, Arxiv\scalebox{0.7}{(07-18)} (with $\sim$125K nodes), remains well within the capacity of modern GPUs (e.g., 24 GB memory available on an RTX 4090): 15.41 GB for GGDA-GCN, 17.49 GB for GGDA-GAT, and 15.70 GB for GGDA-SAGE. The primary memory demand occurs during vertex selection, when regularized scores are computed over the full generated vertex pool. For graphs with millions of nodes, practitioners can adopt straightforward mitigation strategies such as batch-wise vertex scoring or lightweight dimensionality reduction on node embeddings (e.g., PCA or random projection). Both approaches substantially reduce memory overhead without materially compromising selection accuracy. It is also worth noting that GGDA’s domain progression framework is agnostic to the graph encoder architecture. While we used GCN, GAT, and GraphSAGE in our experiments, more memory-efficient or scalable GNN variants (e.g., GraphSAINT \cite{zeng2019graphsaint} or SIGN \cite{frasca2020sign}) could be integrated to further reduce memory and runtime in large-scale industrial settings.

\section{Conclusion}

This work presents a pioneering effort to tackle the significant challenge of large inter-graph distribution shifts by establishing a novel graph gradual domain adaptation (GGDA) framework. Moving beyond the limitations of traditional domain-invariant methods and existing IID-focused gradual adaptation techniques, GGDA is specifically engineered for the complexities of non-IID graph-structured data.

Our solution is built on two foundational pillars. First, we introduce an efficient method for generating a knowledge-preserving intermediate data bridge by computing the weighted Fr\'echet mean over the Fused Gromov-Wasserstein (FGW) metric. This allows us to bypass the inherent implicitness of cross-domain graph structure space and generate intermediate graphs that faithfully integrate node features and structural relations from source and target graphs. Second, upon this bridge, we develop a novel vertex-based domain progression scheme. This scheme constructs an optimal adaptation path by strategically selecting "close" vertices and adaptively advancing the domain, thereby ensuring minimal information loss and maximizing the preservation of transferable, discriminative knowledge throughout the GGDA domain sequence.

Theoretical analysis forms a critical component of our contribution. We show that our framework concretizes $W_p(\mu_t,\mu_{t+1})$, the otherwise intractable inter-domain Wasserstein distance over graph space $\Omega_x \times \Omega_a \times \Omega_y$, by establishing implementable lower and upper bounds for it. The lower bound is derived from the FGW metric, while the upper bound is governed by our vertex selection regularization. This theoretical grounding not only provides a rigorous justification for our approach but also offers a flexible mechanism --- via a simple hyperparameter --- to adjust the inter-domain compactness and optimize the entire GGDA domain sequence without costly graph regenerations.

Extensive experiments validate the superiority of GGDA over state-of-the-art methods across diverse transfer scenarios. Looking forward, the GGDA framework opens up promising avenues for future research and could be extended to other complex non-IID learning scenarios, such as hypergraphs, dynamic graphs, and knowledge graphs.

\begin{acks}
This work was supported by the Science and Technology Development Fund and the University of Macau (Macau SAR: 0126/2024/RIA2, 001/2024/SKL; GDST: 2020B1212030003, 2023A0505030013; Nansha District: 2023ZD001; CG-2026; MYRG-GRG2023-00186-FST-UMDF and MYRG-GRG2025-00234-FST).
\end{acks}

\bibliographystyle{ACM-Reference-Format}
\bibliography{sample-base}


\appendix

\begingroup
\allowdisplaybreaks

\section{Solving Weighted Fr\'echet Mean over the FGW Metric}
Solving for an intermediate graph $\mu_I$ that minimizes the Fr\'echet mean objective over FGW is equivalent to finding the optimal tuple $(X_I, C_I, \pi(I,S), \pi(I,T))$, where $X_I \in \mathbb{R}^{n_I \times d}$ is the feature matrix of $\mu_I$, $C_I \in \mathbb{R}^{n_I \times n_I}$ is the structure matrix of $\mu_I$, and $\pi(I,\cdot)$ is the coupling between $\mu_I$ and the source or target graph. This can be efficiently solved by the block coordinate descent (BCD) algorithm \cite{ferradans2014regularized}. Specifically, in each iteration, we first fix $(X_I, C_I)$ and compute the two FGW distances by solving the optimal $(\pi(I,S), \pi(I,T))$ with conditional gradient (CG) \cite{jaggi2013revisiting, vayer2019}. Then, we fix $(\pi(I,S), \pi(I,T))$ and compute the optimal $(X_I, C_I)$ with exact analytical solutions \cite{peyre2016gromov, cuturi2014fast}. 

\section{Proof of Generalization Error for Graph Gradual Domain Adaptation}\label{proof1}
We will follow previous GDA work \cite{wang2022understanding} to adopt an online learning perspective and show that, in a graph context, the generalization bound for gradual domain adaptation is
\begin{gather*}
\begin{split}
&\epsilon_T(\theta_T) \leq \epsilon_0(\theta_0) + \mathcal{O}\bigg(T\Delta+T\Big(e + \sqrt{\frac{log(1/\delta')}{b}}\Big)\bigg)+\tilde{\mathcal{O}}\Big(\frac{1}{\sqrt{nT}}\Big),\\
&\text{where}~~ \Delta = \frac{1}{T}\sum_{t=0}^{T-1} W_{p}(\mu_t,\mu_{t+1})~~ \text{and} ~~ \mu_t = \sum_{i=1}^{n_t} h^t_i \delta_{(x^t_i,a^t_i,y^t_i)}.
\end{split}
\end{gather*}
\emph{Proof:} \\
Following the backbone of proof in \cite{wang2022understanding}, we start from the sequential learning perspective in \cite{kuznetsov2020discrepancy} and obtain that with probability at least $1-\delta$,
\begin{align} \label{eq_apd2} \tag{A.1}
\begin{split}
\epsilon_T(\theta_T) \leq &\sum_{t=0}^Tq_t \epsilon_t(\theta_T)+disc(q_{T+1})+||q_{T+1}||_2 + 6M\sqrt{4\pi logT}\mathfrak{R}_T^{seq}(\ell \circ \Phi)
+M||q_{T+1}||_2\sqrt{8log \frac{1}{\delta}},
\end{split}
\end{align}
where $q_t=(q_0,...,q_{t-1})$ is a t-dimensional probability vector, $disc(q_t)=sup_{\theta\in\Theta}\big(\epsilon_{t-1}(\theta)-\sum_{\tau=0}^{t-1}q_\tau \cdot \epsilon_\tau(\theta)\big)$ is a discrepancy measure, $\mathfrak{R}_T^{seq}(\cdot)$ is the sequential Rademacher complexity, and $\ell \circ \Phi$ is the composite of loss function and prediction model. 

Now, we first show that the following holds in our non-IID graph setting. Consider graph measures $\mu = \sum_{i=1}^n h_i \delta_{(x_i,a_i,y_i)}$ and $\nu = \sum_{j=1}^m h'_j \delta_{(x'_j,a'_j,y'_j)}$. Under Assumptions 2, 4 and 5, for any $\theta \in \Theta$, the population loss on $\mu$ and $\nu$ given by $M_\theta$ satisfies
\begin{gather}\label{eq_apd1}\tag{A.2}
|\epsilon_{\mu}(\theta) - \epsilon_{\nu}(\theta)| \leq \rho \sqrt{R^2+1} \cdot W_p(\mu,\nu).
\end{gather}
From the definition of population loss, we know that
\begin{align*}
&\qquad|\epsilon_{\mu}(\theta) - \epsilon_{\nu}(\theta)| \\
&= \Big|\mathbb{E}_{x,a,y \sim \mu}[\ell(M_{\theta}(x, a), y)] - \mathbb{E}_{x',a',y' \sim \nu}[\ell(M_{\theta}(x', a'), y')]\Big| \\
&= \Big|\sum_i \ell(M_{\theta}(x_i, a_i), y_i) \cdot h_i - \sum_j \ell(M_{\theta}(x'_j, a'_j), y'_j) \cdot h'_j\Big|.
\end{align*}
Let $\pi \in \Pi(h,h')$ be an arbitrary coupling. Then, under Assumption 2, we have
\begin{align*}
&\qquad|\epsilon_{\mu}(\theta) - \epsilon_{\nu}(\theta)| \\
&= \Big|\sum_i \Big[ \ell(M_{\theta}(x_i, a_i), y_i) \cdot \sum_j \pi_{i,j}\Big] - \sum_j \Big[ \ell(M_{\theta}(x'_j, a'_j), y'_j) \cdot \sum_i \pi_{i,j}\Big] \Big| \\
&= \sum_{i,j} \Big|\ell(M_{\theta}(x_i, a_i), y_i) - \ell(M_{\theta}(x'_j, a'_j), y'_j)\Big| \cdot \pi_{i,j} \\
&\leq \rho \sum_{i,j} \Big( | M_{\theta}(x_i, a_i) - M_{\theta}(x'_j, a'_j) | +
| y_i - y'_j | \Big) \pi_{i,j}\\
&\leq \rho \sum_{i,j} \Big( R \cdot d_{(x,a)}((x_i,a_i), (x'_j,a'_j)) + | y_i - y'_j |\Big)\pi_{i,j} \\ 
&\leq \rho \sqrt{R^2+1} \sum_{i,j} \Big(d_{(x,a)}((x_i,a_i), (x'_j,a'_j)) + | y_i - y'_j |\Big)\pi_{i,j}\\
&= \rho \sqrt{R^2+1} \sum_{i,j} d_{(x,a,y)}((x_i,a_i,y_i), (x'_j,a'_j,y'_j)) \pi_{i,j}. 
\end{align*}
The first inequality follows the triangle inequality and Assumption 5, and the second inequality follows Assumption 4. Now, since $\pi$ can be any coupling, we obtain
\begin{align*}
&\quad|\epsilon_{\mu}(\theta) - \epsilon_{\nu}(\theta)| \\
&\leq \rho \sqrt{R^2+1} \bigg[ \min_{\pi \in \Pi(h,h')} \sum_{i,j} d_{(x,a,y)}((x_i,a_i,y_i), (x'_j,a'_j,y'_j))\pi_{i,j}\bigg] \\
&= \rho \sqrt{R^2+1} \cdot W_1(\mu,\nu) \\
&\leq\rho \sqrt{R^2+1} \cdot W_p(\mu,\nu).
\end{align*} 

Having shown that Eq. \eqref{eq_apd1} is true in a graph scenario, we can adopt Lemma 2 from \cite{wang2022understanding} and obtain 
\begin{align}\label{eq_apd3}\tag{A.3}
disc(q_t^*) &\leq \rho \sqrt{R^2+1}/2\cdot (t\Delta)=\mathcal{O}(t\Delta),
\end{align}
where $q_t^*=(\frac{1}{t},...,\frac{1}{t})$. Now, by letting $q_{T+1} = q_{T+1}^*$ and using Eq. \eqref{eq_apd3}, we can transform Eq. \eqref{eq_apd2} into
\begin{align*}
\epsilon_T(\theta_T) \leq &\sum_{t=0}^Tq_t \epsilon_t(\theta_T)+\mathcal{O}(T\Delta)+\frac{1}{\sqrt{T+1}} + 6M\sqrt{4\pi logT}\mathfrak{R}_T^{seq}(\ell \circ \Phi)
+\frac{M}{\sqrt{T+1}}\sqrt{8log \frac{1}{\delta}},\\
\end{align*}
and by adopting a fine-grained view \cite{wang2022understanding}, it can be further transformed into
\begin{align}\label{eq_apd5}\tag{A.4}
\begin{split}
\epsilon_T(\theta_T) \leq &\sum_{t=0}^T\sum_{i=0}^{n-1}q_{nt+i} \epsilon_t(\theta_T)+\mathcal{O}(T\Delta)+\frac{1}{n\sqrt{T+1}} + 6M\sqrt{4\pi logT}\mathfrak{R}_T^{seq}(\ell \circ \Phi)
+\frac{M}{n\sqrt{T+1}}\sqrt{8log \frac{1}{\delta}}\\
\leq & \frac{1}{T+1}\sum_{t=0}^T\epsilon_t(\theta_T)+\mathcal{O}(T\Delta)+\mathcal{O}\Big(\frac{1}{\sqrt{nT}}\Big)+ 6M\sqrt{4\pi log(nT)}\mathfrak{R}_{nT}^{seq}(\ell \circ \Phi)
+\mathcal{O}\Big(\sqrt{\frac{log(1/\delta)}{nT}}\Big).
\end{split}
\end{align}

Next, we aim to obtain an upper bound for $\mathfrak{R}_{nT}^{seq}(\ell \circ \Phi)$ under our non-IID setting, which applies a graph encoder that takes as input an entire ego-graph. Define a $\mathcal{G}$-valued complete binary tree $g$ of depth $T$, where $g_t$ is the value of the node obtained by following the path of length $t-1$ from the root. The path is denoted by the sequence $r_t = (r_1, ..., r_{t-1}) \in \{\pm1\}^{t-1}$. Then, the sequential Rademacher complexity of a function class $\Phi$ on the tree is
\begin{align*}
\mathfrak{R}_{T}^{seq}(\Phi, g) \equiv \mathbb{E}_r\Bigg[\underset{\phi \in \Phi}\sup\frac{1}{T} \sum_{t=1}^T r_t \phi(g_t(r))\Bigg].
\end{align*}
Here we let $g$ be an ego-graph and $\Phi$ be the function class of graph neural networks (GNNs). This computation assumes each node in the graph to be assigned a scalar value, i.e., $\phi(g_t(r)) \in \mathbb{R}$. With a form analogous to the neural networks considered in \cite{bartlett2002rademacher}, each GNN layer can be written as
$\Phi^{(l)} = \Big\{g \rightarrow \sum_j w_j^{(l)} a_{j}^{(l)} \sigma\big(\phi_j^{(l-1)} (g)\big) \Big\}$,
where $l$ is the GNN layer, $w$ represents learnable parameters, $a_{j}^{(l)}$ represents the relations in the ego-graph (can be learnable), and $\sigma$ is the Lipschitz transfer function (e.g., sigmoid). Then we have
\begin{align*}
\begin{split}
&\mathfrak{R}_{T}^{seq}(\Phi^{(l)}) \\
=& \mathbb{E}_g\big[\mathfrak{R}_{T}^{seq}(\Phi^{(l)}, g)\big]\\
\leq& \underset{g}\sup\ \mathbb{E}_r\Bigg[\underset{\phi \in \Phi^{(l)}}\sup\frac{1}{T} \sum_{t=1}^T r_t \phi(g_t(r))\Bigg]\\
=& \frac{1}{T}\underset{g}\sup\ \mathbb{E}_r\Bigg[\underset{\phi \in \Phi^{(l-1)}}\sup \sum_{t=1}^T r_t \sum_j w_j^{(l)} a_{j}^{(l)} \sigma\big(\phi_j (g_t(r))\big)\Bigg]\\
=& \frac{1}{T}\underset{g}\sup\ \mathbb{E}_r\Bigg[\underset{\phi \in \Phi^{(l-1)}}\sup \sum_j w_j^{(l)} a_{j}^{(l)}\sum_{t=1}^T r_t  \sigma\big(\phi_j (g_t(r))\big)\Bigg]\\
\leq& \frac{1}{T}\underset{g}\sup\ \mathbb{E}_r\Bigg[\underset{\phi \in \Phi^{(l-1)}}\sup \Big|\Big|w^{(l)}a^{(l)}\Big|\Big|_1 \cdot \Big|\Big|\sum_{t=1}^T r_t  \sigma\big(\phi (g_t(r))\big)\Big|\Big|_{\infty}\Bigg]\\
\leq& \frac{1}{T}\underset{g}\sup\ \mathbb{E}_r\Bigg[\underset{\phi \in \Phi^{(l-1)}}\sup \sum_j \big|w_j^{(l)}\big| \cdot \big|a_{j}^{(l)}\big| \cdot\Big|\Big|\sum_{t=1}^T r_t  \sigma\big(\phi(g_t(r))\big)\Big|\Big|_{\infty}\Bigg]\\
\leq& \frac{1}{T}\underset{g}\sup\ \mathbb{E}_r\Bigg[\underset{\phi \in \Phi^{(l-1)}}\sup \Big|\Big|(w^{(l)})^+\Big|\Big|_1 \cdot \Big|\Big|(a^{(l)})^+\Big|\Big|_\infty \cdot\Big|\Big|\sum_{t=1}^T r_t  \sigma\big(\phi(g_t(r))\big)\Big|\Big|_{\infty}\Bigg].
\end{split}
\end{align*}
The second and fourth inequalities follow H\"older's inequality. As we have $||(w^{(l)})^+||_1 \leq B$ under regularization, and the graph relation satisfies $||(a^{(l)})^+||_\infty=max_j(|a_{j}^{(l)}|) \leq 1$, by assuming that $\sigma(\cdot)$ is $\iota$-Lipschitz, we obtain
\begin{align*}
\begin{split}
&\mathfrak{R}_{T}^{seq}(\Phi^{(l)}) \\
\leq& \frac{B}{T}\underset{g}\sup\ \mathbb{E}_r\Bigg[\underset{\phi \in \Phi^{(l-1)}}\sup \Big|\Big|\sum_{t=1}^T r_t  \sigma\big(\phi(g_t(r))\big)\Big|\Big|_{\infty}\Bigg]\\
=& \frac{B}{T}\underset{g}\sup\ \mathbb{E}_r\Bigg[\underset{\phi \in \Phi^{(l-1)}}\sup \underset{j} \max \Bigg|\sum_{t=1}^T r_t  \sigma\big(\phi_j(g_t(r))\big)\Bigg|\Bigg]\\
\leq& 2B\underset{g}\sup\ \mathbb{E}_r\Bigg[\underset{\phi \in \Phi^{(l-1)}}\sup \frac{1}{T} \sum_{t=1}^T r_t  \sigma\big(\phi(g_t(r))\big)\Bigg]\\
\leq& 16B\iota\Big(1+4\sqrt{2}log^{3/2}(eT^2)\Big)\underset{g}\sup\ \mathbb{E}_r\Bigg[\underset{\phi \in \Phi^{(l-1)}}\sup \frac{1}{T} \sum_{t=1}^T r_t  \phi(g_t(r))\Bigg]\\
=& 16B\iota\Big(1+4\sqrt{2}log^{3/2}(eT^2)\Big)\mathfrak{R}_{T}^{seq}(\Phi^{(l-1)}).
\end{split}
\end{align*}
Here the second inequality follows the observation that two mirror trees $g_t(r)$ and $g_t(-r)$ yield the same value, and the third inequality uses Corollary 5 from \cite{rakhlin2015online}. By recursive computation, for an L-layer GNN, we can get
\begin{align*}
\begin{split}
&\mathfrak{R}_{T}^{seq}(\Phi) \\
=&\mathfrak{R}_{T}^{seq}(\Phi^{(L)}) \\
\leq& \Big[16B\iota\Big(1+4\sqrt{2}log^{3/2}(eT^2)\Big)\Big]^{(L-1)}\mathfrak{R}_{T}^{seq}(\Phi^{(1)})\\
\leq& \Big[16B\iota\Big(1+4\sqrt{2}log^{3/2}(eT^2)\Big)\Big]^{(L-1)}\underset{g}{\sup} \Bigg(\bar{\alpha}+\sqrt{\frac{2log\mathcal{N}(\bar{\alpha},\Phi^{(1)},g)}{T}}\Bigg)\\
\leq& \Big[16B\iota\Big(1+4\sqrt{2}log^{3/2}(eT^2)\Big)\Big]^{(L-1)}\underset{g}{\sup} \Bigg(\bar{\alpha}+\sqrt{\frac{2log \big((2eT/\bar{\alpha})^{\bar{d}} \big)}{T}}\Bigg)\\
=& \mathcal{O}\Bigg(\sqrt{\frac{(logT)^{3L-2}}{T}}\Bigg).
\end{split}
\end{align*}
The second inequality follows the bound from Lemma 3 in \cite{rakhlin2015sequential}, and the third inequality follows Corollary 6 in \cite{rakhlin2015sequential}. By applying Corollary 5 from \cite{rakhlin2015online} again, we get
\begin{align*}
\begin{split}
\mathfrak{R}_{T}^{seq}(\ell \circ \Phi)
\leq 8\rho\Big(1+4\sqrt{2}log^{3/2}(eT^2)\Big) \mathcal{O}\Bigg(\sqrt{\frac{(logT)^{3L-2}}{T}}\Bigg)
= \mathcal{O}\Bigg(\sqrt{\frac{(logT)^{3L+1}}{T}}\Bigg).
\end{split}
\end{align*}
Finally, from a fine-grained view, we get
\begin{gather}\label{eq_apd4}\tag{A.5}
\mathfrak{R}_{nT}^{seq}(\ell \circ \Phi) = \mathcal{O}\Bigg(\sqrt{\frac{(log(nT))^{3L+1}}{nT}}\Bigg).
\end{gather}
Plugging Eq. \eqref{eq_apd4} into Eq. \eqref{eq_apd5}, we have
\begin{align}\label{eq_apd7}\tag{A.6}
\begin{split}
\epsilon_T(\theta_T) 
\leq & \frac{1}{T+1}\sum_{t=0}^T\epsilon_t(\theta_T)+\mathcal{O}(T\Delta)+\mathcal{O}\Big(\frac{1}{\sqrt{nT}}\Big) + \mathcal{O}\Bigg(\sqrt{\frac{(log(nT))^{3L+2}}{nT}}\Bigg)
+\mathcal{O}\Big(\sqrt{\frac{log(1/\delta)}{nT}}\Big).
\end{split}
\end{align}
To handle the term $\frac{1}{T+1}\sum_{t=0}^T\epsilon_t(\theta_T)$, we follow the structure in \cite{wang2022understanding} but in a graph setting. Denote the population risk by $\epsilon$ and the empirical risk by $\hat{\epsilon}$. Denote the composite of loss function and prediction model by $\mathcal{A} = \ell \circ \Phi$. Under Assumptions 2, 3, 4 and 5, with probability at least $1-\delta$, we have
\begin{align*}
& \quad \epsilon_{t+1}({\theta}_{t+1}) \\
&\leq \hat{\epsilon}_{t+1}({\theta}_{t+1}) + \mathcal{O} \big(\mathfrak{R}_b(\mathcal{A}) + e \big) + M\sqrt{\frac{log(2/\delta')}{8b}}\\
&\approx \hat{\epsilon}_{t+1}({\theta}_{t}) + \mathcal{O} \big(\mathfrak{R}_b(\mathcal{A}) + e \big) + M\sqrt{\frac{log(2/\delta')}{8b}}\\
&\leq {\epsilon}_{t+1}({\theta}_{t}) + 2\mathcal{O} \big(\mathfrak{R}_b(\mathcal{A}) + e \big) + 2M\sqrt{\frac{log(2/\delta')}{8b}}\\
&\leq {\epsilon}_{t}({\theta}_{t}) + \rho \sqrt{R^2+1} \cdot 
W_p(\mu_t,\mu_{t+1}) + 2\mathcal{O} \big(\rho\mathfrak{R}_b(\Phi) + e \big)+ 2M\sqrt{\frac{log(2/\delta')}{8b}}\\
&\leq {\epsilon}_{t}({\theta}_{t}) + \mathcal{O} \bigg( W_p(\mu_t,\mu_{t+1}) + e + \frac{d_{max}}{\sqrt{n_b}} + \sqrt{\frac{log(1/\delta')}{b}}\bigg),\\
\end{align*}
where $\delta'=\delta-2(b-1)\beta(n_b)$. Here we view each ego-graph as a sample and a graph as a non-stationary mixing process. Then, for the first two inequalities, we follow \cite{kuznetsov2017generalization} --- by partitioning the graph into blocks (i.e., subgraphs formed by sets of connected ego-graphs), the bound is computed by viewing distant blocks as independent (but intra-block dependency still exists). Specifically, $b$ is the number of subgraphs, $n_b$ is the number of ego-graphs in each subgraph, $\beta(n_b)$ is the dependency between ego-graphs separated by $n_b$ steps, and $M$ is a constant bound on the hypotheses set $\mathcal{A}$. The term $(b-1)\beta(n_b)$ represents the potential error by assuming such inter-subgraph independency. With smaller $n_b$ (greater dependency between subgraphs), $b$ and $\beta(n_b)$ are bigger and the bound is looser. With constant $n_b$ but magnifying $\beta$ function (larger dependency between $n_b$-hop nodes), the bound is looser as well. The term $\mathfrak{R}_b(\mathcal{A})$ represents the complexity for intra-subgraph modeling. As for $e$, it represents the largest possible difference in error that any model $\theta \in \Theta$ could produce when applied to two ego-graphs from the same graph, which is affected by the degree of non-IIDness in the underlying data.

In addition, the approximation ($\approx$) above follows the fact that sharpened pseudo labels are used for the next round of training (it becomes equality if label-sharpening is not applied). The third inequality follows Eq. \eqref{eq_apd1}, Talagrand's Lemma, and Assumption 5. The fourth inequality follows Assumption 3. Now by recursion and combining the fact that $T\Delta = \sum_{t=0}^{T-1} W_{p}(\mu_t,\mu_{t+1})$, we obtain the bound
\begin{align}\label{eq_apd6}\tag{A.7}
\frac{1}{T+1}\sum_{t=0}^T\epsilon_t(\theta_T) \leq \epsilon_0(\theta_0) + \mathcal{O}(T\Delta) + \mathcal{O}\Big(Te + T\sqrt{\frac{log(1/\delta')}{b}}\Big).
\end{align}
Plugging Eq. \eqref{eq_apd6} into Eq. \eqref{eq_apd7}, we get
\begin{align*}
\begin{split}
&\epsilon_T(\theta_T) \\
\leq & \epsilon_0(\theta_0) + \mathcal{O}(T\Delta) + \mathcal{O}\Big(Te + T\sqrt{\frac{log(1/\delta')}{b}}\Big)+\mathcal{O}(T\Delta)+\mathcal{O}\Big(\frac{1}{\sqrt{nT}}\Big) \\  &+ \mathcal{O}\Bigg(\sqrt{\frac{(log(nT))^{3L+2}}{nT}}\Bigg)
+\mathcal{O}\Big(\sqrt{\frac{log(1/\delta)}{nT}}\Big)\\
=& \epsilon_0(\theta_0) + \mathcal{O}\bigg(T\Delta+T\Big(e + \sqrt{\frac{log(1/\delta')}{b}}\Big)\bigg)+\tilde{\mathcal{O}}\Big(\frac{1}{\sqrt{nT}}\Big).\qed
\end{split}
\end{align*}

\section{Proof of Proposition 2}\label{proof2}
\emph{Restatement of Proposition 2:} For graph measures $\mu(x,a,y) = \sum_{i=1}^n h_i \delta_{(x_i,a_i,y_i)}$ and $\nu(x',a',y') = \sum_{j=1}^m h'_j \delta_{(x'_j,a'_j,y'_j)}$, along with their marginal measures $\mu(x,a) = \sum_{i=1}^n h_i \delta_{(x_i,a_i)}$ and $\nu(x',a') = \sum_{j=1}^m h'_j \delta_{(x'_j,a'_j)}$,
\begin{gather*}
W_{p}\big(\mu(x,a,y),\nu(x',a',y')\big) \geq FGW_{\alpha,p,1}\big(\mu(x,a),\nu(x',a')\big)/2.
\end{gather*}
\emph{Proof:} \\
Let $\pi \in \Pi(h,h')$ be an arbitrary coupling. Then under Assumption 2, we have
\begin{align*}
&\qquad FGW_{\alpha,p,1}(\mu(x,a),\nu(x',a'))\\
&= \Big\{ \min_{\pi \in \Pi(h,h')}\sum_{i,j,k,l} \Big[(1-\alpha) d_x(i,j)+ \alpha |C(i,k)-C'(j,l)|\Big]^p \pi_{i,j} \pi_{k,l}\Big\}^{\frac{1}{p}}\\
&\leq \Big\{ \sum_{i,j,k,l} \Big[(1-\alpha) d_x(i,j) + \alpha |C(i,k)-C'(j,l)|\Big]^p \pi_{i,j} \pi_{k,l}\Big\}^{\frac{1}{p}}\\
&\leq \Big\{ \sum_{i,j,k,l} \Big[(1-\alpha) d_x(i,j) + \alpha (C(i,j)+C'(k,l))\Big]^p \pi_{i,j} \pi_{k,l}\Big\}^{\frac{1}{p}}\\
&\leq \Big\{ \sum_{i,j,k,l} \Big[(1-\alpha) d_x(i,j) + \alpha C(i,j) +(1-\alpha) d_x(k,l) +\alpha C'(k,l)\Big]^p \pi_{i,j} \pi_{k,l}\Big\}^{\frac{1}{p}}\\
&\leq \Big\{ \sum_{i,j} \Big[(1-\alpha) d_x(i,j) + \alpha C(i,j)\Big]^p \pi_{i,j} \Big\}^{\frac{1}{p}}  + \Big\{ \sum_{k,l} \Big[(1-\alpha) d_x(k,l) + \alpha C'(k,l)\Big]^p \pi_{k,l} \Big\}^{\frac{1}{p}}\\
&= 2\cdot \Big\{ \sum_{i,j} \Big[(1-\alpha) d_x(i,j) + \alpha C(a_i,a_j)\Big]^p \pi_{i,j} \Big\}^{\frac{1}{p}} \\
&= 2 \cdot \Big( \sum_{i,j} d_{(x,a)}((x_i,a_i), (x'_j,a'_j))^p \pi_{i,j} \Big)^{\frac{1}{p}} \\
&\leq 2 \cdot \Big[\sum_{i,j} \Big(d_{(x,a)}((x_i,a_i), (x'_j,a'_j)) + |y_i - y'_j|\Big)^p \pi_{i,j} \Big]^{\frac{1}{p}}\\
&= 2 \cdot \Big[\sum_{i,j} d_{(x,a,y)}((x_i,a_i,y_i), (x'_j,a'_j,y'_j))^p \pi_{i,j} \Big]^{\frac{1}{p}}.
\end{align*}
The second inequality follows the triangle inequality, and the fourth inequality follows Minkowski's inequality. Since $\pi$ can be any coupling, we obtain
\begin{align*}
& \qquad FGW_{\alpha,p,1}(\mu(x,a),\nu(x',a'))\\
&\leq 2 \cdot \Big[ \min_{\pi \in \Pi(h,h')} \sum_{i,j} d_{(x,a,y)}((x_i,a_i,y_i), (x'_j,a'_j,y'_j))^p \pi_{i,j} \Big]^{\frac{1}{p}}\\
&= 2 \cdot W_{p}(\mu(x,a,y),\nu(x',a',y')). \qed
\end{align*}

\section{Proof of Proposition 3}\label{proof3}
\emph{Restatement of Proposition 3:} Consider graph domains $\mu = \sum_{i=1}^n h_i \delta_{(x_i,a_i,y_i)}$ and $\nu = \sum_{j=1}^m h'_j \delta_{(x'_j,a'_j,y'_j)}$ with uniform weights $h$ and $h'$. Let $\mathcal{E}$ be the set of unit-weight edges linking $\mu$ and $\nu$ and $\tilde{\pi}$ be an arbitrary coupling satisfying $\tilde{\pi}_{i,j} = \frac{1}{nm} \mathds{1}_{\{(i,j) \in \mathcal{E}\}}$. For any $\theta \in \Theta$, the population loss on $\mu$ and $\nu$ given by $M_\theta$ satisfies
\begin{gather*}
\begin{split}
&~~~~~~~|\epsilon_{\mu}(\theta) - \epsilon_{\nu}(\theta)| 
\leq \rho(\zeta_1 + \zeta_2), \qquad\text{where} \\ 
\zeta_1 =& \frac{\sqrt{\tilde{R}^2+1}}{n \cdot m} \sum_{(i,j)\in \mathcal{E}}\Big(\lVert z_i(\theta) - z'_j(\theta) \rVert + | y_i - y'_j |\Big), \\
\zeta_2 =& \sqrt{R^2+1} \cdot  \bigg[\min_{\tilde{\pi}}\bigg(\sum_{(i,j)\notin \mathcal{E}} d_{(x,a,y)}\Big((x_i,a_i,y_i), (x'_j,a'_j,y'_j)\Big)^p
\cdot \tilde{\pi}_{i,j}  \bigg)\bigg]^{\frac{1}{p}}.
\end{split}
\end{gather*}
\emph{Proof:} \\
Let $\pi \in \Pi(h,h')$ be an arbitrary coupling, and $\tilde{w}_{i,j}=\frac{w_{i,j}}{n \cdot m}$ where $w_{i,j}=1$ if $(i,j) \in \mathcal{E}$ and $0$ otherwise. Then we obtain
\begin{align*}
&\quad |\epsilon_{\mu}(\theta) - \epsilon_{\nu}(\theta)| \\
&\leq \rho \sum_{i,j} \big( | M_{\theta}(x_i, a_i) - M_{\theta}(x'_j, a'_j) | +
| y_i - y'_j | \big) \pi_{i,j}\\
&=\rho \Bigg[\sum_{i,j}\bigg(\big|\Psi_\theta(z_i) - \Psi_\theta(z'_j)\big| + | y_i - y'_j |\bigg)\tilde{w}_{i,j} \\
&\quad +\sum_{i,j}\bigg(\big|M_{\theta}(x_i, a_i) - M_{\theta}(x'_j, a'_j)\big| + | y_i - y'_j | \bigg)(\pi_{i,j} - \tilde{w}_{i,j}) \Bigg]\\
&\leq\rho \Bigg[ \sum_{i,j}\bigg(\tilde{R} \lVert z_i(\theta) - z'_j(\theta) \rVert + | y_i - y'_j |\bigg)\tilde{w}_{i,j} \\
&\quad + \sum_{i,j}\bigg(R \cdot d_{(x,a)}((x_i,a_i), (x'_j,a'_j)) + | y_i - y'_j | \bigg)(\pi_{i,j} - \tilde{w}_{i,j}) \Bigg],
\end{align*}
where the first inequality follows the proof of Eq. \eqref{eq_apd1}, and the second inequality follows Assumption 4. Since $\pi$ can be any coupling, we have
\begin{align*}
&\quad |\epsilon_{\mu}(\theta) - \epsilon_{\nu}(\theta)| \\
&\leq\rho \Bigg\{  \sqrt{\tilde{R}^2+1} \sum_{i,j}\bigg(\lVert z_i(\theta) - z'_j(\theta) \rVert + | y_i - y'_j |\bigg)\tilde{w}_{i,j} \\
& \quad + \sqrt{R^2+1} \cdot \min_{\tilde{\pi}} \bigg[\sum_{i,j} d_{(x,a,y)}((x_i,a_i,y_i), (x'_j,a'_j,y'_j)) \cdot (\tilde{\pi}_{i,j} - \tilde{w}_{i,j}) \bigg]\Bigg\}\\
&\leq \rho \Bigg\{  \frac{\sqrt{\tilde{R}^2+1}}{n \cdot m} \sum_{(i,j)\in \mathcal{E}}\Big(\lVert z_i(\theta) - z'_j(\theta) \rVert + | y_i - y'_j |\Big) \\
& \quad + \sqrt{R^2+1} \cdot \bigg[\min_{\tilde{\pi}} \bigg(\sum_{(i,j)\notin \mathcal{E}} d_{(x,a,y)}((x_i,a_i,y_i), (x'_j,a'_j,y'_j))^p \cdot \tilde{\pi}_{i,j}  \bigg)\bigg]^{\frac{1}{p}}\Bigg\},
\end{align*}
where the last inequality follows H\"older's inequality. $\qed$

\section{Proof of Proposition 4}\label{proof4}
\emph{Restatement of Proposition 4} Let $\Phi_{\tilde\theta}: \Omega_x \times \Omega_a \rightarrow \Omega_z$ be an $(L,\gamma)$-quasiisometric graph embedding \cite{dructu2018geometric}. Let $\mu(z|\tilde\theta) = \sum_{i=1}^n h_i \delta_{z_i|\tilde\theta}$, $\nu(z'|\tilde\theta) = \sum_{j=1}^m h'_j \delta_{z'_j|\tilde\theta}$, and $\pi^* = \argmin_{\pi \in \Pi(h,h')}\sum_{i,j} d_{z}(z_i|\tilde\theta, z'_j|\tilde\theta)^p \pi_{i,j}$. Then we have
\begin{align*}
\begin{split}
&W_{p}\big(\mu(x,a,y),\nu(x',a',y')\big) \leq \xi
\text{~and~} 
|\epsilon_{\mu}(\theta) - \epsilon_{\nu}(\theta)| \cdot \mathcal{O}(\rho^{-1}) \leq \xi,\\ 
&\text{where } \xi = \Big[W_{\infty}\big(\mu(z|\tilde\theta),\nu(z'|\tilde\theta)\big) + \gamma  + \frac{1}{L}\bigg( \sum_{i,j} |y_i-y'_j|^p \pi_{i,j}^*\bigg)^{\frac{1}{p}}\Big].
\end{split}
\end{align*}
\emph{Proof:} \\
A map $f: X \rightarrow Y$ is an $(L,\gamma)$-quasiisometric embedding if \cite{dructu2018geometric}
\begin{align*}
L^{-1} \cdot d_X(x,x')-\gamma \leq d_Y(f(x), f(x')) \leq L \cdot d_X(x,x') + \gamma, \quad
\forall x, x' \in X.
\end{align*}
Let $W_{p}(\mu(x,a,y),\nu(x',a',y'))$ and $W_{p}(\mu(z|\tilde\theta),\nu(z'|\tilde\theta))$ be the p-Wasserstein distances between graphs $\mu$ and $\nu$ over the space $\Omega_x \times \Omega_a \times \Omega_y$ and $\Omega_z$. Let $\pi \in \Pi(h,h')$ be an arbitrary coupling. Then we have
\begin{align*}
& \quad W_{p}(\mu(x,a,y),\nu(x',a',y')) \\
=& \min_{\pi \in \Pi(h,h')} \sum_{i,j} d_{(x,a,y)}((x_i,a_i,y_i), (x'_j,a'_j,y'_j))^p \pi_{i,j} \\
\leq& \sum_{i,j} \Big(d_{(x,a)}((x_i,a_i), (x'_j,a'_j)) + |y_i-y'_j|\Big)^p \pi_{i,j} \\
\leq& \bigg[ \sum_{i,j} \Big(L \cdot d_{z}(z_i|\tilde\theta, z'_j|\tilde\theta) + L\gamma \Big)^p \pi_{i,j}\bigg]^{\frac{1}{p}} + 
\bigg( \sum_{i,j} |y_i-y'_j|^p \pi_{i,j}\bigg) \\
\leq& \bigg[ \sum_{i,j} \Big(L \cdot d_{z}(z_i|\tilde\theta, z'_j|\tilde\theta)\Big)^p \pi_{i,j}\bigg]^{\frac{1}{p}} + 
\bigg[ \sum_{i,j} (L\gamma)^p \pi_{i,j} \bigg] ^{\frac{1}{p}} + \bigg( \sum_{i,j} |y_i-y'_j|^p \pi_{i,j}\bigg)^{\frac{1}{p}} \\
\leq& L \Bigg[\bigg( \sum_{i,j} d_{z}(z_i|\tilde\theta, z'_j|\tilde\theta)^p \pi_{i,j}\bigg)^{\frac{1}{p}} + 
\gamma + \frac{1}{L}\bigg( \sum_{i,j} |y_i-y'_j|^p \pi_{i,j}\bigg)^{\frac{1}{p}}\Bigg].
\end{align*}
The second and third inequalities follow Minkowski's inequality. Since $\pi$ can be any coupling, by letting $\pi^*$ be the one that minimizes $\sum_{i,j} d_{z}(z_i|\tilde\theta, z'_j|\tilde\theta)^p \pi_{i,j}$, we obtain
\begin{align*}
& \quad W_{p}(\mu(x,a,y),\nu(x',a',y')) \\
&\leq L \Bigg[\bigg(\sum_{i,j} d_{z}(z_i|\tilde\theta, z'_j|\tilde\theta)^p \pi_{i,j}^*\bigg)^{\frac{1}{p}} + 
\gamma + \frac{1}{L}\bigg( \sum_{i,j} |y_i-y'_j|^p \pi_{i,j}^*\bigg)^{\frac{1}{p}}\Bigg] \\
&\leq L \Big[W_{p}\big(\mu(z|\tilde\theta),\nu(z'|\tilde\theta)\big) + \gamma  + \frac{1}{L}\bigg( \sum_{i,j} |y_i-y'_j|^p \pi_{i,j}^*\bigg)^{\frac{1}{p}}\Big] \\
&\leq L \Big[W_{\infty}\big(\mu(z|\tilde\theta),\nu(z'|\tilde\theta)\big) + \gamma  + \frac{1}{L}\bigg( \sum_{i,j} |y_i-y'_j|^p \pi_{i,j}^*\bigg)^{\frac{1}{p}}\Big],
\end{align*}
and $|\epsilon_{\mu}(\theta) - \epsilon_{\nu}(\theta)| \cdot \mathcal{O}(\rho^{-1}) \leq \xi$ follows from Eq. \eqref{eq_apd1}. $\qed$

\section{Further Discussion on Assumptions}
GGDA is built upon several foundational assumptions (Section 4 in the main text). While these are standard in domain adaptation theory and graph representation learning, we recognize that in practice, real-world graphs may not satisfy them strictly. Below, we discuss the practical impact of potential violations:

\paragraph{(Assumption 1)} Covariate shift is a standard assumption in unsupervised domain adaptation, and most existing methods are designed for this type of distribution gap rather than prior or concept shift \cite{farahani2021brief}. While GGDA is framed under this assumption, our domain progression is robust to its moderate violation. In cases where $p(y|x,a)$ shifts gradually, GGDA can still adapt by prioritizing vertices whose local structural and feature patterns remain stable across nearby domains, thereby preserving discriminative knowledge. For more pronounced shifts, we recommend tuning $\kappa$ to construct finer-grained intermediate steps, which promotes smoother adaptation and helps recover pseudo-label stability. This adjustability, combined with GGDA’s inherent reliance on locality and structure, makes the framework applicable even when the covariate shift assumption is not strictly satisfied.

\paragraph{(Assumption 2)}
Mathematically, the linear combination assumption is not an arbitrary simplification of the data distribution.
Unlike a hand‑crafted non-linear metric combination $f(d_x, C)$ that relies on unprovable assumptions about the exact interaction between features and structures, the linear combination is the canonical way to construct a metric on the product space $\Omega_X \times \Omega_A$: it guarantees that two nodes are considered "close" if and only if they are proximate in both features and structure, and, most importantly, it is mathematically rigorous by always yielding a true metric (whereas an arbitrary non-linear combination could risk violating the triangle inequality and cause the entire theoretical foundation to collapse). 



Note that our framework does not ignore the complex, non-linear underlying coupling between features and topology. On one hand, this coupling is dynamically captured by the GNN encoder, which acts as a data-driven Lipschitz continuous pushforward (notably, our theory does not restrict the encoder to be linear). 
On the other hand, during data generation, we adopt the squared cost setting (i.e., $q=2$), which yields a smoother optimization landscape and promotes more stable convergence \cite{courty2016optimal}, while the outer exponent ensures it remains a proper mathematical metric. 
The theoretical gap between $q=1$ and $q=2$ is bridged by the topological equivalence of different $q$-norms for compact spaces (with normalized graph features and structures) --- they induce identical convergence behavior, albeit with different constants (i.e., $\sqrt{a^2 + b^2} \leq a + b \leq \sqrt{2} \sqrt{a^2 + b^2}$) and potentially different convergence rates. 
This ensures that the theoretical integrity of our framework is fully preserved in practice. Overall, the successful learning of the underlying feature-structure relations is strongly validated by our extensive empirical GGDA results.

\paragraph{(Assumption 3)}
Modern locally unordered GNNs, including GCN, GAT, GraphSAGE, and GIN, often satisfy this assumption in practice \cite{garg2020generalization}. If practitioners adopt novel models outside this family, potentially unbounded Rademacher complexity could elevate the risk of overfitting to source-specific patterns and degrade adaptation performance. To mitigate this risk, GGDA’s algorithmic components can be actively tuned to regularize the learning process. Specifically, increasing $\eta$ helps avoid overconfident decisions on out-of-distribution samples, while reducing $\kappa$ ensures finer inter-domain steps to ease model adaptation. Additionally, lowering $\beta$ retains more past samples, thereby increasing the effective sample size in each intermediate domain and further counteracting overfitting. Through such appropriate tuning, GGDA can help limit overfitting even when the theoretical bound is not tight.

\paragraph{(Assumption 4 and 5)}
Lipschitz assumptions are standard in deep learning literature for deriving tractable generalization bounds \cite{shalev2014understanding}. Empirically, while strict global Lipschitzness may not be guaranteed, GGDA maintains strong performance across complex, real-world datasets, indicating that the theoretical insights retain practical relevance. This empirical success is supported by design choices that inherently promote Lipschitz-like behavior, including normalized aggregation in message passing, intrinsically 1-Lipschitz components (e.g., ReLU activations and softmax transformations), and implicit smoothing via regularization (e.g., dropout and weight decay).

\begin{algorithm*}[t]
\caption{The Proposed GGDA Framework.}
\label{alg:algorithm}
\begin{flushleft}
\textbf{Input}: labeled source graph $\mu_S$, unlabeled target graph $\mu_T$, number of partitions $P_S$ and $P_T$, number of intermediate graphs $K-1$, tolerance of unlabeled ratio in target graph $r_u^T$, score penalty $\eta$, selection ratio $\kappa$, mass decay $\beta$\\
\textbf{Output}: node classification predictions on target graph
\end{flushleft}
\begin{algorithmic}[1] 
\State Partition $\mu_S$ and $\mu_T$ into $P_S$ and $P_T$ subgraphs by METIS.
\For{$P_t \in \{1,2,...,P_T\}$}
    \For{$m(P_t) \in \mathcal{M}(P_t)$}
        \State Compute the information loss $S_{loss}(m(P_t),P_t)$ by Eq. (7) and (8).
    \EndFor
    \State Set the preliminary matching to $\dot{m}(P_t)=\argmin_{m(P_t)\in\mathcal{M}(P_t)}S_{loss}(m(P_t),P_t)$.
\EndFor
\For{$k = 1:(K-1)$}
    \State For $P_t \in \{1,2,...,P_T\}$, set $m(P_t)$ to $\dot{m}(P_t)$ with a probability $\propto 1/S_{loss}(\dot{m}(P_t),P_t)$, and set it to a new matching otherwise.
    \State Generate intermediate graph $\tilde{\mu}_k$ over the FGW metric by Eq. (9).
    \State Using results from graph generation, compute $S_{loss}(m(P_t),P_t)$ by refined Eq. (7) and (8) w.r.t. intermediate graph entropy.
    \State Update the matching $\dot{m}(P_t)=m(P_t)$ if $S_{loss}(m(P_t),P_t) < S_{loss}(\dot{m}(P_t),P_t)$.
\EndFor
\State Define $\mu_B$ as the graph sequence, i.e., $\mathcal{V}_B := supp(\mu_B) = \bigcup_{k=1}^{K-1} supp(\tilde\mu_k) \cup supp(\mu_S) \cup supp(\mu_T)$.
\State Initialize mass decay mask $w^0_b = 1$ for all $v_b \in \mathcal{V}_B$.
\State Initialize the first domain $\mu_0 = \mu_S$.
\State Initialize $t = 0$.
\While{$1 - |supp(\mu_t) \cap supp(\mu_T)| / |supp(\mu_T)| \geq r_u^T$}
    \State Define $\mathcal{V}^t_l := supp(\mu_t)$; $\mathcal{V}^t_u := \mathcal{V}_B\symbol{92}(\bigcup_{t'=0}^{t}\mathcal{V}^{t'}_l)$.
    \State Initialize and train $\theta_t$ on $\mu_t$ with loss $\mathbb{E}_{x,a,\hat{y} \sim \mu_t}\big[\ell(M_{\theta_t}(x,a), \hat{y})\big]$.
    \State Compute the regularized score $c_u^t$ for $v_u \in \mathcal{V}^t_u$ using Eq. (12).
    \State Sort $v_u \in \mathcal{V}^t_u$ by $c_u^t$ and choose top-scored ones within constraint $\kappa$ as $\mathcal{V}^t_{pl}$. Assign pseudo-labels $\hat{y}_{pl} = sign(M_{\theta_t}(x_{pl}, a_{pl}))$ for $v_{pl}\in \mathcal{V}^t_{pl}$.
    \State Define $\mathcal{V}_{pll}^t  := \mathcal{V}_l^t \cup \mathcal{V}_{pl}^t$. 
    \State Store the label score $\hat{c}^t_{pll} = \hat{y}_{pll} \cdot M_{\theta_t}(x_{pll}, a_{pll})$ for $v_{pll} \in \mathcal{V}_{pll}^t$.
    \If {$t \geq 1$}
        \State Compute the mass decay $\lambda_l^t$ for $v_l \in \mathcal{V}^t_l$ using Eq. (14).
        \State Update the mask $w^{t}_l = w^{t-1}_l \cdot \lambda_l^t$ for $v_l \in \mathcal{V}^t_l$.
    \EndIf
    \State Define the next domain distribution $\mu_{t+1}$ using Eq. (15).
    \State Set $t = t+1$.
\EndWhile
\State Apply $M_{\theta_t}$ on $\mu_T$ to obtain classification predictions.
\end{algorithmic}
\label{alg}
\end{algorithm*}

\paragraph{(Binary vs. Multi-Class Classification)} The binary classification assumption in Section 3.2 is purely a writing simplification for notational convenience. It is not among the formal assumptions listed in Section 4 that underpin our proofs. Therefore, it does not restrict the mathematical validity of our theoretical results to binary tasks. Nevertheless, we briefly discuss some technical nuances between binary and multi-class settings for clarity. In our theoretical derivations (Appendices \ref{proof1}, \ref{proof2}, \ref{proof3}, and \ref{proof4}), the label space distance is denoted as the scalar absolute difference $|y_i - y_j'|$. In a multi-class setting with $C$ classes, the label space simply becomes a vector space (i.e., $\mathbb{R}^C$), and this scalar difference is replaced by a standard vector norm $\|y_i - y_j'\|_p$. Because any valid vector norm inherently satisfies the triangle inequality, Minkowski’s inequality, and Hölder’s inequality --- which are the exact foundational inequalities our proofs rely on --- every derivation step in Propositions 2, 3, and 4 holds identically for both binary and multi-class settings. The structure of the bounds remains completely unchanged. Similarly, for the generalization error bound (Appendix \ref{proof1}), extending the scalar Rademacher complexity to vector-valued (multi-class) functions is a standard result in statistical learning theory. It merely introduces a constant scaling factor (dependent on $C$) to the complexity term, while the core theoretical dynamic of our GGDA framework --- the trade-off between the number of intermediate domains $T$ and the domain gap $\Delta$ --- remains structurally identical.

\section{Algorithm}\label{apd_alg}
The algorithm of our proposed GGDA framework is shown in Algorithm \ref{alg}. The indexing of equations corresponds to that
in the main text.

\section{Datasets and Statistics} \label{apd_dataset}
Below are the detailed descriptions of the datasets. All graphs are converted into undirected ones.
\begin{itemize}
\item\textbf{ACM}, \textbf{Citation}, and \textbf{DBLP} (denoted by \textbf{A}, \textbf{C}, and \textbf{D}): citation networks extracted from ArnetMiner, where each node represents a paper and each edge represents a citation relationship. The node attributes are sparse bag-of-words features and the label is the research areas. The networks originate from ACM (pre-2008), Microsoft Academic Graph (post-2010), and DBLP (2004 to 2008) respectively \cite{shen2020adversarial}. Each node is assigned multiple labels to represent the relevant research areas.
\item\textbf{ACM2} and \textbf{DBLP2} (denoted by \textbf{A2}, \textbf{D2}): another pre-processed version of citation networks from ACM (2000 to 2010) and DBLP (post-2010) \cite{wu2020unsupervised}. Different from the version above, each node is assigned only a single label to represent the relevant research area.
\item\textbf{Blog1} and \textbf{Blog2} (denoted by \textbf{B1}, \textbf{B2}): social networks from BlogCatalog \cite{li2015unsupervised}, with each node representing a blogger and each edge representing a friendship. The node attributes are keywords extracted from the blogger's self-description and the label is the blogging category. Following previous work \cite{shen2020adversarial}, 30\% of the binary attributes in each network are randomly selected and flipped to enlarge domain discrepancy.
\item \textbf{Arxiv}: Arxiv computer science citation networks partitioned by the time \textbf{1950-2007}, \textbf{1950-2016}, and \textbf{1950-2018} \cite{hu2020open, liu2024pairwise}. The node attributes are word2vec vectors with the average embedding of the paper’s title and abstract, and the label is the research area. 
\item \textbf{Cora} and \textbf{CiteSeer}: benchmark citation networks for node classification tasks \cite{sen2008collective}, where the nodes, edges, attributes, and labels are defined similarly as the citation networks above. During shift generation, the shift vector at each step is scaled to be 3 unit-norm.
\item \textbf{ENGB, DE}, and \textbf{PTBR}: benchmark social networks with different densities \cite{rozemberczki2021multi}, where each node represents a Twitch user and each edge represents a friendship. The node attributes are user descriptions and the label is an indicator of whether the streamer uses explicit language. During shift generation, the shift vector at each step is scaled to be 5 unit-norm.
\end{itemize}

\section{Baseline Details} 
Below are the descriptions of all baselines in this paper:
\begin{itemize}
\item \textbf{CDAN} \cite{long2018conditional}: employs multi-linear conditioning to capture the cross-covariance between feature representations and classifier predictions; here the vanilla encoder is replaced by a graph encoder;
\item \textbf{MDD} \cite{zhang2019bridging}: incorporates margin disparity discrepancy (MDD) tailored to the distribution
comparison with the asymmetric margin loss; here the vanilla encoder is replaced by a graph encoder;
\item \textbf{UDA-GCN} \cite{wu2020unsupervised}: encodes both local and global consistency within the graphs by incorporating PPMI-based convolutions \cite{zhuang2018dual} and captures semantic information on the unlabeled target domain via an extra entropy loss;
\item \textbf{ACDNE} \cite{shen2020adversarial}: utilizes dual feature extractors and an additional pairwise constraint to preserve node proximity;
\item \textbf{ASN} \cite{zhang2021adversarial}: separates domain-private and domain-shared information by adding a private encoder for each network and forcing them to extract different features;
\item \textbf{GRADE} \cite{wu2023non}: incorporates an encoder that captures subtree discrepancy from the perspective of the Weisfeiler-Lehman graph isomorphism test;
\item \textbf{StruRW} \cite{liu2023structural}: proposes an edge-reweighing scheme upon the source graph to overcome the conditional structure shift (CSS) between graphs;
\item \textbf{SpecReg} \cite{you2023graph}: employs spectral regularization on spectral smoothness and maximum frequency response to modulate the GNN Lipschitz constant;
\item \textbf{GraphAlign} \cite{huang2024can}: generates a new source graph by aligning its distribution with the target graph while retaining information from the original source graph;
\item \textbf{A2GNN} \cite{liu2024rethinking}: introduces an asymmetric encoder architecture for the source and target graphs;
\item \textbf{Pair-Align} \cite{liu2024pairwise}: extends StruRW with a bootstrapping process of edge weight assignment and an adjustable classification loss to address label shift;
\item \textbf{TDSS} \cite{chen2025smoothness}: performs structural smoothing on target graph by neighboring node generation and Laplacian smoothness
constraint to mitigate structural distribution shifts.
\end{itemize}

\section{Implementations Details}
By the definition of unsupervised graph domain adaptation, only labeled samples on the source graph are used for training. For the labeled samples on the target graph, we randomly hold out 20\% for validation, and the remaining 80\% are used for testing \cite{liu2023structural}. The best model is searched on the validation set, and the final experiment outcomes are reported on the test set. All reported outcomes are summarized over 10 independent trials with fixed random seeds, ensuring consistent data splits across all compared methods.

For a fair comparison, we employ the same hyperparameter configurations for the graph encoder among all methods. Specifically, the number of hidden layers is set to 2 for all datasets except for Blogs, which is set to 1. The hidden dimension is fixed at 128 on all layers. Depending on the datasets, the learning rate is chosen from \{1e-3, 5e-4, 1e-4\}, and the number of training epochs is chosen from \{200, 1000, 2000\}. An Adam optimizer with a weight decay rate of 5e-4 is applied. Dropout is employed at a rate of 0.5 to prevent model overfitting. 

For our GGDA framework, when generating the FGW-based intermediate graphs, the trade-off parameter between feature and structure (i.e., $\alpha$) is searched within \{0.05, 0.5, 0.95\}, while the partition size is searched within \{150, 300, 1000\}. The maximum number of iterations is set to 30 for updating $(X_I, C_I)$ and 300 for optimizing the couplings $(\pi(I,S), \pi(I,T))$. If both objective errors are below 1e-5, the algorithm terminates as well. Our proposed framework is implemented with Python 3.9, PyTorch 2.0.0, and Python Optimal Transport (POT) 0.9.1. We use a machine with Nvidia GeForce RTX 4090 GPU and Intel Core i7-13700 CPU.

\section{Visualization of Target Representation Space}
Figure \ref{emb_cls} illustrates the final target representation space given by the symbolic DIR model, UDA-GCN, and our proposed GGDA framework. Specifically, we project the node embeddings onto a two-dimensional space via the t-SNE algorithm. In the visualizations, the node embeddings produced by UDA-GCN exhibit greater overlap and mixing between different classes, while the ones generated by GGDA demonstrate a well-separated clustering with more pronounced class boundaries. The improved ability of GGDA to learn class-discriminative target representations is attributed to its careful domain sequence constructions that aim for the optimal adaptation path to minimize information loss and reduce irrelevant noise. The preservation of transferable information from the labeled source graph allows GGDA to achieve stable transfer under arbitrary source-target discrepancies.
\begin{figure}
    \setlength{\tabcolsep}{1.2em}
        \begin{tabular}{cc} 
        \includegraphics[height=82pt]{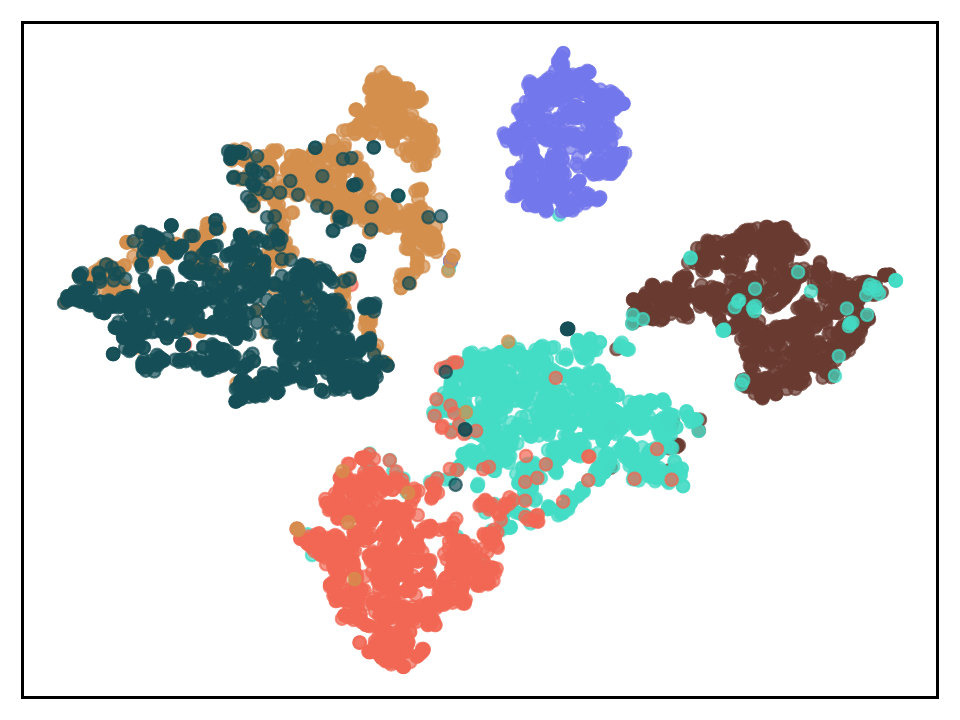} & 
        \includegraphics[height=82pt]{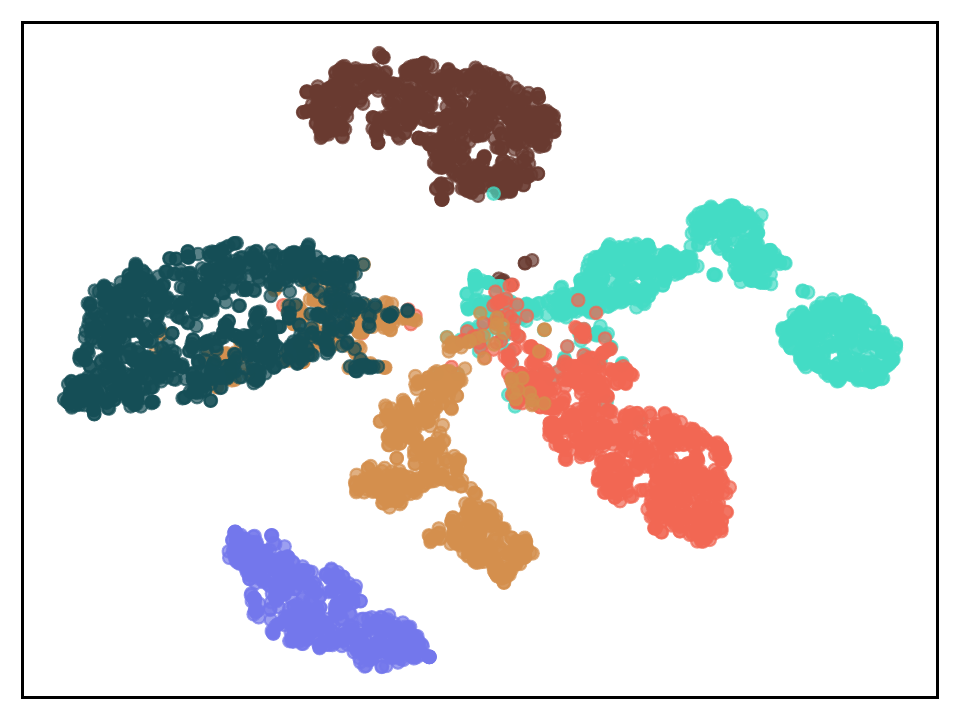}
        \end{tabular}
    \caption{Visualization of target graph node embeddings for the task A2 to D2, where distinct colors signify different class labels. The left side shows the result from UDA-GCN and the right side shows that from GGDA-GCN.}
    \Description[Two t-SNE scatter plots comparing the node embedding clusters of UDA-GCN and GGDA-GCN.]{Two side-by-side scatter plots show the 2D t-SNE projection of target node embeddings for the A2 to D2 task, with colors representing different class labels. The left plot (UDA-GCN) shows clusters that are loosely formed with significant overlap and mixing between different colors. In contrast, the right plot (GGDA-GCN) shows much tighter, well-separated clusters with clear white space between different color groups. This visual difference highlights GGDA's superior ability to learn discriminative target representations with pronounced class boundaries.}
    \label{emb_cls}
\end{figure}

\section{Analysis: The Case of Pure Structural Shift}
To investigate the handling of pure structural shifts, we conduct an experiment using synthetic datasets generated via a contextual stochastic block model (CSBM) \cite{deshpande2018contextual}. We first construct identical source and target graphs, each containing 300 nodes evenly distributed among three classes. Node attributes for each class are sampled from three Gaussian distributions: $\mathbb{P}_0=\mathcal{N}([0, -6], I/2)$, $\mathbb{P}_1=\mathcal{N}([0, 0], I/2)$, and $\mathbb{P}_2=\mathcal{N}([0, 6], I/2)$. In both graphs, the intra-class edge probability is \begin{wrapfigure}{r}{0.4\textwidth}
    \centering
    \includegraphics[height=180pt]{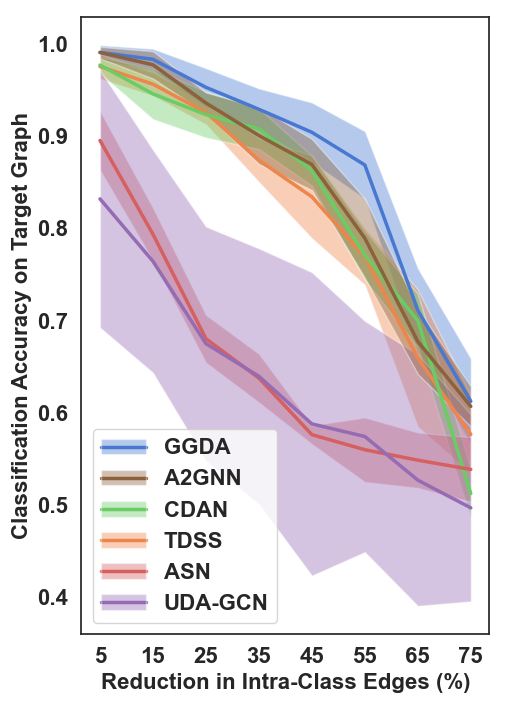}
    \caption{Target graph classification accuracy for GGDA and five strongest baselines as intra-class edges are progressively removed from the target graph (x-axis). Node attributes are held fixed across domains to isolate the effect of structural distribution shift.}
    \Description[A line graph showing model accuracy as intra-class edges are removed from the target graph, comparing GGDA against five strong baselines.]{A line plot illustrates target graph classification accuracy (y-axis, 0.4–1.0) against the percentage reduction in intra-class edges (x-axis, 5\%–75\%), with shaded areas around the lines indicating the variance. The blue line representing GGDA remains the highest and most stable across the entire range, maintaining accuracy above 0.85 until a 55\% reduction. Baselines like UDA-GCN and ASN show sharp, early performance drops, while others like A2GNN and TDSS are more robust but still consistently underperform compared to GGDA.}
    \label{structural_shift}
\end{wrapfigure} set to 20\%, while the inter-class edge probability is set to 2\%. 

To introduce a controlled structural shift, we modify the target graph by randomly adding 2\% inter-class edges and progressively removing intra-class edges --- simulating a gradual reduction of homophily. The node attributes remain unchanged, isolating the effect of structural distribution gaps. The degree of shift is quantified by the percentage reduction in intra-class edge probability, as shown on the x‑axis of Figure \ref{structural_shift}.

Figure \ref{structural_shift} compares the performance under pure structural shift across GGDA and the five strongest baselines. The results reveal distinct patterns of robustness. Methods such as ASN and UDA‑GCN exhibit significant performance degradation as the structure changes. This sensitivity likely stems from UDA‑GCN’s reliance on a global PPMI matrix and ASN’s propensity to form subclusters --- both of which are highly vulnerable to structural perturbations. In contrast, A2GNN, CDAN, and TDSS demonstrate greater robustness. A2GNN performs structural aggregation only on the target graph, which mitigates the impact of cross‑domain structural discrepancies. TDSS extends A2GNN while enforcing feature smoothness among neighboring nodes, thereby reinforcing community structure and becoming less sensitive to intra‑class edge reduction. CDAN avoids mode mismatch by conditioning its domain discriminator on classifier uncertainty, prioritizing easy‑to‑transfer examples (e.g., community centers) over noisy peripheral connections. Nevertheless, GGDA surpasses all baselines in stability under increasing structural shift. This advantage stems from its explicit modeling of gradual structural transitions between source and target domains. Unlike other DIR baselines that attempt abrupt cross-domain marginal structural alignment, GGDA enables safer and more structured knowledge transfer over the manifold.

\section{Analysis: Pseudo-Label Accuracy in the GGDA Domain Sequence}\label{PL_acc}
To examine how GGDA controls information loss during adaptation, we analyze pseudo-label accuracy across constructed intermediate domains. Since ground-truth labels for these domains are unavailable, we conduct this analysis on \emph{multi-step shifting} datasets, where ground-truth intermediate graphs with node labels are available (i.e., Cora and CiteSeer). This enables us to estimate pseudo-label accuracy for each GGDA-constructed domain by matching its samples to the nearest embedded points from the ground-truth intermediate sequence and comparing their labels. Results for Cora and CiteSeer are shown in Figure \ref{inter_PL}.

As expected, pseudo-label accuracy decreases as domains progress from source to target --- a natural characteristic of gradual domain adaptation (GDA). The key contribution of GGDA, however, is not to prevent this decay entirely, but to control its rate such that accumulated error is minimized when the model finally reaches the target. As observed from the experiment, higher quality of the initially constructed domains helps improve the overall quality of later constructed domains and curbs the accuracy decay. This quality is influenced by factors such as encoder selection and inter-domain compactness within GGDA.

\begin{wrapfigure}{r}{0.55\textwidth}
    \centering
    \setlength{\tabcolsep}{0em}
    \begin{tabular}{cc} 
    \includegraphics[height=150pt]{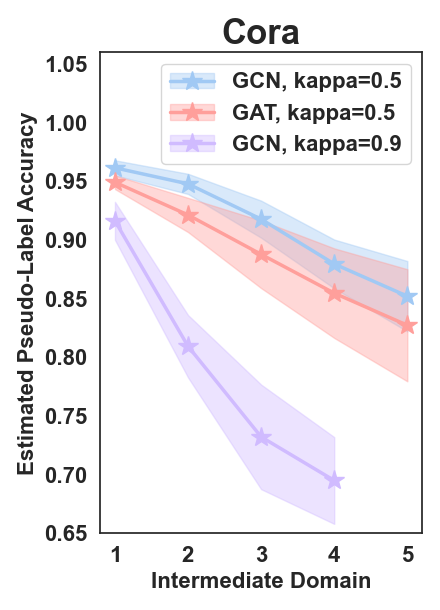} &
    \includegraphics[height=150pt]{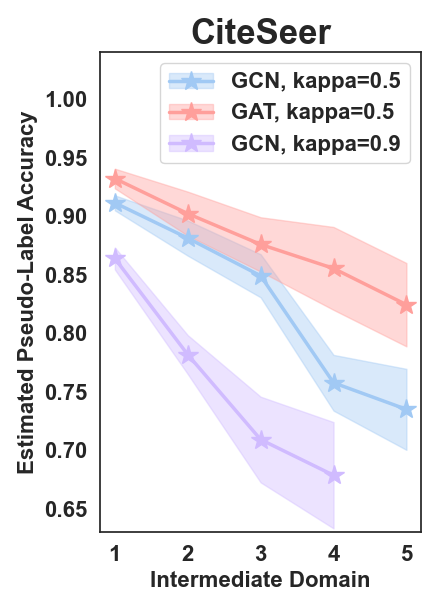}
    \end{tabular}
    \caption{Estimated pseudo-label accuracy across GGDA intermediate domains on Cora and CiteSeer, evaluated under encoder and $\kappa$ variations, where larger $\kappa$ results in fewer domains due to accelerated progression.}
    \Description[Line graphs of pseudo-label accuracy across GGDA intermediate domains for Cora and CiteSeer]{The figure displays two line graphs illustrating the estimated pseudo-label accuracy across intermediate domains (1 to 5) for the Cora and CiteSeer datasets. The x-axis represents the intermediate domain sequence, while the y-axis shows the estimated pseudo-label accuracy. Both graphs compare GGDA in three settings: GCN with kappa=0.5, GAT with kappa=0.5, and GCN with kappa=0.9. Shaded regions indicate variance. In both datasets, accuracy decreases as domains progress. For Cora, GCN with kappa=0.5 maintains the highest accuracy, dropping gradually from approximately 0.96 to 0.85. Conversely, on CiteSeer, GAT with kappa=0.5 performs best, decreasing from about 0.93 to 0.82. In both graphs, the GCN variant with kappa=0.9 setting exhibits a significantly sharper decline in accuracy compared to the other two variants.}
    \label{inter_PL}
\end{wrapfigure}For instance, GCN outperforms GAT on Cora, while the opposite holds on CiteSeer, underscoring the need to align encoder inductive bias with graph characteristics to preserve discriminative signals. Meanwhile, the Wasserstein distance $W_p(\mu_t, \mu_{t+1})$ controlled via $\kappa$ directly affects information retention. When $\kappa$ is too large (e.g., $\kappa=0.9$), the domain gap widens, causing sharp initial accuracy drops that propagate errors forward. Conversely, a moderate $\kappa$ ensures gradual transitions, sustaining higher pseudo-label consistency across domains.

In essence, GGDA strategically slows the decay of pseudo-label accuracy through its cohesive design and tunable progression mechanisms, which together construct a compact domain sequence that minimizes information loss and curbs error accumulation. This results in a more reliable model at the final adaptation step, leading to the superior target accuracy observed in our experiments.

\section{Analysis: Guiding Encoder Selection for GGDA}
In this section, we provide actionable metrics to guide encoder selection for GGDA in a new GDA task with unlabeled target. We discuss three complementary indicators --- source‑domain accuracy, uncertainty calibration on early shifted domains, and graph structural statistics --- that explain the performance differences among GCN, GAT, and GraphSAGE in GGDA, and help practitioners choose the most suitable encoder for a new transfer task.
\paragraph{Source-Domain Accuracy.} Since the target domain is unlabeled, one reliable prior indicator is the encoder's performance on the labeled source graph. As briefly discussed in Appendix \ref{PL_acc} ("Analysis: Pseudo-Label Accuracy in the GGDA Domain Sequence"), the quality of the initial model directly bounds the subsequent pseudo-labeling quality. Figure \ref{encoder_choice} ("Src." column) demonstrates a positive correlation between an encoder’s source‑accuracy (trained with 40\% of source labels) and its final GGDA target accuracy. Practitioners should first evaluate candidate encoders on the source graph and prioritize those with the higher accuracy while discarding those with notably low accuracy.

\paragraph{Uncertainty Calibration on Shifted Domains.} A more nuanced indicator is the encoder’s generalization behavior under the early distribution shifts encountered in GGDA. We measure the average drop in maximum softmax probability (MSP), i.e., average increase in uncertainty, when applying the source‑trained model to the first three intermediate domains (column “$\Delta$Uncer.” in Figure \ref{encoder_choice}). As established in out‑of‑distribution (OOD) literature \cite{hendrycks2016baseline}, an encoder that exhibits very little MSP decay (near‑zero drop) is often overconfident and overfitted to source‑specific spurious patterns; it will collapse as the domain shift accumulates. For instance, in the A2 to D2 task, GraphSAGE exhibits high source accuracy but fails during adaptation, a vulnerability successfully flagged by its abnormally low $\Delta$Uncer (0.0158). Conversely, an encoder that shows excessive MSP decay fails to capture transferable structure and generalizes poorly. The ideal encoder for GGDA maintains a moderate, calibrated MSP drop (empirically between 0.02 and 0.10 in our datasets) --- one that reflects a realistic increase in uncertainty under a gradual shift. Practitioners should avoid the two extremes and select the encoder that demonstrates stable, moderately decaying MSP across the first few shifted domains.

\paragraph{Graph Structural Statistics.} Beyond empirical metrics, certain global properties of the source and target graphs (density, homophily, fragmentation) can predict how an encoder will behave during the progressive domain adaptation in GGDA. These properties affect not only static representation learning but also the stability of pseudo‑label propagation across constructed domains.
\begin{enumerate}
    \item {Dense graphs (e.g., Blog datasets):} Avoid GAT in dense domain sequences, as its attention mechanism often overfits to spurious neighbor connections, leading to unstable pseudo‑labels in GGDA. Instead, select GCN, which acts as a stable low-pass filter to smooth structural noise, or GraphSAGE, whose neighbor sampling regularizes excessive noise from dense hub-nodes.
    \item {Sparse graphs (e.g., citation networks A, C, D, A2, and D2):} Avoid GraphSAGE in sparse domain sequences. These graphs rely on the propagation of sparse but informative structural signals. GCN’s symmetric normalization ($1/\sqrt{d_u d_v}$) and GAT's normalized attentions amplify connections between low‑degree or important nodes, preserving discriminative structure across intermediate domains. GraphSAGE, lacking such amplifications and using random neighbor sampling, introduces unnecessary variance that destabilizes inter-domain knowledge transfer in GGDA.
    \item {Fragmented or low-homophily graphs (e.g., CiteSeer):} Avoid GCN in these domain sequences. In highly fragmented or low-homophily graphs, a node's own features are often more reliable than its local structure. GCN forcibly blends ego‑features with neighbor features using fixed weights, diluting useful signals. In contrast, GAT can assign high attention to self‑loops, and GraphSAGE explicitly concatenates ego‑features, both preventing error accumulation in adaptation when the domain sequence has weak or heterogeneous structural cues.
\end{enumerate}

\begin{figure*}[t]
    \centering
    \includegraphics[height=43pt]{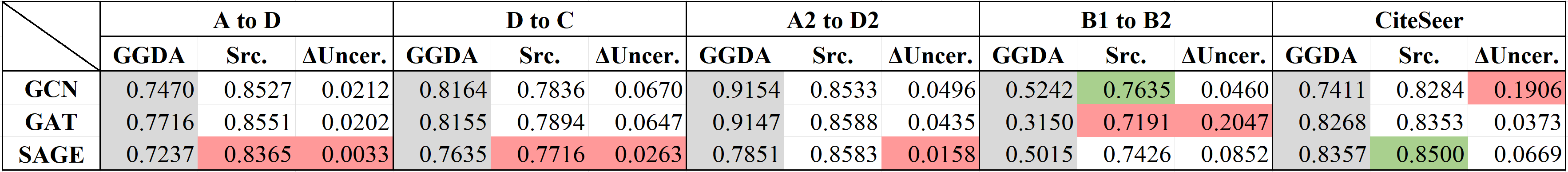}
    \caption{Correlation analysis between GGDA performance (grey) and two predictive metrics --- source‑domain accuracy (“Src.”) and average increase of uncertainty (drop in MSP, “$\Delta$Uncer.”) --- across five transfer tasks and three encoders. Red indicates signals for an unsuitable encoder (low source accuracy or extreme uncertainty increase), while green indicates signals for a suitable encoder. The combination of these two metrics consistently predicts the empirical GGDA performance with different encoders validated on the true target domain.}
    \label{encoder_choice}
\end{figure*}

\begin{figure*}
    \centering
    \setlength{\tabcolsep}{0.66em}
        \begin{tabular}{ccc} 
        \includegraphics[height=80pt]{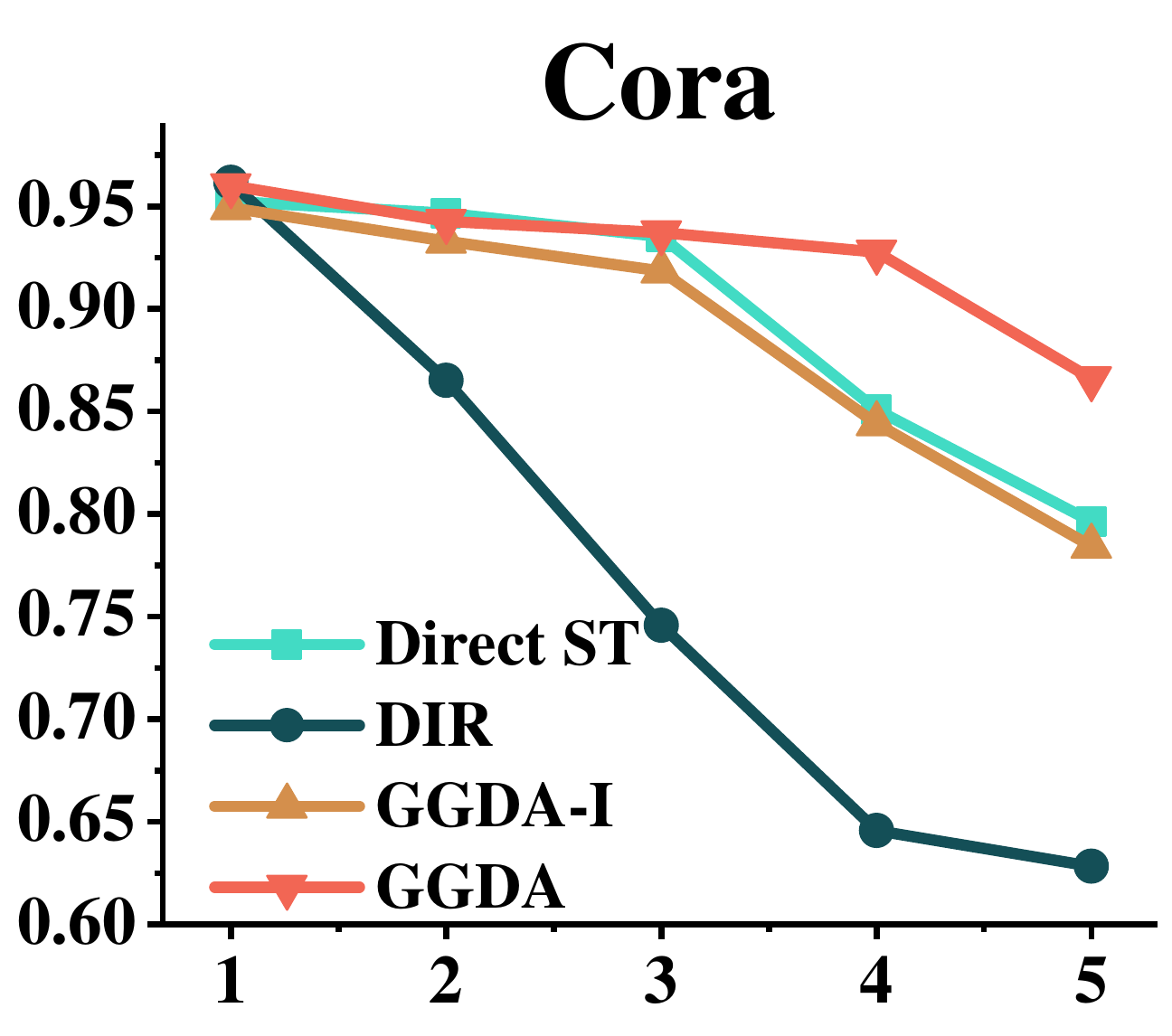} & 
        \includegraphics[height=80pt]{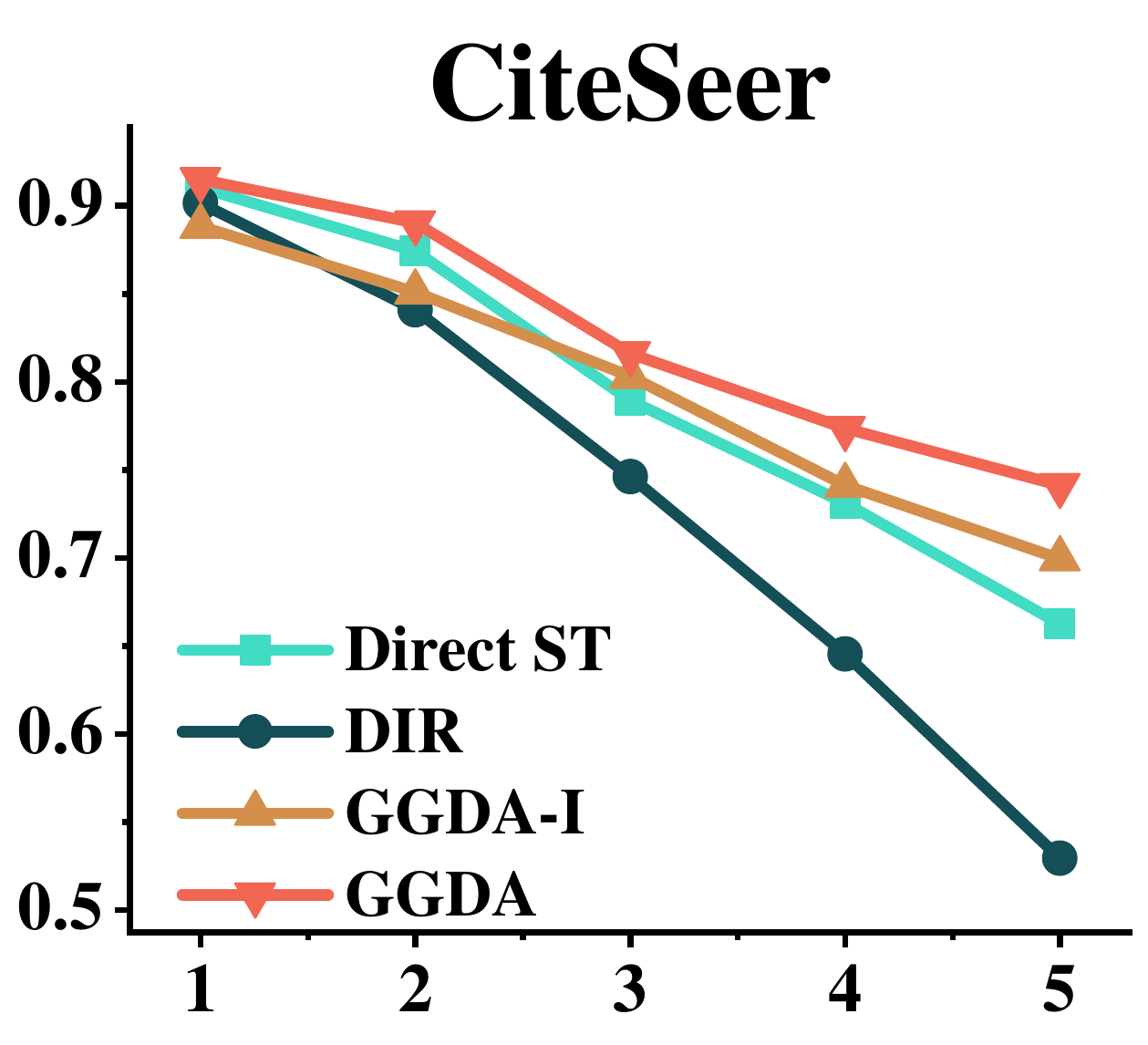} &
        \includegraphics[height=80pt]{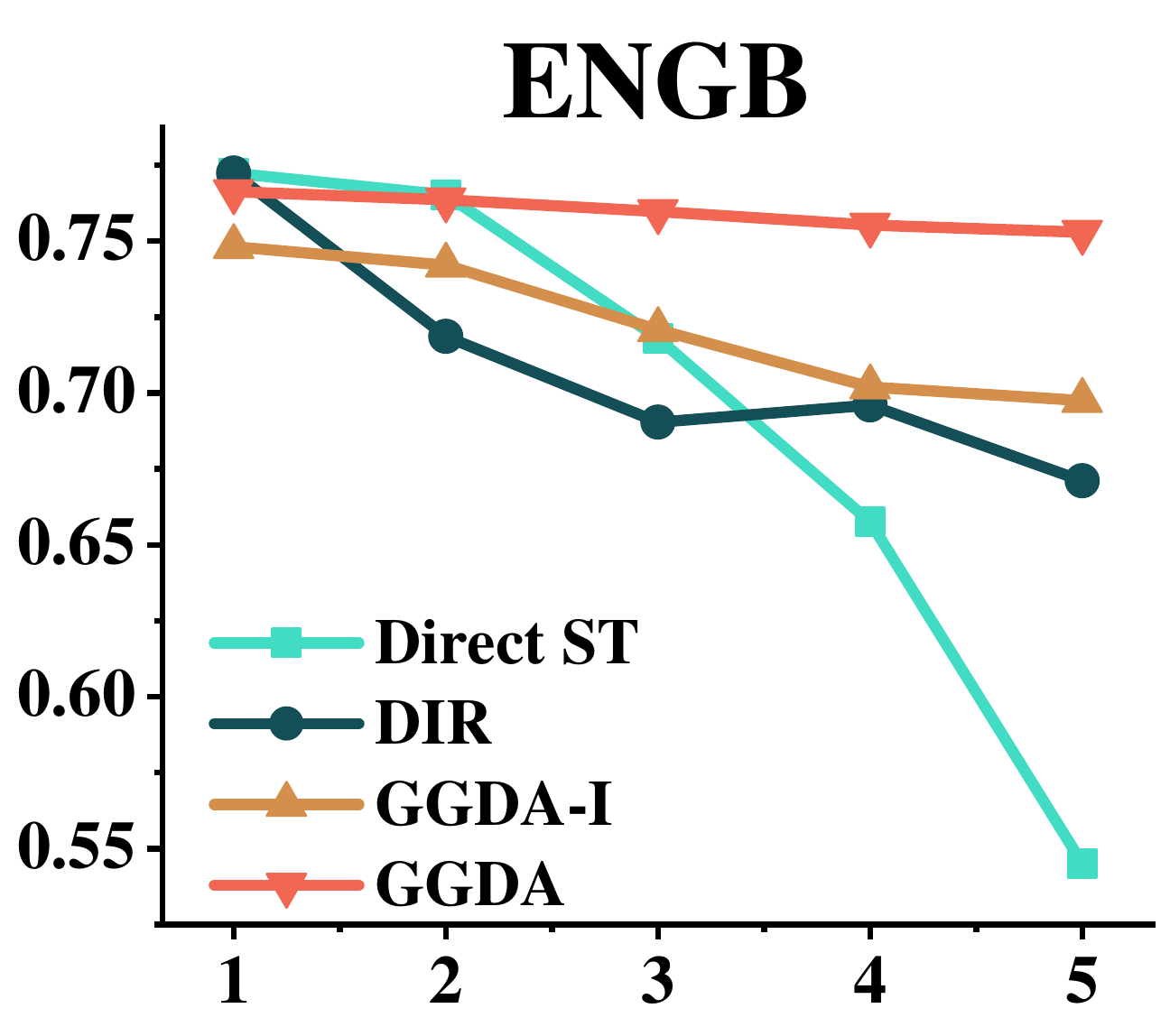} \\ 
        \includegraphics[height=80pt]{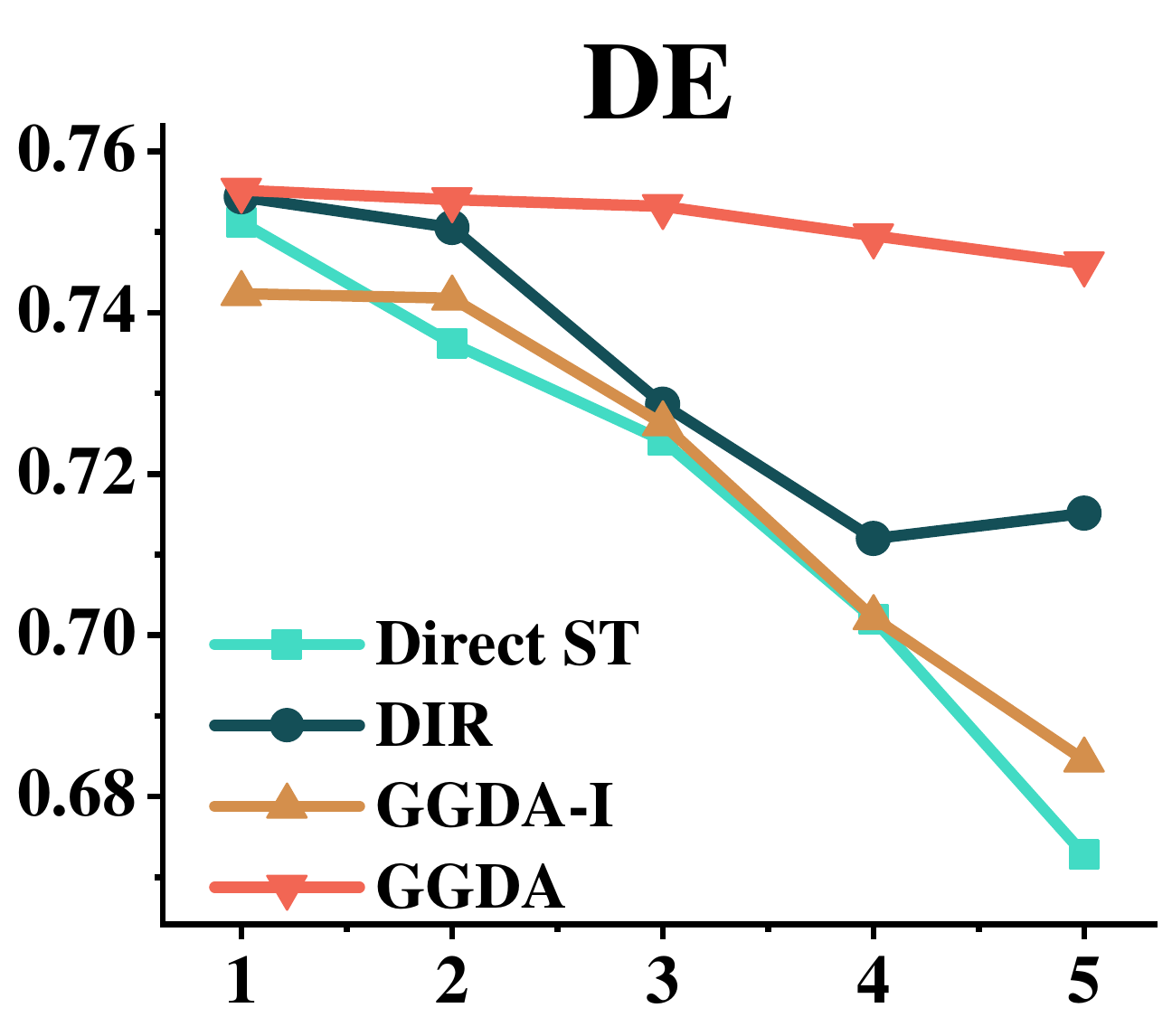} & 
        \includegraphics[height=80pt]{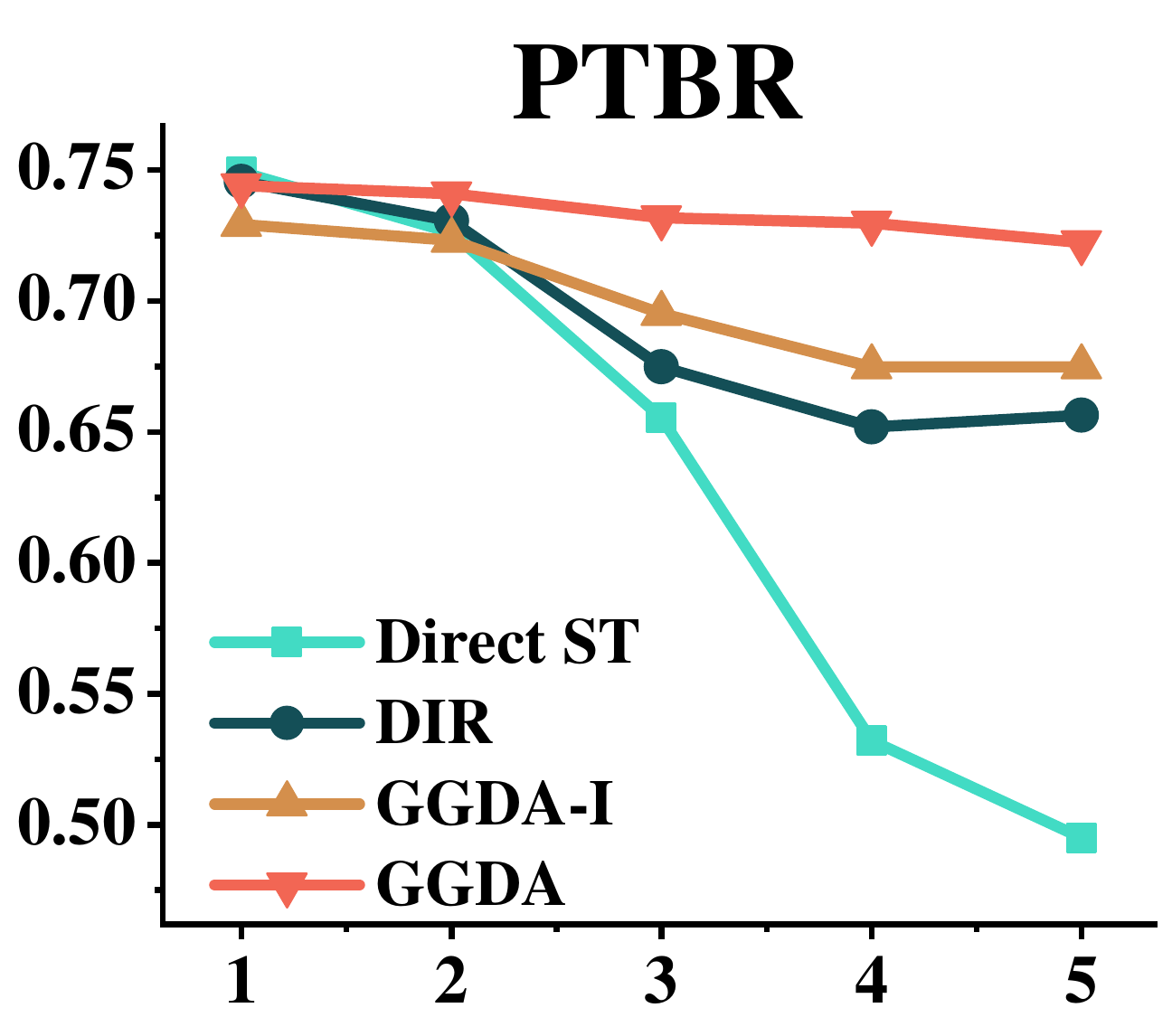} &
        \end{tabular}
    \caption{Node classification accuracy on target graphs with varying levels of discrepancy from the source graph. The x-axis shows the source-target discrepancy level ranked from 1 (closest) to 5 (farthest). The y-axis shows the classification accuracy.}
    \Description[Line graphs of node classification accuracy across varying source-target discrepancy levels]{The figure presents five line graphs detailing node classification accuracy for the Cora, CiteSeer, ENGB, DE, and PTBR datasets. The x-axis represents the source-target discrepancy level, increasing from 1 (closest) to 5 (farthest), while the y-axis displays the resulting target classification accuracy. Four methods are compared: Direct ST, DIR, GGDA-I, and GGDA. Across all datasets, accuracy generally declines as the discrepancy level increases. However, the proposed GGDA framework consistently achieves the highest accuracy and exhibits the most stable performance, showing minimal degradation and noticeable improvements over GGDA-I, Direct ST, and DIR.}
    \label{lineplot_general}
\end{figure*}

\section{Additional Performance Comparison between Different Frameworks}
We further delve into the performance of different general frameworks as the discrepancy level between the source and target graphs increases. As shown in Figure \ref{lineplot_general}, while GGDA-I, which omits the vertex-based domain progression, already demonstrates some improvement over Direct ST by integrating the FGW intermediate graphs, GGDA enjoys further performance and stability boosts by refining the domain construction to preserve richer transferrable information over the gradual domain adaptation. On the other hand, employing DIR could result in worse performance than Direct ST, indicating that the domain alignment technique is only effective within certain distribution contexts. Throughout our empirical assessment, we noticed that adopting DIR under large distribution shifts can potentially form a non-convex problem that is hard to optimize over, causing the optimization to be trapped in local minima and leading to unstable outcomes. 

\section{Potential Extensions of GGDA to Other Graph Tasks}
While GGDA is presented in the context of node-level classification --- the most widely studied setting in graph domain adaptation \cite{liu2024revisiting}, its core principles, namely intermediate graph generation and gradual progression, can conceptually inspire adaptations for other important graph learning tasks. Below we outline how GGDA could be extended, emphasizing necessary reformulations and highlighting open challenges.

In link prediction, the goal is to predict whether an edge exists between a pair of nodes. Unlike node classification, the basic unit of adaptation becomes a node pair $(u,v)$ characterized by pairwise features such as edge attributes $x_{uv}$ (if available) or concatenated node features $[x_u, x_v]$, relational structure such as the local subgraph surrounding the pair, and a binary label $y_{uv}\in [0,1]$. To adapt GGDA to this setting, several key modifications would be required. The domain 
$\mu$ would need to be defined over a product space $\Omega_{pair} \times \Omega_{struct} \times \Omega_y$, where $\Omega_{pair}$ captures pairwise features and $\Omega_{struct}$ encodes the relational context of the pair. The generalization bound in Eq. (5) would need to be re‑derived for an edge‑level loss, taking into account the dependency between linked pairs. Moreover, the vertex selection mechanism would be replaced by pair selection, where pseudo‑labels are assigned to candidate edges based on a link prediction model trained on the current domain. Critically, how the proximity between different linked pairs across domains is quantified dictates the construction of intermediate domains, and this should be guided directly by the pairwise dependencies emerging from the theoretical bound to preserve link-predictive structural motifs. To mitigate the high risk of noisy pseudo-edges under distribution shift, structural heuristics (e.g., common neighbors or Adamic-Adar) could also be integrated to regularize the selection of intermediate edges.

In graph classification, each sample is an entire graph $G_i$ with a label $y_i$. Crucially, graphs are typically treated as independent and identically distributed (IID) samples, which differs fundamentally from the non‑IID setting of node classification. Adapting GGDA to graph‑level adaptation would involve several key adjustments. The domain $\mu$ would become a distribution over graphs rather than over nodes within a graph, though the FGW distance could still be applied directly to compare graph distances. Given the IID nature between graph samples, the theoretical analysis would resemble traditional IID gradual domain adaptation, simplifying the error bound but requiring a reformulation of the Wasserstein distance over graph‑level distributions. The vertex selection step would then be replaced by graph selection, where whole graphs are pseudo‑labeled and added to the next domain. Quantifying proximity between graphs is essential, as it determines how the next domain should be constructed to preserve transferable information such as biochemical motifs in molecules or functional subgraphs in proteins. Accordingly, the graph selection mechanism could be regularized by motif‑matching scores or graph kernel similarities, but, again, the specific choice should follow directly from what the theoretical derivation prescribes (e.g., using the ground metric that minimizes the FGW distance).

Despite the conceptual feasibility, several practical challenges remain. Pseudo‑labeling for edges or whole graphs is inherently noisier than for nodes, so incorporating task‑specific regularization or confidence calibration mechanisms is necessary for ensuring the quality of constructed domains, and the design of such mechanisms should be guided by rigorous theoretical grounding. Moreover, operating on pairs or whole graphs increases computational complexity, calling for further approximations or sampling strategies. We view these extensions as promising avenues for future research, each requiring careful reformulation of GGDA’s theoretical foundations and algorithmic components to respect the unique characteristics of the target task.

\section{More on Related Works}
Below are some further discussions on related works.

\subsection{Graph Domain Adaptation}
Node-level graph domain adaptation leverages a source graph with sufficient labeled nodes to perform inference on a target graph from a shifted distribution with few or no labeled nodes. To address the task, a widely adopted framework is to perform adversarial training and learn a graph encoder that fools the domain discriminator, such that domain-invariant representations can be obtained \cite{zhang2019dane, shen2020adversarial, wu2020unsupervised, zhang2021adversarial, dai2022graph, guo2022learning, mao2022augmenting, qiao2023semi, wu2023non, liu2023structural, you2023graph, shi2023improving, liu2024pairwise, liu2024rethinking, wang2024open, chen2025smoothness}. Among the earliest works, DANE includes a bidirectional training with an LSGAN-based loss function of the discriminator \cite{zhang2019dane}, whereas ACDNE presents the use of dual feature extractors and an additional pairwise constraint to preserve both attributed and topological proximity between nodes \cite{shen2020adversarial}. UDA-GCN encodes both local and global consistency within the graphs by incorporating PPMI-based convolutions \cite{zhuang2018dual}, and it captures semantic information on the unlabeled target domain via an extra entropy loss \cite{wu2020unsupervised}. ASN separates domain-private and domain-shared information by adding a private encoder for each network and forcing them to extract different features \cite{zhang2021adversarial}. GraphAE incorporates dual cross-graph alignment strategies, namely message routing alignment and message aggregating alignment, and presents a framework for transferring from multiple source graphs \cite{guo2022learning}. SGDA applies adaptive shift parameters to each of the source nodes, and to handle a higher level of label scarcity, it performs pseudo-labeling with confidence given by the distance from class centroids \cite{qiao2023semi}. GRADE provides a theoretical analysis of cross-network knowledge transfer from the perspective of the Weisfeiler-Lehman graph isomorphism test and develops a transfer framework based on subtree discrepancy \cite{wu2023non}. StruRW identifies a new type of shift in graph domain adaptation, named conditional structure shift (CSS), and proposes a reweighing scheme upon source graph edges to handle such a shift \cite{liu2023structural}. Pair-Align extends StruRW with a bootstrapping process of edge weight assignment and an adjustable classification loss to address label shift \cite{liu2024pairwise}. SpecReg is an approach based on graph filter theory, where a spectral regularization regarding spectral smoothness and maximum frequency response is employed to modulate the GNN Lipschitz constant for bounding the adaptation error \cite{you2023graph}. A2GNN investigates the effect of propagation layers in transfer performance and introduces a model with an asymmetric architecture for source and target graphs \cite{liu2024rethinking}. TDSS performs structural smoothing directly on the target graph by neighboring node generation and Laplacian smoothness constraint to mitigate structural distribution shifts. \cite{chen2025smoothness}. SDA is a separate domain alignment strategy that addresses open-set graph domain adaptation, where the target domain contains new classes not included in the source domain \cite{wang2024open}.

Meanwhile, some works handle graph domain adaptation from alternative perspectives \cite{zhu2021transfer, zhang2024collaborate}. EGI develops a novel graph training objective based on ego-graph information maximization and analyzes the transferability of the learned GNN regarding the local graph Laplacians \cite{zhu2021transfer}. GraphCTA tackles source-free graph domain adaptation, where the source graph is inaccessible, through iterative model adaptation with updates on node attributes and graph topology based on neighborhood contrastive learning \cite{zhang2024collaborate}. GraphAlign, the first data-centric graph domain adaptation method, explores the alignment principle and the rescaling principle, based on which a new smaller source graph is generated by aligning its distribution with the target graph and incorporating information from the original source graph \cite{huang2024can}.

On the other hand, a separate line of research is graph-level domain adaptation, which focuses on graph classifications and treats each graph as an individual sample, but the corresponding shifts and methods are significantly different from node-level tasks (e.g., can be modeled as IID) \cite{yin2022deal, yin2023coco}.

\subsection{Intermediate Domains for IID Domain Adaptation}
The use of intermediate domains has shown promise in addressing IID domain adaptation tasks \cite{gong2012geodesic, gopalan2011domain, tan2017distant, hsu2020progressive, xu2020adversarial, na2021fixbi, abnar2021gradual, zhu2017unpaired, chen2021gradual, he2023gradual, kumar2020understanding, peyre2019computational, wang2022understanding, tan2015transitive}. For the generation of intermediate domains, some earlier works construct them as generative subspaces along a geodesic path on a specific (e.g., Grassmann) manifold \cite{gong2012geodesic, gopalan2011domain}, while later works tend to generate them on a sample level to better integrate with deep learning techniques \cite{gong2019dlow, xu2020adversarial, na2021fixbi, abnar2021gradual, he2023gradual}. 
For instance, DLOW leverages CycleGAN \cite{zhu2017unpaired} to learn a generator of intermediate samples via weighted adversarial loss that acts as a distributional distance measurement \cite{gong2019dlow}. Another standardized approach is to apply linear interpolation between source and target samples for generating mixup samples \cite{xu2020adversarial, na2021fixbi, abnar2021gradual, he2023gradual}. Specifically, some methods select interpolated samples from source and target domains via an iterative cost-based alignment method with optimal transport \cite{abnar2021gradual, he2023gradual}. 

Intermediate domains have been utilized in multiple ways to enhance the performance of domain adaptation. DM-ADA constrains domain invariance not only on source and target domains but also on intermediate domains by learning a domain discriminator that outputs soft scores for generated images \cite{xu2020adversarial}. FixBi trains a source-dominant model and a target-dominant model with dual intermediate domains, allowing the two models to refine themselves by incorporating each other's outputs \cite{na2021fixbi}. A line of research concentrates on gradual domain adaptation (GDA), where iterative self-training is adopted along a bridge of intermediate domains to gradually shift the source model towards the target domain \cite{abnar2021gradual, kumar2020understanding, chen2021gradual, wang2022understanding, he2023gradual, sagawa2022gradual, shi2024adversarial, tan2015transitive}. Specifically, TTL selects appropriate intermediate domains by assessing domain difficulty and pairwise closeness, then propagates labels by learning overlapping features across domains \cite{tan2015transitive}. IDOL is a coarse-to-fine framework for indexing a sequence of intermediate domains for GDA \cite{chen2021gradual}. Kumar et al. explored the effect of regularization and label sharpening on GDA and derived the first generalization bound on the error \cite{kumar2020understanding}. Later, Wang et al. improved this bound by transforming its linear dependency on the number of intermediate domains from exponential to linear \cite{wang2022understanding}. Shi et al. proposed to replace vanilla self-training with adversarial self-training to address the challenge of incorrect pseudo-labels \cite{shi2024adversarial}.

\subsection{Optimal Transport on Graphs}
The optimal transport (OT) framework describes distribution discrepancy by solving the minimal cost of transporting one measure to another. The Wasserstein distance was first proposed as a metric between distributions dwelling in a common metric space \cite{kantorovich1960mathematical, rubner2000earth}. Later, the Gromov-Wasserstein (GW) distance was developed to measure the distance between structured objects (e.g., non-attributed graphs) via the underlying geometric captured by pair-to-pair transports \cite{peyre2019computational, memoli2011gromov}. To integrate both graph attributes and topology within the metric, Fused Gromov-Wasserstein (FGW) distance was proposed recently as a combination of the two distances above \cite{vayer2019, vayer2020}. 

The OT framework has found versatile applications in graph learning \cite{vayer2019, vayer2020, xu2019scalable, xu2019gromov, chowdhury2021generalized, vincent2021online, chen2020optimal, ma2024fused}. For example, graph clustering and graph partitioning tasks can proceed effectively based on GW or FGW distance \cite{vayer2019, vayer2020, xu2019scalable}. Meanwhile, OT has great potential for solving matching and alignment problems given the coupling computations within its optimization process. This makes it a popular tool for shape matching in computer vision \cite{su2017order, chen2020graph} and for graph matching or node alignment in graph learning \cite{xu2019gromov, xu2019scalable, chowdhury2021generalized}. Specifically, Xu et al. proposed to perform graph matching with a mutual learning and regularization scheme between GW discrepancy and node embeddings, such that the resulting representations encode both the intra-graph topology and inter-graph correspondence \cite{xu2019gromov}. The OT metrics have also been integrated into other graph-based learning tasks such as graph representation learning \cite{chen2020optimal, ma2024fused} and online dictionary learning \cite{vincent2021online}. In particular, FGWMixup performs graph data augmentation with FGW barycenters to enhance the robustness of graph neural networks (GNNs) and improve graph classification accuracy \cite{ma2024fused}. Another line of research focuses on boosting the computational efficiency of graph optimal transports with means such as low-rank restriction on couplings and other relaxed algorithms \cite{peyre2016gromov, li2023convergent}.

\endgroup

\end{document}